\documentclass[10pt,twocolumn,letterpaper]{article}

\usepackage[pagenumbers]{cvpr} 

\usepackage{times}
\usepackage{graphicx}
\usepackage{xspace}
\usepackage{comment}
\usepackage{soul}
\usepackage{pifont}
\usepackage{xparse}
\usepackage{floatpag}
\usepackage{amsmath,bm}
\usepackage{lipsum}

\usepackage{multirow}
\usepackage{rotating}
\usepackage[table]{xcolor}
\usepackage{tabularx}
\usepackage{arydshln}
\usepackage{siunitx, booktabs}

\usepackage{dblfloatfix}
\usepackage{fontawesome}
\usepackage{array}

\newcommand\blfootnote[1]{%
  \begingroup
  \renewcommand\thefootnote{}\footnote{#1}%
  \addtocounter{footnote}{-1}%
  \endgroup
}

\definecolor{cvprblue}{rgb}{0.21,0.49,0.74}
\definecolor{codegreen}{rgb}{0,0.6,0}
\definecolor{teaserblue}{RGB}{53,116,188}
\definecolor{teaserred}{RGB}{223,64,52}
\usepackage[pagebackref,breaklinks,colorlinks,citecolor=cvprblue]{hyperref}

\newcommand{\uio}{\mbox{\sc{Unified-IO}}}
\newcommand{\uiot}{\mbox{\sc{Unified-IO 2}}}
\newcommand{\uiotacronym}{\mbox{\sc{UIO-2}}}

\newcommand{\tablefont}{\small}

\newcommand{\uiolarge}{\mbox{\sc{UIO-2}$_\texttt{L}$}}
\newcommand{\uioxl}{\mbox{\sc{UIO-2}$_\texttt{XL}$}}
\newcommand{\uioxxl}{\mbox{\sc{UIO-2}$_\texttt{XXL}$}}

\newcommand{\vae}{\mbox{\sc{VAE}}}
\newcommand{\vqgan}{\mbox{\sc{VQ-GAN}}}
\newcommand{\boldheader}[1]{\noindent\textbf{#1.}}

\title{Unified-IO 2: 
Scaling Autoregressive Multimodal Models \\with Vision, Language, Audio, and Action}

\author{Jiasen Lu$^1$\footnotemark[1] \quad Christopher Clark $^1$\footnotemark[1] \quad Sangho Lee$^{1}$\footnotemark[1] \quad Zichen Zhang$^{1}$\footnotemark[1] \\ 
Savya Khosla$^2$ \quad Ryan Marten$^2$ \quad Derek Hoiem$^2$ \quad Aniruddha Kembhavi$^1$$^3$ \\
$^1$Allen Institute for AI \quad $^2$ University of Illinois Urbana-Champaign \quad $^3$ University of Washington\\
\tt\small \{jiasenl, chrisc, sanghol, chralesz, anik\}@allenai.org
\\
\tt\small \href{https://unified-io-2.allenai.org/}{unified-io-2.allenai.org}
}

\begin{document}

\twocolumn[{
\renewcommand\twocolumn[1][]{#1}
\maketitle
\vspace*{-0.4in}
\centering
\captionsetup{type=figure}\includegraphics[width=\linewidth]{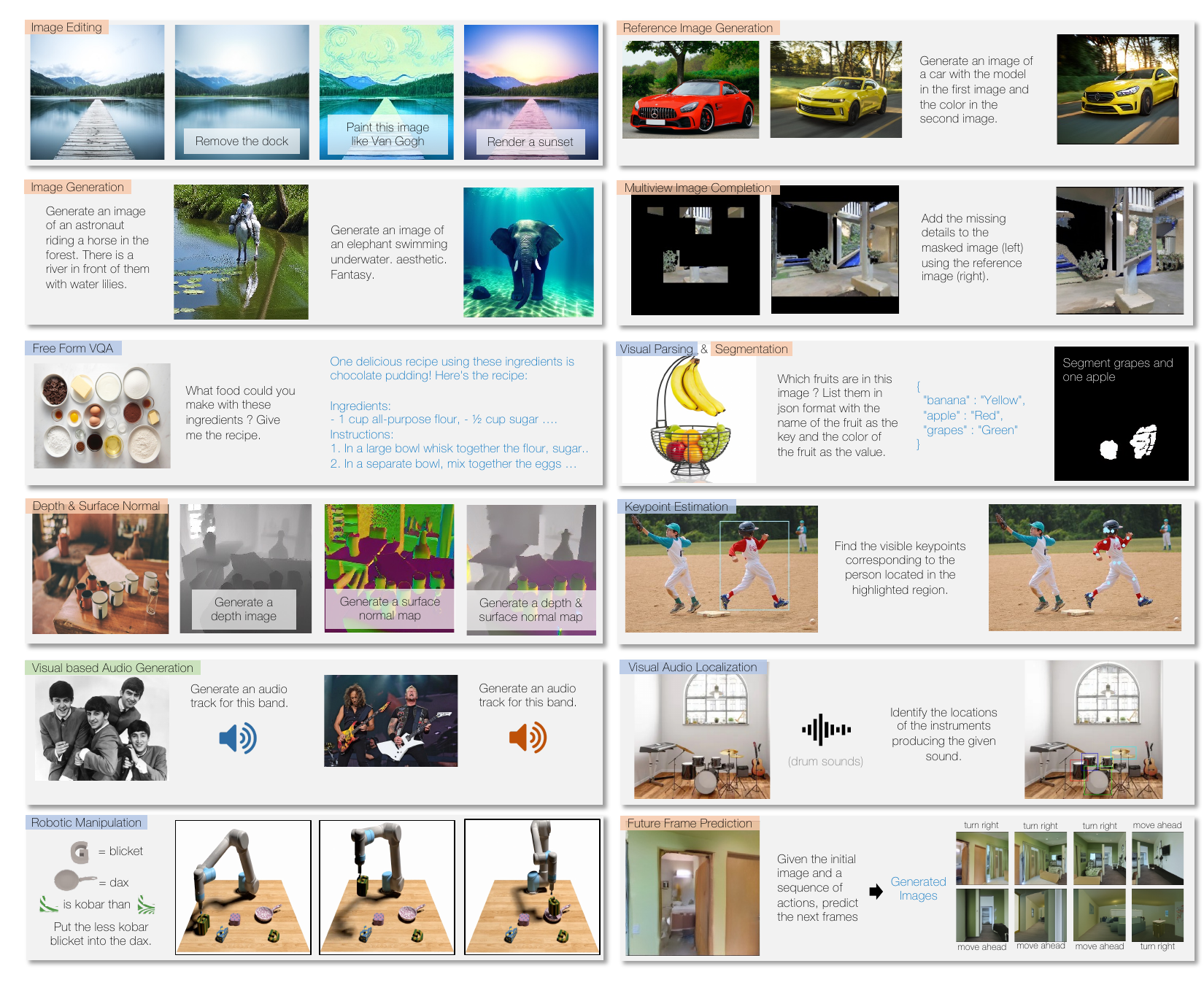}
\vspace{-25pt}
\captionof{figure}{\uiot{} is an instruction-following model with a huge breadth of abilities and supported modalities. It can generate images (\textcolor{teaserred}{red box}), including image editing, image generation, depth estimation, surface normal estimation, and future frame prediction \etc. It can also generate texts (\textcolor{teaserblue}{blue box}), including long-form answers to queries, keypoint estimation, visual audio localization, predicting actions for robotic manipulation \etc. It can generate audio (\textcolor{codegreen}{green box}) from images or text. Click \href{https://ai2-prior-uio.s3.us-west-2.amazonaws.com/public/samples/blue_audio+(beatles).wav}{\textcolor{blue}{\faVolumeUp}} and \href{https://ai2-prior-uio.s3.us-west-2.amazonaws.com/public/samples/red_audio+(metallica).wav}{\textcolor{red}{\faVolumeUp}} for the corresponding audio samples.
}
\label{fig:teaser}
\vspace{-10pt}
}]
\blfootnote{* Leading Authors, equal contribution. A description of each author’s contribution is available in Appendix \ref{supp:contributions}.
Corresponding to \tt Jiasen Lu.}

\maketitle

\clearpage
\begin{abstract}

We present \uiot{}, the first autoregressive multimodal model that is capable of understanding and generating image, text, audio, and action. 
To unify different modalities, we tokenize inputs and outputs -- images, text, audio, action, bounding boxes \etc, into a shared semantic space and then process them with a single encoder-decoder transformer model.
Since training with such diverse modalities is challenging, we propose various architectural improvements to stabilize model training.
We train our model from scratch on a large multimodal pre-training corpus from diverse sources with a multimodal mixture of denoisers objective. 
To learn an expansive set of skills, such as following multimodal instructions, 
we construct and finetune on an ensemble of 120 datasets with prompts and augmentations. 
With a single unified model, \uiot~achieves state-of-the-art performance on the GRIT benchmark and strong results in more than 35 benchmarks, including image generation and understanding, natural language understanding, video and audio understanding, and robotic manipulation. 
We release all our models to the research community.

\vspace{-0.5cm}
\end{abstract}    
\section{Introduction}
\label{sec:intro}

As AI researchers, we seek to build intelligent agents that can perceive their environment, communicate with others, act in the world, and reason about their interactions. The world is multimodal, so our agents must partake in rich interactions that are multimodal in nature via vision, language, sound, action \etc.
Psychologists have argued that the 
redundancy of our sensory systems serves as supervisory mechanisms to improve each other~\cite{piaget1952origins, edelman1993neural, smith2005development}. This provides a natural motivation to create models with similar learning capabilities, supporting many different modalities that can supervise each other during training.

Building models that can parse and produce many modalities is a complex undertaking. Training Large Language Models (LLMs) with billions of parameters, despite only supporting a single modality, is extremely challenging across many fronts -- from sourcing and processing massive datasets, ensuring data quality and managing biases, designing effective model architectures, maintaining stable training processes, and instruction tuning to enhance the model's ability to follow and understand user instructions. These challenges are hugely amplified with the addition of each new modality.

In light of these difficulties, a line of recent works in building multimodal systems has leveraged pre-trained LLMs, with some augmenting with new modality encoders \cite{Alayrac2022FlamingoAV, driess2023palm, llava}, some adding modality specific decoders \cite{koh2023gill, borsos2023audiolm} and others leveraging the LLM's capabilities to build modular frameworks \cite{gupta2023visprog, Singh2022ProgPromptGS, vipergpt}. Another line of works on training multimodal models from scratch has focused on generating text output \cite{peng2023kosmos1, peng2023kosmos2} with a few recent works supporting the understanding and generation of two modalities -- text and images \cite{lu2023chameleon, lu2022unified}. 
Building generative models with a wider coverage of modalities, particularly when training from scratch, remains an open challenge. 

In this work, we present \uiot{}, a large multimodal model (LMM) that can encode text, image, audio, video, and interleaved sequences and produce text, action, audio, image, and sparse or dense labels. It can output free-form multimodal responses and handle tasks unseen during training through instruction-following.
\uiot{} contains 7 billion parameters and is pre-trained from scratch on an extensive variety of multimodal data -- 1 billion image-text pairs, 1 trillion text tokens, 180 million video clips, 130 million interleaved image \& text, 3 million 3D assets, and 1 million agent trajectories. We further instruction-tune the model with a massive multimodal corpus by combining more than 120 datasets covering 220 tasks across vision, language, audio, and action.

Our pre-training and instruction tuning data, totaling over 600 terabytes, presents significant challenges for training due to its diversity and volume. 
To effectively facilitate self-supervised learning signals across multiple modalities, we develop a novel multimodal mixture of denoiser objective that combines denoising and generation across modalities. 
We also develop dynamic packing -- an efficient implementation that provides a 4x increase in training throughput to deal with highly variable sequences. 
To overcome the stability and scalability issues in training, we propose to apply key architectural changes, including 2D rotary embeddings, QK normalization, and scaled cosine attention mechanisms on the perceiver resampler. 
For instruction tuning, we ensure every task has a clear prompt, either using existing ones or crafting new ones. 
We also include open-ended
tasks and create synthetic tasks for less common modalities
to enhance task and instruction variety. 

We evaluate \uiot{} on over 35 datasets across the various modalities it supports. 
Our single model sets the new state of the art on the GRIT \cite{gupta2022grit} benchmark, which includes diverse tasks such as keypoint estimation and surface normal estimation.
On vision \& language tasks, it matches or outperforms the performance of many recently proposed VLMs that leverage pre-trained LLMs. 
On image generation, it outperforms the closest competitor \cite{tang2023codi} that leverages the pre-trained stable diffusion model \cite{SD15}, especially in terms of faithfulness as per the metrics defined in \cite{hu2023tifa}. 
It also shows effectiveness in video, natural language, audio, and embodied AI tasks, showcasing versatility despite its broad capability range.
Moreover, \uiot{} can follow free-form instructions, including novel ones. Figure~\ref{fig:teaser} offers a glimpse into how it handles various tasks. 
Further examples, along with the code and models, are accessible on our \href{https://unified-io-2.allenai.org/}{project website}.
\section{Related Work}
\label{sec:related}
Inspired by the success of language models as general-purpose text processing systems~\cite{flanv2,touvron2023llama,bubeck2023sparks}, there has been a recent wave of multimodal systems trying to achieve similar general-purpose capabilities with additional modalities. A common approach is to use a \textit{vision-encoder} to build features for input images and then an \textit{adapter} to map those features into embeddings that can be used as part of the input to an LLM. The network is then trained on paired image/language data to adapt the LLM to the visual features. These models can already perform some tasks zero-shot or with in-context examples~\cite{frozen,BLIP2,magma}, but generally a second stage of visual instruction tuning follows using instructions, visual inputs, and target text triples to increase zero-shot capabilities~\cite{llava,llava15,llama_adapter,dai2023instruct_blip,ye2023mplug_owl,zhu20223minigpt4,chen2023minigpt_v2}.

Building upon this design, many researchers have expanded the breadth of tasks these models can support.
This includes creating models that can do OCR~\cite{zhang2023llavar,bai2023qwen}, 
visual grounding~\cite{chen2023shikra,wang2023visionllm,zhang2023gpt4roi,peng2023kosmos2,you2023ferret,bai2023qwen,zang2023context_det}, image-text-retrieval~\cite{koh2023fromage}, additional languages~\cite{li2023m}, embodied AI tasks~\cite{brohan2023rt,padalkar2023rtx,mu2023embodiedgpt,reed2022generalist} or leverage other expert systems~\cite{llama_adapter2}. 
Other efforts have added new input modalities. This includes video inputs~\cite{Li2023VideoChatCV,Luo2023ValleyVA}, audio~\cite{huang2023audiogpt} or both~\cite{zhang2023video_llama}. 
PandaGPT~\cite{su2023pandagpt} and ImageBind-LLM~\cite{han2023imagebind} use the universal encoder ImageBind~\cite{girdhar2023imagebind} to encode many kinds of input modalities, and ChatBridge~\cite{zhao2023chatbridge} uses a similar universal encoder based on language. While these efforts are effective for understanding tasks, they do not allow complex multimodal generation and often exclude modalities long considered central to computer vision (\eg, ImageBind cannot support sparse annotation of images). 

Fewer works have considered multimodal generation. \uio{}~\cite{lu2022unified}, LaVIT~\cite{jin2023lavit}, OFA~\cite{wang2022OFA}, Emu~\cite{sun2023emu} and CM3Leon \cite{Yu2023ScalingAM} train models to generate tokens that a \vqgan{}~\cite{esser2021taming,van2017neural} can then decode into an image, while GILL~\cite{koh2023gill}, Kosmos-G~\cite{pan2023kosmosg} and SEED~\cite{ge2023seed} generate features that a diffusion model can use, and JAM~\cite{Aiello2023JointlyTL} fuses pre-trained language and image generation models. 
\uiot{} also uses a \vqgan{}, but supports text, image, and audio generation.

Overall, this shows a strong trend towards expanding the number of supported tasks and modalities. \uiot{} pushes this trend to its limit, including the capabilities of these prior works with few exceptions and the ability to generate outputs in more modalities.
Recently, CoDi~\cite{tang2023codi} also achieved similar any-to-any generation capabilities by using multiple independently trained diffusion models and aligning their embedding spaces. 
\uiot{} has stronger language abilities and can perform well on many more tasks. 

A notable feature of \uiot{} is that the model is trained from scratch instead of being initialized with a pre-trained LLM. Prior works~\cite{wang2022OFA,wang2022image,wang2021simvlm,liang2022high} following this approach are typically not designed to produce complex generations like free-form text responses, images or sounds, or follow text instructions.
Compared to recent general-purpose multimodals models~\cite{peng2023kosmos1,peng2023kosmos2, Yu2023ScalingAM}, \uiot{} has a significantly broader scope of tasks and outputs. Training from scratch means that the method can be reproduced without a costly preliminary stage of language model pre-training and is a more natural fit for how humans learn modalities simultaneously through their co-occurrences, not one at a time.

\begin{figure*}[t]
    \centering
    \includegraphics[width=\textwidth]{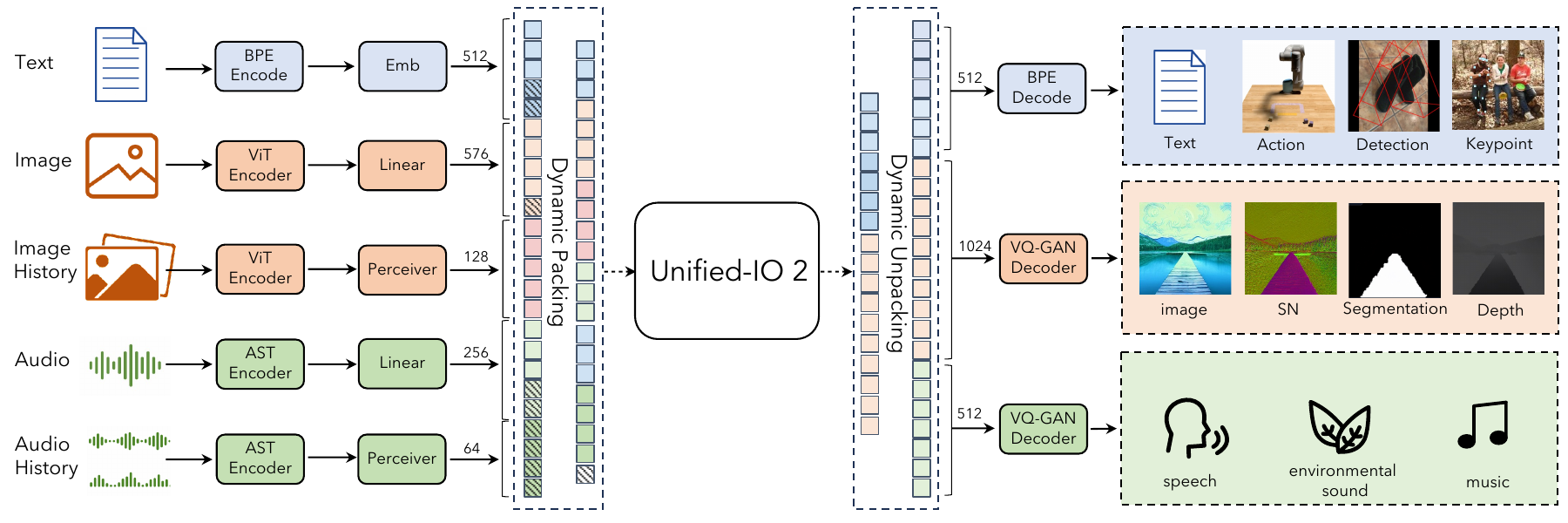}
    \captionof{figure}{\uiot{} architecture. Input text, images, audio, or image/audio history are encoded into sequences of embeddings which are concatenated and used as input to an encoder-decoder transformer model. The transformer outputs discrete tokens that can be decoded into text, an image, or an audio clip. 
    }
    \label{fig:model}
    \vspace{-2mm}
\end{figure*}

\section{Approach}

In this section, we discuss the unified task representation (\ref{sec:unified_task_representation}), the model architecture and techniques to stabilize training (\ref{sec:architecture}), the multimodal training objective (\ref{sec:objective}) and the efficiency optimizations (\ref{sec:implementation}) used in \uiot{}.

\subsection{Unified Task Representation}
\label{sec:unified_task_representation}
\uiot{} processes all modalities with a single, unified encoder-decoder transformer~\cite{vaswani2017attention}. 
This is achieved by encoding various inputs and outputs -- images, text, audio, action, boxes \etc, into sequences of tokens in a shared representation space.
Our encoding procedure follows the design of \uio{}~\cite{lu2022unified}, with several modifications to improve performance and new encoders and decoders for additional modalities. 
Figure~\ref{fig:model} shows an overview of the model. Details about how modalities are encoded are given below.

\boldheader{Text, Sparse Structures, and Action}
Text inputs and outputs are tokenized using the byte-pair encoding \cite{sennrich2015neural} from LLaMA \cite{touvron2023llama}, which we chose since it supports Unicode symbols and preserves whitespace. 
Sparse structures such as bounding boxes, keypoints, and camera poses are discretized and then encoded using 1000 special tokens added to the vocabulary \cite{chen2021pix2seq,lu2022unified}. 
Points are encoded with a sequence of two such tokens (one for $x$ and one for $y$), boxes are encoded with a sequence of four tokens (upper left and lower right corners), and 3D cuboids are represented with 12 tokens that encode the projected center, virtual depth, log-normalized box dimension, and continuous allocentric rotation~\cite{brazil2023omni3d}.
For embodied tasks, discrete robot actions \cite{brohan2023rt} are generated as text commands (\eg, ``move ahead'' to command the robot to move forward in navigation). Special tokens are used to encode the robot's state, such as its position and rotation. Details are in Appendix~\ref{supp:detail_task_representation}.

\boldheader{Images and Dense Structures}
Images are encoded with a pre-trained Vision Transformer (ViT) \cite{ilharco_gabriel_2021_5143773}. We concatenate the patch features from the second and second-to-last layers of the ViT to capture both low and high-level visual information. These features are passed through a linear layer to get embeddings that can be used as part of the input sequence for the transformer.
To generate images, we use VQ-GAN \cite{esser2021taming} to convert images into discrete tokens. These tokens are added to the vocabulary and then used as the target output sequence in order to generate an image.
For better image quality, we use a dense pre-trained VQ-GAN model with $8\times8$ patch size that encodes a $256 \times 256$ image into 1024 tokens with a codebook size of 16512.

Following \cite{lu2022unified}, we represent per-pixel labels, which include depth, surface normals, and binary segmentation masks, as RGB images that can be generated or encoded with our image generation and encoding abilities.
For segmentation, \uiot{} is trained to predict a binary mask given a class and bounding box. An entire image can be segmented by first doing detection, and then querying the model for a segmentation mask for each detected bounding box and class. 
See Appendix~\ref{supp:detail_task_representation} for details.

\boldheader{Audio} 
\uiot~encodes up to 4.08 seconds of audio into a spectrogram (See Appendix~\ref{supp:detail_task_representation} and Table~\ref{tab:input_representation}). 
The spectrogram is then encoded with a pre-trained Audio Spectrogram Transformer (AST) \cite{gong2021ast}, and the input embeddings are built by concatenating the second and second-to-last layer features from the AST and applying a linear layer just as with the image ViT. To generate audio, we use a ViT-VQGAN \cite{yu2021vector} to convert the audio into discrete tokens. Since there is no public codebase, we implement and train our own ViT-VQGAN with $8\times8$ patch size that encodes a $256 \times 128$ spectrogram into 512 tokens with a codebook size of 8196.

\boldheader{Image and Audio History}
We allow up to four additional images and audio segments to be given as input, which we refer to as the image or audio history. These elements are also encoded using the ViT or AST, but we then use a perceiver resampler \cite{Alayrac2022FlamingoAV}, see Table~\ref{tab:input_representation} for hyperparameters, to further compress the features into a smaller number of tokens (32 for images and 16 for audio). This approach greatly reduces the sequence length and allows the model to inspect an image or audio segment in a high level of detail while using elements in the history for context.
This history is used to encode previous video frames, previous audio segments, or reference images for tasks such as multi-view image reconstruction or image-conditioned image editing. Eight special tokens are added to the text vocabulary and used to reference the individual elements in these histories in the text input or output.

\begin{figure*}[t]
    \centering
    \includegraphics[width=1.0 \textwidth]{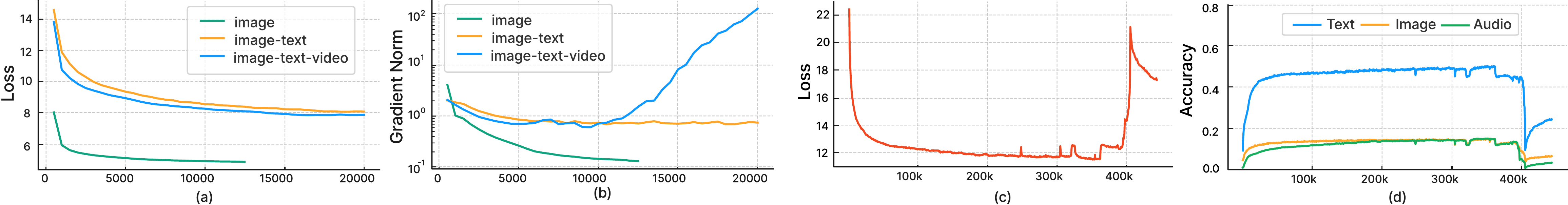}
    \vspace{-5mm}
    \captionof{figure}{\small \textbf{Left}: Training loss (a) and gradient norms (b) on different modality mixtures. \textbf{Right}: Training loss (c) and next token prediction accuracy (d) of \uioxxl on all modalities. Results were obtained before applying the proposed architectural improvements.
 }
    \label{fig:loss_explode}
    \vspace{-1mm}
\end{figure*}

\begin{figure}[t]
    \centering
    \includegraphics[width=0.78 \columnwidth]{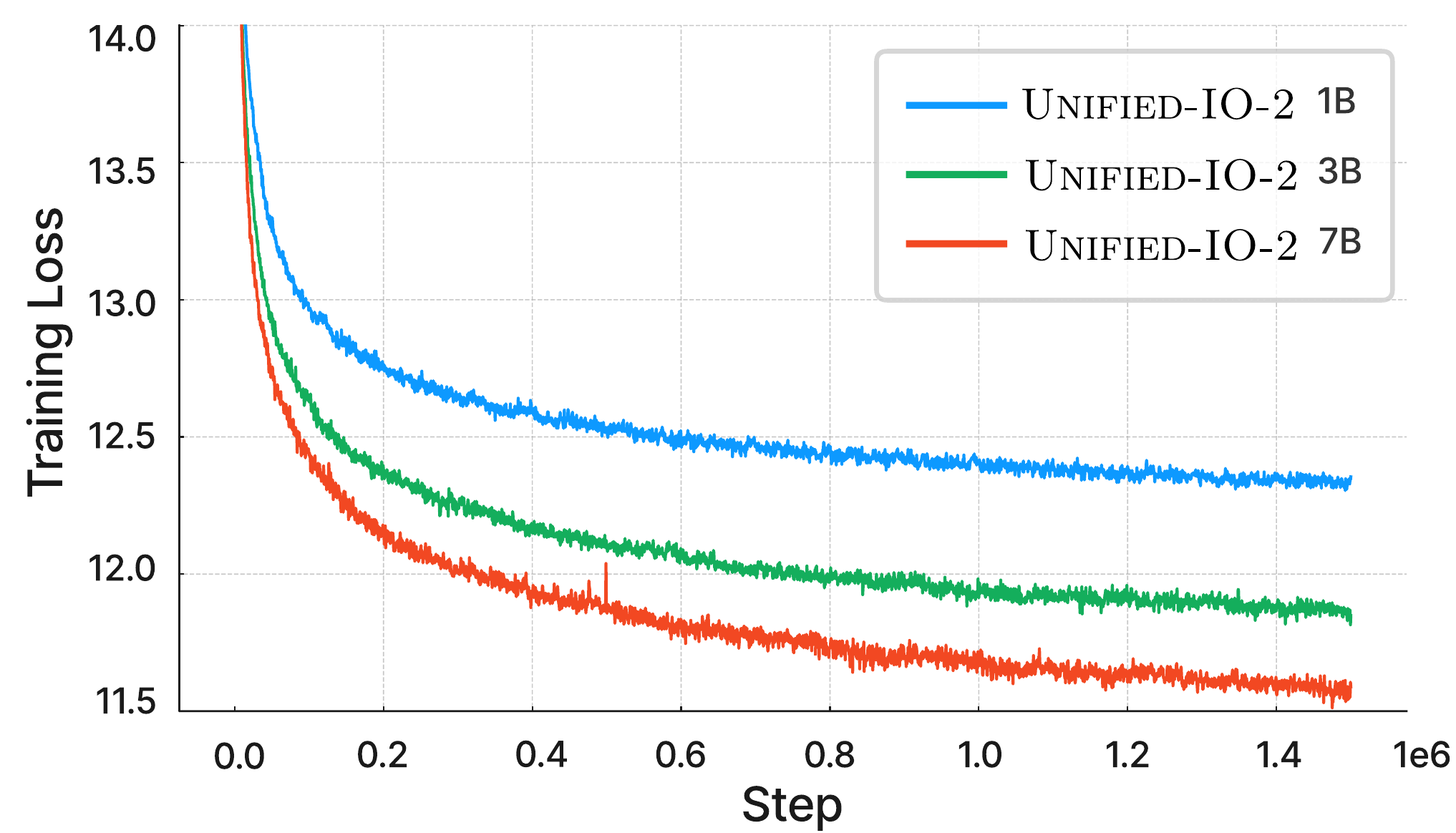}
    \vspace{-1mm}
    \captionof{figure}{\small Training loss curves for the three models, which are pretrained with dynamic packing and a batch size of 512.}
    \label{fig:training_loss}
    \vspace{-1mm}
\end{figure}

\subsection{Architecture}
\label{sec:architecture}

\uiot{} uses a transformer encoder-decoder architecture.
However, we observe that using a standard implementation following \uio{} leads to increasingly unstable training as we integrate additional modalities. 
As shown in Figure~\ref{fig:loss_explode} (a) and (b), training only on image generation (green curve) results in stable loss and gradient norm convergence.
Introducing a combination of image and text tasks (orange curve) slightly increases the gradient norm compared to a single modality, but remains stable.
However, the subsequent inclusion of the video modality (blue curve) leads to an unrestrained escalation in the gradient norm.
When an \texttt{XXL} version of this model is trained on all modalities, as shown in Figure~\ref{fig:loss_explode} (c) and (d), the loss explodes after 350k steps, and the next token prediction accuracy significantly drops at 400k steps. 
To address this, we include various architectural changes that significantly stabilize multimodal training.  

\boldheader{2D Rotary Embedding} Instead of relative positional embedding \cite{2020t5}, we apply rotary positional embeddings (RoPE) \cite{su2021roformer} at each transformer layer. For non-text modalities, we extend RoPE to two-dimensional positions: 
For any 2D indexes $(i, j)$, we split each of the query and key embeddings of the transformer attention heads in half and apply separate rotary embeddings constructed by each of the two coordinates to the halves, see Appendix~\ref{supp:rotary}. 

\boldheader{QK Normalization} We observe extremely large values in the multi-head attention logits when including image and audio modalities, which leads to attention weights becoming either 0 or 1 and contributes to training instability. To solve this, following \cite{dehghani2023scaling}, we apply LayerNorm~\cite{ba2016layer} to the queries and keys before the dot-product attention computation.

\boldheader{Scaled Cosine Attention} We use perceiver resampler \cite{jaegle2021perceiver} to compress each image frame and audio segment into a fixed number of tokens. We found that even with QK normalization, the attention logits in the perceiver can grow to extreme values.
Therefore, we apply more strict normalization in the perceiver by using scaled cosine attention \cite{liu2022swin}, which significantly stabilizes training. 

To avoid numerical instabilities, we also enable float32 attention logits. Jointly updating the pre-trained ViT and AST can also cause instabilities. Thus, we freeze the ViT and AST during pretraining and finetune them at the end of instruction tuning. Figure~\ref{fig:training_loss} shows that the pre-training loss for our model is stable despite the heterogeneity of input and output modalities.

\subsection{Training Objective}
\label{sec:objective}

A strong multimodal model has to be exposed to solving diverse sets of problems during pre-training. UL2 \cite{tay2022ul2} proposed the Mixture of Denoisers (MoD), a unified perspective to train LLMs, which combines the span corruption \cite{2020t5} and causal language modeling \cite{brown2020language} objectives. Motivated by this, we propose a generalized and unified perspective for multimodal pre-training.

\boldheader{Multimodal Mixture of Denoisers} MoD uses three paradigms: \texttt{[R]} -- standard span corruption, \texttt{[S]} -- causal language modeling, and \texttt{[X]} -- extreme span corruption. 
For text targets, we follow the UL2 paradigms. For image and audio targets, we define two analogous paradigms: \texttt{[R]} -- masked denoising where we randomly mask $x$\% of the input image or audio patch features and task the model to re-construct it and \texttt{[S]} -- where we ask the model to generate the target modality conditioned only on other input modalities.
During training, we prefix the input text with a modality token (\texttt{[Text]}, \texttt{[Image]}, or \texttt{[Audio]}) and a paradigm token (\texttt{[R]}, \texttt{[S]}, or \texttt{[X]}) to indicate the task.

\begin{figure}[t]
    \centering
    \includegraphics[width=1.0\columnwidth]{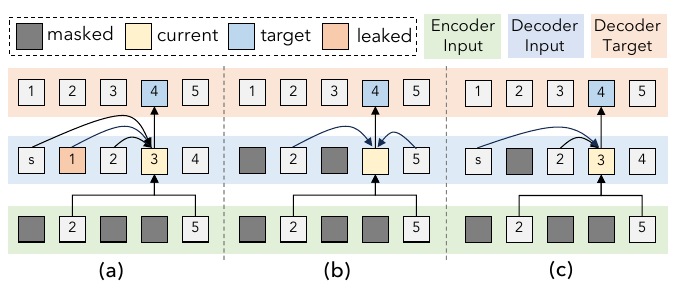}
    \captionof{figure}{\small Different training paradigms in masked image modeling: (a) autoregressive, (b) mask auto-encoder,  (c) autoregressive with dynamic masking. Our proposed paradigms can maintain causal generation while avoiding information leaks in the decoder.}
    \vspace{-3mm}
    \label{fig:dynamic_mask}
\end{figure}

\begin{figure*}[t]
    \centering
    \includegraphics[width=\textwidth]{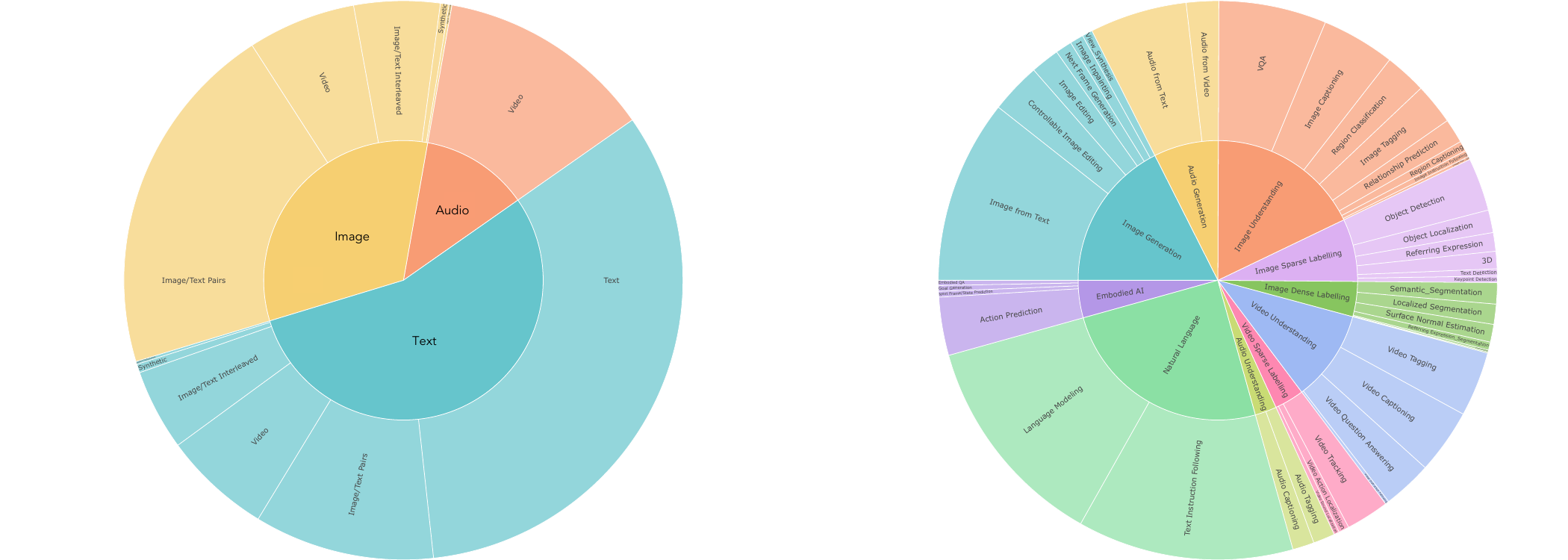}
    \captionof{figure}{\small Distribution of pre-training and instruction tuning data. Segments proportional to sampling rates. The inner section shows the target modality, and the outer section shows the data type. Please refer to Figure~\ref{fig:pretraining_sunburst} and Figure~\ref{tab:instruction-tuning-data} in the Appendix for particular datasets.}
    \vspace{-1mm}
    \label{fig:data_main}
\end{figure*}

\boldheader{Autoregressive with Dynamic Masking}
One problem with image and audio masked denoising in an autoregressive manner is an information leak on the decoder side; see Figure~\ref{fig:dynamic_mask} (a). 
The current decoder's input token (\texttt{3}) is conditioned on enocoder's information (\texttt{2}, \texttt{5}) and all previous tokens (\texttt{s} $\rightarrow$ \texttt{2}) to predict target (\texttt{4}). 
As a result, the predicted token will be conditioned on \texttt{1} even though it was masked in the encoder since it appears in the decoder, which will simplify the task and harm representation learning.
Simply masking the token in the decoder, as shown in Figure~\ref{fig:dynamic_mask} (b), avoids this information leakage but causes the generation and de-noising tasks to interfere with one another. For example, we found that joint training with generation (50\% MAE and 50\% causal modeling) significantly reduced image generation performance. Our solution is to mask the token in the decoder except when predicting that token, as shown in Figure~\ref{fig:dynamic_mask} (c), which does not interfere with causal prediction whilst mostly eliminating data leakage. 
For image and audio generation, we also use row, column, and conv-shaped masked sparse attention \cite{ramesh2021zero} in the decoder.

\subsection{Efficient Implementation}
\label{sec:implementation}

Training on heavily multimodal data results in highly variable sequence lengths for the transformer's inputs and outputs, both because modalities are often missing for individual examples and because the number of tokens used to encode particular modalities can vary from just a few tokens (for a sentence) to 1024 tokens (for an output image). To handle this efficiently, we use packing, a process where the tokens of multiple examples are packed into a single sequence, and the attentions are masked to prevent the transformer from cross-attending between examples.

Typically, packing is done during pre-processing, but it is challenging in our setup since our encoders and decoder do not always support it. Instead, we do packing right before and after the transformer encoder-decoder stage, which allows the modality encoders/decoder to run on the unpacked data. During training, we use a heuristic algorithm to re-arrange data being streamed to the model so that long examples are matched with short examples they can be packed with.
Packing optimization was also explored in~\citep{krell2021efficient}, but not in the streaming setup. Dynamic packing leads to an almost 4x increase in training throughput (Details in Appendix~\ref{supp:dynamic_packing}).

\begin{table}[t]
\setlength\tabcolsep{4pt}
\renewcommand{\arraystretch}{1.15}
\center
  \resizebox{1.0\columnwidth}{!}{
  \begin{tabular}
  {l c c c c c c }
\toprule
{Model} & {model dims} & {mlp dims} & {encoder lyr} & {decoder lyr} & {heads} & {Params}\\
\midrule
\uiotacronym$_\texttt{L}$ & 1024 & 2816 & 24 & 24 & 16 & 1.1B  \\
\uiotacronym$_\texttt{XL}$ & 2048 & 5120 & 24 & 24 & 16 & 3.2B  \\
\uiotacronym$_\texttt{XXL}$ & 3072 & 8192 & 24 & 24 & 24 & 6.8B \\
\bottomrule
\end{tabular}}
\vspace{-1mm}
\caption{\small{Size variant of \uiot.}}
\vspace{-2mm}
\label{tab:model_size}
\end{table}

\subsection{Optimizer}
We use Adafactor \citep{shazeer2018adafactor} as our optimizer with a linear warm-up for the first 5,000 steps and a learning rate decay of $1/\sqrt{k}$. We train with $\beta_1 = 0.9$ and $\beta_2 = 1.0-k^{-0.8}$, where $k$ is the step number. We use global norm gradient clipping with a threshold of 1.0 and find that this is crucial to stabilized training. Table~\ref{tab:model_size} gives the details of our different models.
For all models, we train $3.0$M steps -- $1.5$M for pre-training and 1.5M for instruction tuning, respectively. 
More details in Appendix~\ref{supp:full_model_details}.

\section{Multimodal Data}
\label{sec:model_training}

One critical difference between \uiot~and prior work is that we train the model with a diverse set of multimodal data from scratch. This requires curating high-quality, open-source multimodal data for both pre-training (\ref{sec:training_data}) and instruction tuning (\ref{sec:tuning_data}).

\begin{figure*}[t]
    \centering
    \includegraphics[width=\textwidth]{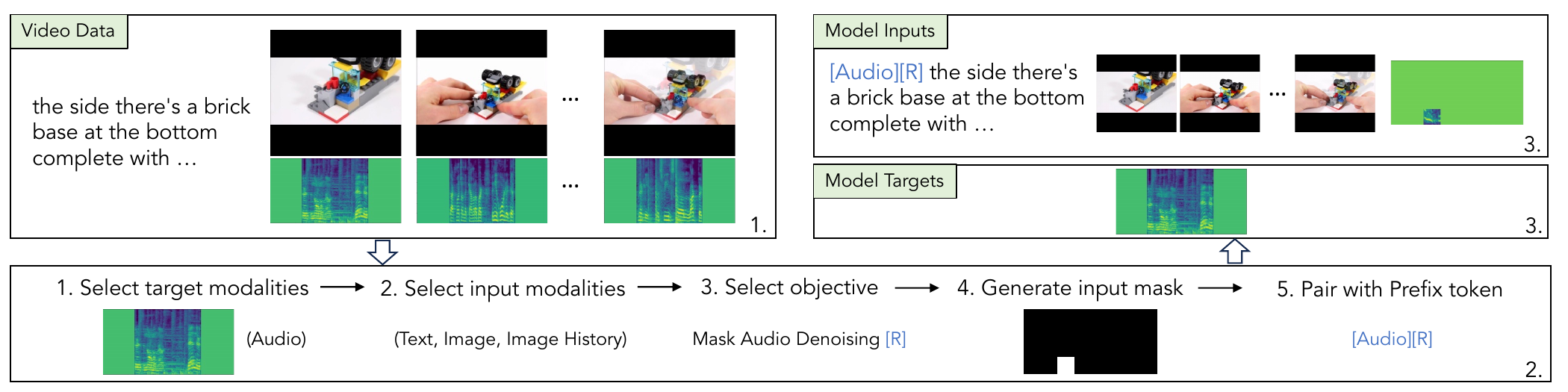}
    \vspace{-5mm}
    \captionof{figure}{\small Construction of training samples from video data for the model's input and target. Given the video, we first extract the video frames and the corresponding audio spectrograms and transcript. Then, the data pass through a random selection process to determine the target modality, input modalities, training objective, input mask \etc. The model's final input and target are shown in the top right.}
    \label{fig:pretraining_sample}
    \vspace{-1mm}
\end{figure*}

\subsection{Pre-training Data}
\label{sec:training_data}

Our pre-training data comes from various sources and covers many modalities. We provide a high-level overview and details in Appendix~\ref{supp:pretraining}.

\boldheader{NLP [33\%]} We use the publicly available datasets that were employed to train MPT-7B \cite{MosaicML2023Introducing}. This dataset emphasizes English natural language text but also contains code and markdown. It includes text from the RedPajama dataset~\cite{together2023redpajama}, C4~\cite{habernal2016c4corpus}, Wikipedia, and stack overflow. We follow the proportion suggested by \cite{MosaicML2023Introducing} and remove multi-lingual and scientific data. 

\boldheader{Image \& Text [40\%]} Text and image paired data comes from LAION-400M~\cite{schuhmann2021laion400m}, CC3M~\cite{sharma2018cc3m}, CC12M~\cite{changpinyo2021cc12m}, and RedCaps~\cite{desai2021redcaps}. 
To help train the image-history modality, we also use the interleaved image/text data from OBELICS~\cite{laurenccon2023obelics}. We use the last image as the image input and the remaining images as the image history. Special tokens are used to mark where those images occur in the text.

\boldheader{Video \& Audio [25\%]} Video provides strong self-supervisory signals with high correlations between audio and visual channels. We sample audio and video data from various public datasets including YT-Temporal-1B~\cite{Zellers_2022_CVPR}, ACAV100M~\cite{lee2021acav100m}, AudioSet~\cite{gemmeke2017audio}, WebVid-10M~\cite{bain2021webvid}, HD-VILA-10M~\cite{xue2022hdvila} and Ego4D~\cite{Grauman_2022_CVPR}.

\boldheader{3D \& Embodiment [1\%]} For self-supervised 3D and embodiment pre-training, we use CroCo~\cite{weinzaepfel2022croco} for cross-view generation and denoising; Objaverse~\cite{deitke2023objaverse} for view synthesis; and random trajectories in ProcTHOR~\cite{deitke2022} and Habitat~\cite{savva2019habitat} for the next action and frame predictions.

\boldheader{Augmentation [1\%]}
While there is a lot of unsupervised data on the web for images, text, video, and audio, options are much more limited for dense and sparse annotations. 
We propose to solve this through large-scale data augmentation. 
We consider two types of data augmentation: 1. Automatically generated segmentation data from SAM~\cite{Kirillov_2023_ICCV} to train the model to segment an object given a point or bounding box.
2. Synthetic patch-detection data which tasks the model to list the bounding boxes of synthetically added shapes in an image. We additionally
train the model to output the total number of patches in the
image to pre-train its counting abilities. 

\boldheader{Training Sample Construction}
During pre-training, most of our data contains various modalities without a supervised target. In these cases, we randomly pick one of the modalities present to be the target output. Then, we either remove that modality from the example or replace it with a corrupted version. Other modalities that might be present in the example are randomly kept or masked to force the model to make predictions using whatever information is left.
Figure \ref{fig:pretraining_sample} shows an example when using a video that contains a sequence of image frames, the corresponding audio, and a text transcript. The pre-training sample is constructed by following the procedure: 1. select the target modality; 2. select which other input modalities to keep; 3. select the objective; 4. generate the random input mask depending on the task of denoising or generation; 5. add a prefix token indicating the task.

\begin{figure*}[t]
    \centering
    \includegraphics[width=\textwidth]{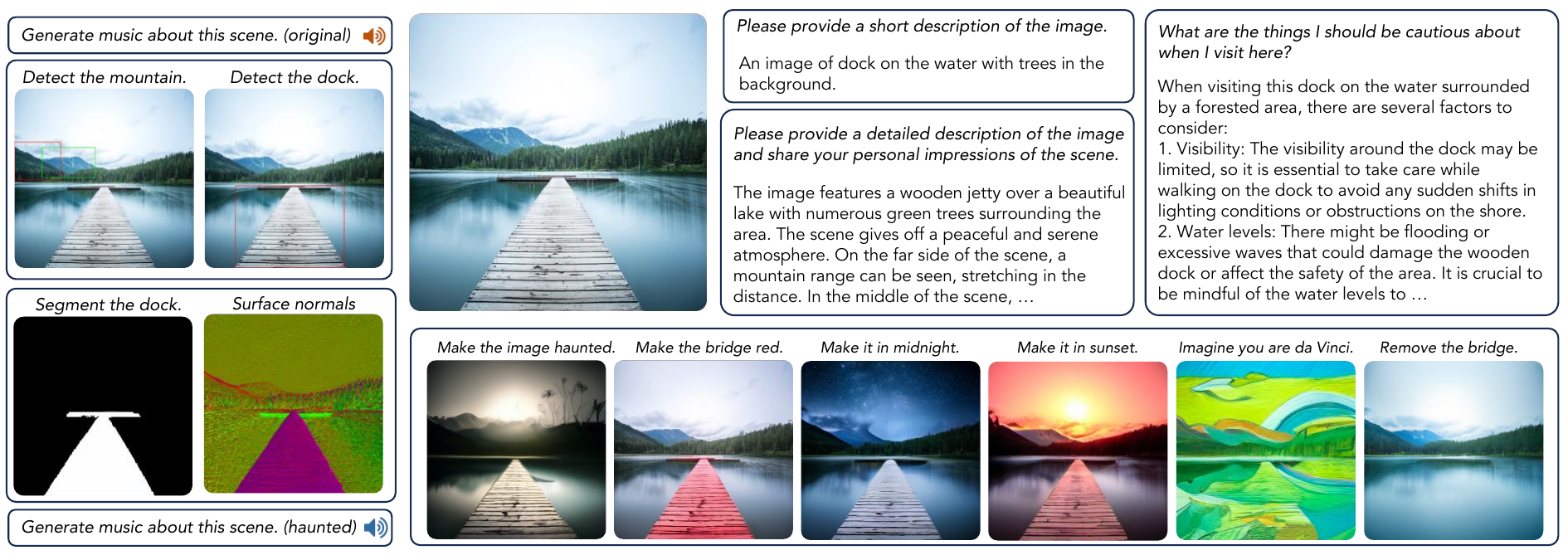}
    \vspace{-5mm}
    \captionof{figure}{\small Our single model can perform a multitude of multimodal tasks: captioning the image, following free-form instructions, image editing, object detection, semantic segmentation, surface normal, and image-based audio generation, \etc Here, we show the outputs of our model for a variety of prompts. Click \href{https://ai2-prior-uio.s3.us-west-2.amazonaws.com/public/samples/bridge-1.wav}{\textcolor{red}{\faVolumeUp}} and \href{https://ai2-prior-uio.s3.us-west-2.amazonaws.com/public/samples/bridge-2.wav}{\textcolor{blue}{\faVolumeUp}} for the corresponding audio samples. }
    \label{fig:instruction_tuning_sample}
    \vspace{-1mm}
\end{figure*}

\subsection{Instruction Tuning Data}
\label{sec:tuning_data}
Multimodal instruction tuning is the key process to equip the model with diverse skills and capabilities across various modalities and even adapt to new and unique instructions.
We construct the multimodal instruction tuning dataset by combining a wide range of supervised datasets and tasks. 
We ensure every task has a clear prompt, either using existing ones or writing new ones. We also include open-ended tasks and create synthetic tasks for less common modalities to enhance task and instruction variety.
Our mixture includes 220 tasks drawn from over 120 external datasets. We provide a high-level overview and examples here and leave details in Appendix \ref{supp:instruction-tuning}.

\boldheader{Natural Language [25.0\%]}
For natural language, we use the mixture from FlanV2~\cite{flanv2} and various other instruction following datasets~\cite{dolly,peng2023instruction}. In addition, we continue pre-training on our unsupervised NLP mixture to help prevent the model from forgetting information learned from pre-training during the extensive instruction tuning stage.

\boldheader{Image Generation [17.6\%]}
For text-to-image generation, we use the same image \& text pairs we used during pre-training. We also include data from \cite{open_images, coco, visual_genome} that provide better caption quality. 
We additionally train the model to generate images through view synthesis~\cite{weinzaepfel2022croco,deitke2023objaverse}, image editing~\cite{brooks2023instructpix2pix,zhang2023magicbrush},  segmentation-based image generation~\cite{lu2022unified} and inpainting \cite{lu2022unified}. 

\boldheader{Audio Generation [7.5\%]}
This includes text-to-audio datasets with audio in the wild~\cite{audiocaps, drossos2020clotho, martin2021diversity}, music~\cite{agostinelli2023musiclm}, and human speech~\cite{ljspeech}. We also add pre-training data with the task of predicting the next audio clip in a video.
More specifically, we divide the audio into segments and then generate one of them given both the text and previous segments as input.

\boldheader{Image Understanding [17.8\%]}
We include various data sources from visual question answering \cite{antol2015vqa}, image tagging \cite{deng2009imagenet}, region classification \cite{visual_genome}, and datasets with open-ended chat-like responses \cite{llava,zhang2023llavar}. We also include the multimodal instruction tuning datasets M$^3$IT~\cite{li2023m} and MIMIC-IT~\cite{li2023mimic}.

\boldheader{Video Understanding [10.6\%]}
We include data sources from video captioning \cite{wang2019vatex, xu2016msr}, video tagging \cite{soomro2012ucf101, li2022uniformerv2, Damen2022rescaling}, and video question answering \cite{xu2017video, wu2021star}.
We also use examples from M$^3$IT~\cite{li2023m} and MIMIC-IT~\cite{li2023mimic} for video instruction following. 

\boldheader{Audio Understanding [10.6\%]}
We include data sources from audio tagging~\cite{gemmeke2017audio, chen2020vggsound}, and audio captioning~\cite{audiocaps,drossos2020clotho}. We also include data from video action classification~\cite{arandjelovic2017look} with audio in the dataset. 

\boldheader{Image Sparse Labelling [7.25\%]}
These tasks require outputting sparse coordinates based on an input image. We mainly consider object detection \cite{coco}, referring expression \cite{referitgame}, 3D detection \cite{brazil2023omni3d}, camera pose prediction \cite{deitke2023objaverse}, text detection \cite{veit2016cocotext} and human keypoints \cite{coco}.

\boldheader{Image Dense Labelling [4.06\%]}
We do several image labeling tasks, including surface normal estimation~\cite{huang2019framenet, yao2020blendedmvs}, depth estimation~\cite{nyu_depth}, and optical flow~\cite{dosovitskiy2015flyingchairs, butler2012mpisintel}. 
We also train our models on various segmentation tasks, including semantic segmentation, localization segmentation, and referring expression segmentation.

\boldheader{Video Sparse Labelling [3.42\%]}
We do video detection \cite{real2017youtube}, single object tracking \cite{fan2021lasot, huang2019got} and video action localization \cite{gu2018ava}.

\boldheader{Embodied AI [4.33\%]}
For VIMA-Bench \cite{jiang2023vima}, we use the image input as the initial observation of the environment and the image history for the images or videos in the prompt. 
We add large-scale manipulation datasets \cite{lynch2023interactive, walke2023bridgedata, gupta2019relay} with continuous control in both simulated and real-world environments. We also train on the PointNav task from Habitat Gibson scenes.

The distribution of the instruction tuning data is in Figure~\ref{fig:data_main}. 
Overall, our instruction tuning mixture is composed of 60\% prompting data, meaning supervised datasets combined with prompts. To avoid catastrophic forgetting, 30\% of the data is carried over from pre-training. Additionally, 6\% is task augmentation data we build by constructing novel tasks using existing data sources, which enhances existing tasks and increases task diversity. The remaining 4\% consists of free-form text to enable chat-like responses.
\section{Experiments}
In this section, we evaluate our pre-trained and instruction-tuned models on a broad range of tasks that require parsing and producing all modalities: images, video, audio, text, and actions.
\textbf{We do not perform task-specific finetuning in any experiments.}
Details about experimental setups, additional result details, results on natural language tasks, and additional studies for \uiot{}'s instruction capabilities are in Appendix~\ref{supp:results}. 

\subsection{Pre-training Evaluation}
\label{exp:pretrain}

We demonstrate the effectiveness of our pre-training by evaluating \uiot{} on commonsense natural language inference (HellaSwag~\cite{zellers2019hellaswag}), text-to-image generation (TIFA~\cite{hu2023tifa}) and text-to-audio generation (AudioCaps~\cite{audiocaps}). We also assess spatial and temporal understanding on SEED-Bench~\cite{li2023seed}, a benchmark for comprehensively evaluating perception and reasoning on image and video modalities. Table~\ref{tab:pretrain-results} shows that \uiot{} achieves comparable or even better performance on both generation and comprehension tasks compared to the task-specific specialist~\cite{SD15} or the universal multimodal model~\cite{Awadalla2023OpenFlamingoAO}. 

Despite extensive multitasking, the results on HellaSwag suggest that \uiot{} has language modeling capabilities between typical 3B and 7B language models. This may be due to that the model sees far fewer tokens compared to language-based LLMs -- approximately 250 billion tokens in total. Qualitative results of pre-training are in Appendix~\ref{supp:pretrain_visualization}.

\begin{table}[t]
\tablefont
\setlength\tabcolsep{3pt}
\renewcommand{\arraystretch}{1.15}
\center
\resizebox{\columnwidth}{!}{
\begin{tabular}{l  c c c c c}
\toprule
Method  & HellaSwag$\uparrow$ & TIFA$\uparrow$ & SEED-S$\uparrow$  & SEED-T$\uparrow$ & AudioCaps$\downarrow$ \\

\midrule
LLaMA-7B \cite{touvron2023llama} & \textbf{76.1} & - & - & - & - \\
OpenLLaMa-3Bv2 \cite{openlm2023openllama} & 52.1 & - & - & - & - \\
SD v1.5 \cite{SD15} & - & 78.4 & - & - & - \\
OpenFlamingo-7B \cite{Awadalla2023OpenFlamingoAO} & - & - & 34.5 & 33.1 & - \\


\midrule    
\uiotacronym$_\texttt{L}$   &  38.3       &     70.2 & 37.2 & 32.2  & 3.08
\\
\uiotacronym$_\texttt{XL}$   &  47.6       & 77.2  & 40.9  & 34.0 & 3.10 \\
\uiotacronym$_\texttt{XXL}$  &  54.3       & \textbf{78.7}  &  \textbf{40.7}  &  \textbf{35.0} & \textbf{3.02} \\
\bottomrule
\end{tabular}}
\vspace{-2mm}
\caption{\small Zero-shot performance on commonsense sentence completion (HellaSwag \cite{zellers2019hellaswag}), text-to-image generation (TIFA~\cite{hu2023tifa}), spatial and temporal comprehension (Seed-Bench~\cite{li2023seed}), and text-to-audio generation (AudioCaps~\cite{audiocaps}).}
\vspace{-1mm}
\label{tab:pretrain-results}
\end{table}

\begin{table}[tb]
\tablefont
\setlength\tabcolsep{5pt}
\renewcommand{\arraystretch}{1.15}
\center
\resizebox{\columnwidth}{!}{
\begin{tabular}{c l c c c c c c c c }
\toprule
& Method & Cat. & Loc. & Vqa & Ref. & Seg. & KP & Norm. & All \\
 \midrule
\multirow{3}{*}{\small \rotatebox{90}{\footnotesize Ablation}}& 
\uiolarge & 70.1 & 66.1 & 67.6 & 66.6 & 53.8 & 56.8 & 44.5 & 60.8 \\ 
& \uioxl  & 74.2 & 69.1 & 69.0 & 71.9 & 57.3 & 68.2 & \textbf{46.7} & 65.2 \\
& \uioxxl & \textbf{74.9} & \textbf{70.3} & \textbf{71.3} & \textbf{75.5} & \textbf{58.2} & \textbf{72.8} & 45.2 & \textbf{66.9} \\
 \midrule
\multirow{3}{*}{\small \rotatebox{90}{\footnotesize Test}} 
 & GPV-2 \cite{Kamath2022WeblySC} & 55.1 & 53.6 & 63.2 & 52.1 & - & - & - & - \\
& UIO$_{\texttt{XL}}$ \cite{lu2022unified}& 60.8 & 67.1 & \textbf{74.5} & \textbf{78.9} & 56.5 & 67.7 & 44.3 & 64.3 \\
& \uioxxl & \textbf{75.2} & \textbf{70.2} & 71.1 & 75.5 & \textbf{58.8} & \textbf{73.2} & \textbf{44.7} & \textbf{67.0} \\
 \bottomrule
\end{tabular}}
\vspace{-2mm}
\caption{\small Results on the GRIT ablation and test sets~\cite{gupta2022grit}.}
\vspace{-2mm}
\label{tab:grit_results}
\end{table}

\subsection{GRIT Results}
\label{exp:grit}

We evaluate on the General Robust Image Task (GRIT) Benchmark~\cite{gupta2022grit}, which includes seven tasks: categorization, localization, VQA, referring expression, instance segmentation, keypoint, and surface normal estimation. Completing all 7 tasks requires understanding image, text, and sparse inputs and generating text, sparse, and dense outputs. Although this is a subset of the modalities \uiot{} supports, we evaluate on GRIT because it provides a standardized and comprehensive benchmark on this set of capabilities.
See Appendix \ref{supp:grit} for additional inference details on GRIT.

Results are shown in Table~\ref{tab:grit_results}. Overall, \uiot{} is state-of-the-art on GRIT, surpassing the previous best model, \uio{}, by 2.7 points. On individual tasks, we can observe gains in localization (3 points), categorization (14 points), segmentation (2 points), and keypoint (5 points). On VQA, our GRIT evaluations show \uiot{} is better on same-source (84.6 \vs 81.2) questions, suggesting the gap is due to reduced performance on the new-source questions that were constructed from Visual Genome; see Appendix \ref{supp:grit} for additional discussion.
Despite being slightly behind \uio{}, \uiot{} still obtains strong referring expression scores that compare favorably to prior work on generalist multimodal models, see Table~\ref{tab:results}.
Surpassing \uio{} while also supporting much higher quality image and text generation, along with many more tasks and modalities, illustrates the impressive multi-tasking capabilities of our model. \uiot{} even maintains better overall performance with the 3-billion parameter model (65.2 \vs 64.5), which is roughly equal in size to \uio{}.
Ablation results show average performance, and all individual tasks improve with model size, showing that \uiot{} benefits from scale.

\subsection{Generation Results}
\label{exp:generation}
\begin{table}[t]
\tablefont
\setlength\tabcolsep{5pt}
\renewcommand{\arraystretch}{1.15}
    \centering
      \resizebox{\columnwidth}{!}{
    \begin{tabular}{l c c c c c c }
        \toprule
        \multirow{3}{1.3cm}{Method} & \multicolumn{2}{c}{\textbf{Image}} & \multicolumn{3}{c}{\textbf{Audio}} & \textbf{Action} \\
        \cmidrule(r){2-3}
        \cmidrule(r){4-6}
        \cmidrule(r){7-7}

                                            & FID$\downarrow$   & TIFA$\uparrow$ & FAD$\downarrow$  & IS$\uparrow$   & KL$\downarrow$ & Succ.$\uparrow$  \\
        \midrule
        minDALL-E \cite{mindalle}          &   -  & 79.4 &    - & - &  - & -     \\
        SD-1.5 \cite{SD15}                  &  -      & 78.4 &   -   &  -  &  -  & - \\
        AudioLDM-L \cite{liu2023audioldm}   &   -   &  -   & 1.96 & 8.13 & 1.59 & - \\
        AudioGen \cite{kreuk2023audiogen}   &    -   &  -    & 3.13 &  -    & 2.09 & - \\
        DiffSound \cite{yang2023diffsound}  &    -  &  -   & 7.75 & 4.01 & 2.52 & - \\
        VIMA~\cite{jiang2023vima} & - & - & - & - & - &  72.6 \\
        VIMA-IMG~\cite{jiang2023vima} & - & - & - & - & - & 42.5 \\
        \midrule
        CoDi \cite{tang2023codi}            &  11.26 & 71.6 & 1.80 & 8.77 & 1.40 & -  \\
        Emu \cite{sun2023emu}    &  11.66 & 65.5 &  -  &  -  &  -  & - \\
        \midrule
        \uiotacronym$_\texttt{L}$           & 16.68 & 74.3 & 2.82 & 5.37 & 1.93 &  50.2 \\
        \uiotacronym$_\texttt{XL}$          & 14.11 & 80.0 & 2.59 & 5.11 & 1.74 & 54.2 \\
        \uiotacronym$_\texttt{XXL}$         & 13.39 & 81.3 & 2.64 & 5.89 & 1.80 &  56.3   \\

        \bottomrule
    \end{tabular}}
    \vspace{-1mm}
    \caption{\small Results on text-to-image generation (MS COCO~\cite{coco} and TIFA~\cite{hu2023tifa}), text-to-audio generation (AudioCaps~\cite{audiocaps}) and action generation (VIMA-Bench~\cite{jiang2023vima}).}
    \label{tab:img_audio_gen}
    \vspace{-2mm}
\end{table}
\newcommand{\band}{\rowcolor{gray!10}}

\newcolumntype{C}[1]{>{\centering\arraybackslash}p{#1}}
\begin{table*}[t!]
\tablefont
\setlength\tabcolsep{2pt}
\renewcommand{\arraystretch}{1.15}
\centering
\resizebox{\textwidth}{!}{

  \begin{tabular}{l c c c c c c c c c | c c c }
\toprule
Method          & VQA$^\text{v2}$ & OKVQA  & SQA & SQA$^\text{I}$ & Tally-QA  & RefCOCO & RefCOCO+ & RefCOCO-g & COCO-Cap. & POPE & SEED & MMB \\
\midrule
InstructBLIP (8.2B)     & -   & -  & - & 79.5 & 68.2$^\dagger$ & - & - & - & 102.2 & -  & 53.4 & 36 \\
Shikra (7.2B)           & 77.4  & 47.2     & -   & -  & - & 87.0 & 81.6 & 82.3 & 117.5 & 84.7  & -   & 58.8 \\
Ferret (7.2B)          & -  & - & -& -   & - & 87.5 & 80.8 & 83.9 & - & 85.8  & -   & - \\
Qwen-VL (9.6B)          & 78.8  & 58.6  & - & 67.1$^*$ &  - & 89.4  &  83.1 & 85.6  & 131.9 & - &  & 38.2 \\
mPLUG-Owl2 (8.2B)       & 79.4  & 57.7    & - & 68.7$^*$ & - & - & - & - & 137.3  & 86.2   &  57.8 & 64.5 \\
LLaVa-1.5  (7.2B)      & 78.5  & -  & -  &  66.8$^*$ & - & - & - & - & - & 85.9 & 58.6 & 64.3 \\
LLaVa-1.5 (13B)      & 80.0  & - & - & 71.6$^*$ & 72.4$^\dagger$ & - & - & - & -  & 85.9 & 61.6 & 67.7 \\
\midrule
Single Task SoTA & 86.0~\cite{chen2023pali}  &  66.8~\cite{hu2023visual}  & 90.9~\cite{llava} & 90.7~\cite{dai2023instruct_blip} & 82.4~\cite{hu2023visual} & 92.64~\cite{UNINEXT} & 88.77~\cite{wang2023one} & 89.22~\cite{wang2023one} &  149.1~\cite{chen2023pali} & - & - & - \\
\midrule
\uiotacronym$_\texttt{L}$ (1.1B) & 75.3  & 50.2 & 81.6 & 78.6 & 69.1 & 84.1 & 71.7 & 79.0$^\diamondsuit$ & 128.2 & 77.8 & 51.1 & 62.1 \\
\uiotacronym$_\texttt{XL}$ (3.2B) & 78.1  & 53.7 & 88.8 & 87.4 & 72.2 &  88.2 & 79.8 & 84.0$^\diamondsuit$ & 130.3 & 87.2 & 60.2 & 68.1 \\
\uiotacronym$_\texttt{XXL}$ (6.8B) & 79.4  & 55.5   & 88.7 & 86.2 & 75.9  & 90.7 & 83.1  & 86.6$^\diamondsuit$ & 125.4 & 87.7 & 61.8 & 71.5 \\
\bottomrule
\end{tabular}
}

\vspace{-1mm}
\caption{Vision-language results on nine tasks~\cite{balanced_vqa_v2,Marino2019OKVQAAV,lu2022learn,acharya2019tallyqa,mao2016generation, Chen2015MicrosoftCC,referitgame,yu2016modeling,nagaraja2016modeling} and three evaluation-only benchmarks~\cite{li2023seed,Liu2023MMBenchIY,Li2023EvaluatingOH}. Results marked with $^*$ are zero-shot and $^\dagger$ are evaluated with the open-source releases, and $^\diamondsuit$ indicates that our RefCOCO-g results are on the Google split rather than the UMD split.
}
\vspace{-2mm}
\label{tab:results}
\end{table*}


Table~\ref{tab:img_audio_gen} shows results on tasks that require generating image, audio, and action outputs.
We evaluate using TIFA~\cite{hu2023tifa}, which measures faithfulness to the prompt using VQA models and has been shown to correlate well with human judgments, and FID~\cite{NIPS2017_8a1d6947} on MS COCO~\cite{coco}. On TIFA, we find that \uiot{} scores close to minDALL-E \cite{mindalle}, and about 10 points ahead of other generalist models such as CoDi~\cite{tang2023codi} and Emu~\cite{sun2023emu}. We attribute this strong image generation ability to extensive pre-training and the use of a fine-grained VQ-GAN. We include examples of our generation results from the TIFA benchmark in the Appendix \ref{supp:image_gen}. \uiot{}'s FID scores are slightly higher than the compared models, although we note that qualitatively the generated images are still very smooth and detailed.

For text-to-audio generation, we evaluate on the AudioCaps~\cite{audiocaps} test set. AudioCaps
consists of 10-second audio clips, while our model can generate 4.08-second audio at a time, so we cannot do a direct evaluation on this benchmark. Instead, we generate an audio segment based on the text description and previous audio segments as additional input; see Appendix~\ref{supp:audio_gen} for more details.
While this is not a directly comparable setup to related work, it still gives a reasonable quantitative measure of our audio generation abilities.
\uiot{} scores higher then specialist models except the recent latent diffusion model~\cite{liu2023audioldm}, which shows it's competitive audio generation ability.

For action, we evaluate using VIMA-Bench~\cite{jiang2023vima}, a robot manipulation benchmark containing 17 tasks with text-image interleaved prompts. Since VIMA's action space is action primitives, \uiot{} directly predicts all actions at once given the initial observation and multimodal prompt. We report the average success rate for 4-level evaluation protocol~\cite{jiang2023vima} and compare with the original casual VIMA policy with object-centric inputs, as well as VIMA-IMG, a Gato~\cite{reed2022generalist}-like policy with image inputs like ours.

\subsection{Vision Language Results}
\label{exp:vision_langauge}

We evaluate vision language performance and compare it against other vision/language generalist models, \ie, models that are also designed to perform many tasks and can follow instructions.
Results on a collection of 12 vision/language benchmarks are shown in Table~\ref{tab:results}. SoTA results from specialist models are shown for reference.

\uiot{} achieves strong results on VQA, only passed by the much larger 13B LLaVa model~\cite{llava15} on VQA v2~\cite{balanced_vqa_v2}, and ahead of all other generalist models on ScienceQA~\cite{lu2022learn} and TallyQA~\cite{acharya2019tallyqa}. OK-VQA~\cite{Marino2019OKVQAAV} is the exception. We hypothesize that because it requires external knowledge, extensive language pre-training is important for this task, and therefore our reduced performance is since \uiot{} was not pre-trained as extensively on text as the dedicated language models used by Qwen-VL~\cite{bai2023qwen} and mPLUG-Owl2~\cite{ye2023mplug_owl2}.

\begin{table}[t]
\tablefont
\setlength\tabcolsep{3pt}
\renewcommand{\arraystretch}{1.15}
\centering
\resizebox{\columnwidth}{!}{
\begin{tabular}{l c c c c c c c c c c }
\toprule
\multicolumn{1}{l}{} & \multicolumn{7}{c}{\textbf{Video}} & \multicolumn{3}{c}{\textbf{Audio}} \\
\cmidrule(r){2-8}
\cmidrule(r){9-11}
Method & \rotatebox{90}{\raisebox{-0.5pt}{Kinetics-400~\cite{kay2017kinetics}}} & \rotatebox{90}{\raisebox{-0.5pt}{VATEXCaption~\cite{wang2019vatex}}} & \rotatebox{90}{\raisebox{-0.5pt}{MSR-VTT~\cite{xu2016msr}}} & \rotatebox{90}{\raisebox{-0.5pt}{MSRVTT-QA~\cite{xu2017video}}} & \rotatebox{90}{\raisebox{-0.5pt}{MSVD-QA~\cite{xu2017video}}} & \rotatebox{90}{\raisebox{-0.5pt}{STAR~\cite{wu2021star}}}  & \rotatebox{90}{\raisebox{-0.5pt}{SEED-T~\cite{li2023seed}}} & \rotatebox{90}{\raisebox{-0.5pt}{VGG-Sound~\cite{chen2020vggsound}}} & \rotatebox{90}{\raisebox{-0.5pt}{AudioCaps~\cite{audiocaps}}} 
& \rotatebox{90}{\raisebox{-0.5pt}{Kinetics-Sounds~\cite{arandjelovic2017look}}} 
\\
\midrule
MBT~\cite{Nagrani2021AttentionBF} & - & - & - & - & - & - & - & 52.3 & - & 85.0 \\
CoDi~\cite{tang2023codi} & - & - & 74.4 & - & - & - & - & - & 78.9 & - \\ 
ImageBind~\cite{han2023imagebind}$^*$ & 50.0 & - & - & - & - & - & - & 27.8 & - & - \\
BLIP-2~\cite{BLIP2}$^*$ & - & - & - & 9.2 & 18.3 & - & 36.7 & - & - & - \\
InstructBLIP~\cite{dai2023instruct_blip}$^*$ & - & - & - & 22.1 & 41.8 & - & 38.3 & - & - & - \\
Emu~\cite{sun2023emu}$^{**}$ & - & - & - & 24.1 & 39.8 & - & - & - & - & - \\
Flamingo-9B~\cite{Alayrac2022FlamingoAV}$^{**}$ & - & 57.4 & - & 29.4 & 47.2 & 41.2 & - & - \\
Flamingo-80B~\cite{Alayrac2022FlamingoAV} & - & 84.2 & - & 47.4 & - & - & - & - & - & - \\
\midrule
\uiotacronym$_\texttt{L}$ & 68.5 & 37.1 & 44.0 & 39.6 & 48.2 & 51.0 & 37.5 & 37.8 & 45.7 & 86.1 \\
\uiotacronym$_\texttt{XL}$ & 71.4 & 41.6 & 47.1 & 39.3 & 50.4 & 52.0 & 45.6 & 44.2 & 45.7 & 88.0 \\
\uiotacronym$_\texttt{XXL}$ & 73.8 & 45.6 & 48.8 & 41.5 & 52.1 & 52.2 & 46.8 & 47.7 & 48.9 & 89.3 \\
\bottomrule
\end{tabular}
}
\vspace{-1mm}
\caption{Results on action classification, video captioning, VQA, visual comprehension, audio classification, and audio captioning. $^*$: zero-shot, $^{**}$: few-shot in-context learning.}
\label{tab:audio_video}
\vspace{-2mm}
\end{table}

\label{exp:audio_video}

On referring expression, \uiot{} is ahead of Shikra~\cite{chen2023shikra} and Ferret~\cite{you2023ferret} and matches the scores achieved by Qwen-VL.
On captioning, \uiot{} also achieves a strong CIDEr score~\cite{vedantam2015cider} of 130.3, ahead of Shikra and InstructBLIP~\cite{dai2023instruct_blip} but behind Qwen-VL and mPLUG-Owl2.

Finally, we evaluate using three recently proposed evaluation-only benchmarks. MMB (MMBench~\cite{Liu2023MMBenchIY}) tests multiple facets of 
vision language understanding with multiple choice questions, while SEED-Bench additionally tests video understanding. We show a detailed breakdown of our score in the Appendix \ref{supp:multimodal_benchmark}. Regarding the overall score, \uiot{} has the strongest score of any 7B model on the SEED-Bench leaderboard\footnote{as of 11/17/23}, and scores the highest on MMB by 3.8 points. Notably, it excels LLaVa-1.5 13B model in both benchmarks.
\uiot{} also reaches 87.7 on the POPE object hallucination benchmark~\cite{Li2023EvaluatingOH}, showing that it is not very prone to object hallucination.

Overall, \uiot{} can match or surpass other vision \& language generalist models on these benchmarks despite encompassing many more modalities and supporting high-quality image and audio generation. This shows that its wide breadth of capabilities does not come at the expense of vision/language performance.

\subsection{Video, Audio and other Results}

\begin{table}[t]
\footnotesize
    \centering
    \setlength\tabcolsep{3pt}
    \resizebox{0.9\columnwidth}{!}{
    \begin{tabular}{l | c c c c  }
        \toprule
         &  AP3D & AP3D@15 & AP3D@25 & AP3D@50 \\
         \midrule
         Cube-RCNN~\cite{brazil2023omni3d}  & 50.8 & 65.7 & 54.0 & 22.5 \\
         \midrule
        \uiolarge{} & 42.9 & 54.4 & 45.7 & 21.7  \\
        \uioxl{} & 43.3 & 54.4 & 46.8 & 21.8  \\
        \uioxxl{} & 42.4 & 54.0 & 45.6 & 20.9 \\
        \bottomrule
    \end{tabular}}
    \vspace{-1mm}
    \caption{Single-object 3D detection results on Objectron~\cite{ahmadyan2021objectron}.}
    \vspace{-3mm}
    \label{tab:Objectron}
\end{table}

\uiot{} shows reasonable performance on audio and video classification and captioning, as well as video question answering, as shown in Table~\ref{tab:audio_video}. Notably, \uiot{} outperforms BLIP-2~\cite{BLIP2} and InstructBLIP~\cite{dai2023instruct_blip} on Seed-Bench Temporal~\cite{li2023seed} by 8.5 points. \uiot{} also achieves better performance on Kinetics-Sounds~\cite{arandjelovic2017look} than MBT~\cite{Nagrani2021AttentionBF}, which is trained solely on that dataset.

We show the single-object 3D detection results in Table~\ref{tab:Objectron}. 
Our model shows decent results, similar to Cube-RCNN~\cite{brazil2023omni3d}, on the Objectron benchmark~\cite{ahmadyan2021objectron}. However, its performance drops significantly in multi-object 3D detection tasks, like those on nuScenes~\cite{nuScenes} and Hypersim~\cite{hypersim}. This could be because only 1.0\% of our training data focuses on 3D detection. A potential solution might be to combine 2D and 3D detection techniques. 

In COCO object detection, excluding the `stuff' categories, our model reached an average precision (AP) of 47.2, with AP50 at 57.7 and AP75 at 50.0. However, it has difficulties with images containing many objects. Previous research, like Pix2Seq~\cite{chen2021pix2seq}, suggests that autoregressive models face similar challenges, which can be improved with extensive data augmentation. Our model's data augmentation on object detection is comparatively more limited.

Our model shows weak performance in depth estimation, with an RMSE of 0.623 on NYUv2 depth dataset~\cite{nyu_depth}. However, fine-tuning specifically for this task improved the RMSE to 0.423. In our experiment, we simply normalize the depth map with the max depth value in each dataset. Due to the incompatibility of dense ground-truth depth across different datasets~\cite{ranftl2020towards}, our model failed to capture the exact scale in the current prompt, which could potentially be solved by using better normalization and metric evaluation.  

Appendix \ref{supp:results} shows qualitative visualizations of other tasks, such as single object tracking, future state prediction of robotic manipulation, and image-based 3D view synthesis, \etc

\section{Limitation}

\begin{itemize}
\item Due to memory constraints, we use the base versions of the ViT and AST models for image and audio features throughout the project. Using a larger version of these image and audio encoders could substantially improve performance.

\item While our image generation is more faithful compared to SD-based methods, its quality doesn't match that of the stable diffusion model. Additionally, our audio generation is capped at approximately 4 seconds, which restricts the practical application of the audio outputs.

\item Limited computational resources constrained our exploration of the model's hyperparameters. It's likely that using a significantly larger batch size could enhance the model's performance. 

\item Our model is much less reliable for modalities like depth, video or when requiring more niche abilities like 3D object detection, \etc. This is probably due to the limited variety of tasks we have in these areas. 
\item Improving the quality of our data could enhance the model's performance. However, despite considerable efforts, our human-written prompts still fall short in diversity. We notice a notable decrease in the model's performance when dealing with new instruction tasks, as opposed to those it was trained on. 
\end{itemize}

\section{Conclusion}
We introduced \uiot{}, the first autoregressive multimodal model that is capable of understanding and generating image, text, audio, and action. This model was trained from scratch on a wide range of multimodal data and further refined with instruction tuning on a massive multimodal corpus. We developed various architectural changes to stabilize the multimodal training and proposed a multimodal mixture of denoiser objective to effectively utilize the multimodal signals. Our model achieves promising results across a wide range of tasks. We show that going from LLMs to LMMs enables new capabilities and opportunities. In the future, we would like to extend \uiot{} from the encoder-decoder model to a decoder-only model. Additionally, we plan to expand the model's size, enhance the data quality, and refine the overall model design. 

{
\footnotesize
\noindent \textbf{Acknowledgement}
We thank Klemen Kotar for helping gather Embodied AI pre-training data, Jonathan Frankle from MosaicML for suggesting the mixture of NLP pre-training data, Jack Hessel for interleaved image \& text dataset and Micheal Schmitz for helping support the compute infrastructure.
We also thank Tanmay Gupta for helpful discussions, as well as Hamish Ivison, and Ananya Harsh Jha for their insightful discussions about model design.
We additionally thank Oscar Michel, Yushi Hu and Yanbei Chen for their help editing the paper, and Matt Deitke for help setting up the webpage. Savya Khosla and Derek Hoiem were supported in part by ONR award N00014-23-1-2383.
This research was made possible with cloud TPUs from \href{https://sites.research.google/trc/about/}{Google’s TPU Research Cloud (TRC)}.
}

{
    \small
    \bibliographystyle{ieeenat_fullname}
    \bibliography{main}
}
\clearpage
\appendix

\noindent The appendix includes the following sections:
\begin{itemize} 
\itemsep0em 
    \item Sec~\ref{supp:contributions} - Contributions
    \item Sec~\ref{supp:modelimpldetails} - Model Implementation Details
    \item Sec~\ref{supp:pretraining} - Pre-training Details 
    \item Sec~\ref{supp:instruction-tuning} - Tasks and Instruction Tuning 
    \item Sec~\ref{supp:results} - Experiment Details and Additional Results 
\end{itemize}


\section{Contributions}
\label{supp:contributions}
Jiasen Lu, Christopher Clark, Sangho Lee, and Zichen Zhang collectively contributed to dataset construction, prompt development, and conducted numerous exploratory experiments for this project.

\noindent \textbf{Jiasen Lu}
led and designed the main idea and scope of the project. 
Developed the majority of the model pipeline -- image and audio tokenizer, main architecture, model stabilization, and training objective. 
Led and designed the pre-training and instruction tuning data pipelines.
Conducted experiments with various model and data hyperparameters, oversaw the model training process, and wrote the paper. Coordinate with the whole team. 

\noindent \textbf{Christopher Clark}
co-led and designed the infrastructure, instruction tuning, and evaluation.
Developed the dynamic packing system, modality processing pipeline, and classifier-free guidance for image and audio inference.   
Added the NLP and many V\&L datasets, added many synthetic tasks, and built prompts for instruction-tuning tasks. 
Ran the evaluation in \S~\ref{exp:pretrain} (NLP), \ref{exp:grit}, and \ref{exp:audio_video} (detection, depth) and wrote the paper. 

\noindent \textbf{Sangho Lee}
core contribution to the pre-training data pipeline. 
Added all large-scale multimodal pretraining datasets, and video and audio instruction tuning datasets. Developed sample construction pipeline for pre-training. 
Helped with the model implementation -- position encoding, perceiver resamplers, and model stabilization. 
Ran the evaluation in \S~\ref{exp:pretrain} (audio), \ref{exp:generation} (audio and image FID), \ref{exp:audio_video} (video and audio understanding) and wrote parts of the paper.

\noindent \textbf{Zichen Zhang}
core contribution to the instruction tuning data pipeline. 
Added many V\&L, embodiment, video, audio, data augmentation, and all instruction tuning datasets. 
Built prompts for instruction tuning. 
Investigated the model architectures and training pipelines and stabilized the training. 
Ran the experiments in \S~\ref{exp:pretrain} (image TIFA, SeedBench), \ref{exp:generation} (image TIFA, action), \ref{exp:vision_langauge},
wrote parts of the paper, developed the model demo and project page.

\noindent \textbf{Savya Khosla}
added 3D object detection, optical flow, and multi-point tracking datasets, ran the evaluation of 3D detection, and initiated the demo. 

\noindent \textbf{Ryan Marten}
added part of video and tracking datasets. 

\noindent \textbf{Derek Hoiem}
advised on the research direction.

\noindent \textbf{Aniruddha Kembhavi}
advised on the research direction and evaluation, helped manage compute resources and wrote the paper.

\section{Model Implementation Details}
\label{supp:modelimpldetails}

In this section, we present the implementation details of our model.

\subsection{Detailed of Unified Task Representation}
First, we provide details about how different modalities are represented in our model.
\label{supp:detail_task_representation}

\boldheader{Text representation} 
The Byte Pair Encoding (BPE) vocabulary size is 32000. Similar to \cite{2020t5}, we add 200 additional special tokens to indicated masked spans when de-noising. We further add 10 special tokens that can be used to reference the image, audio, and history input in the text. Two special tokens are to indicate the $\langle\texttt{Image\_Input}\rangle$ and $\langle\texttt{Audio\_Input}\rangle$, and 8 special tokens represent individual elements in the image and audio history inputs, both of which have a maximum of 4 frames. We use a maximum of 512 input and output tokens.

\boldheader{Sparse structures representation}
We use an additional 1000 special tokens to represent all continuous values, such as points, boxes, camera transformation, and 3D cuboids. 
Points are represented with $[y, x]$ coordinates and boxes with $[y_1, x_1, y_2, x_2]$ coordinates with values normalized by the image size.
Camera transformations are represented as polar angle $\theta$, azimuth angle $\phi$, and distance $r$. 1000 special tokens to represent discretized angle from $-\pi$ to $\pi$. Following \cite{brazil2023omni3d}, 3D cuboids are represented with 12 parameters including projected center $[u, v]$, virtual depth $z$, log-normalized box dimension $[\bar{w}, \bar{h}, \bar{l}]$, and continuous allocentric rotation $\bm{p}$. 
\begin{itemize}
    \item $[u, v]$ represent the projected 3D center on the image plane relative to the 2D RoI
    \item $z \in \mathbb{R}_+$ is the object's center depth in meters.
    \item $[\bar{w}, \bar{h}, \bar{l}] \in \mathbb{R}_+$ are the log-normalized physical box dimensions in meters. 
    \item $\bm{p} \in \mathbb{R}^6$ is the continuous 6D allocentric rotation.
\end{itemize}
For 3D cuboid detection, we use prompts to indicate the target format, such as ``\textit{Locate all objects in 3D using projected 3D center, virtual depth, log-normalized box size, and rotation in the image.}''

\begin{figure*}[tb]
    \centering
    \includegraphics[width=\textwidth]{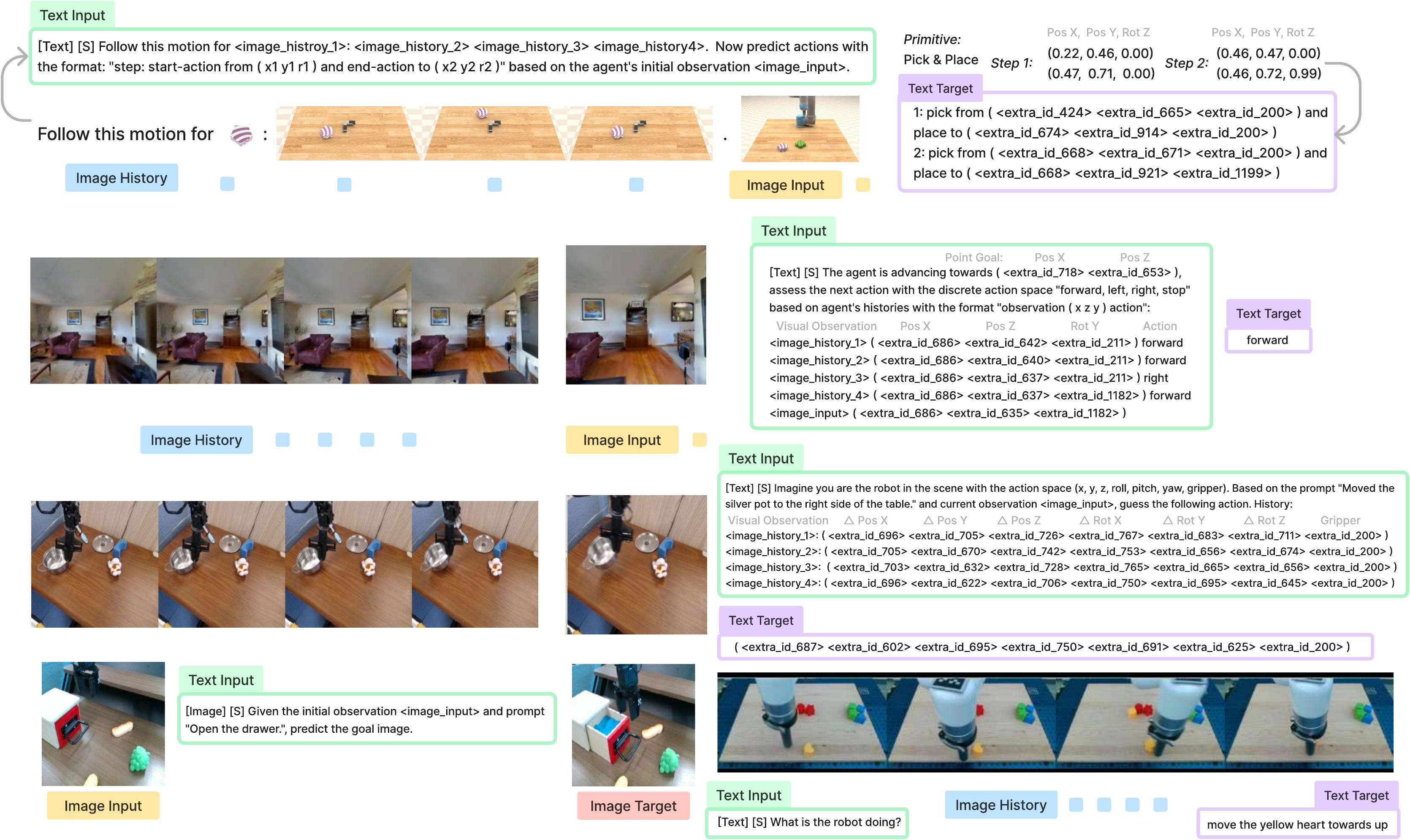}
    \captionof{figure}{Examples of input and target representations for embodied and robot tasks.}
    \vspace{-2mm}
    \label{fig:embodied_supp_illustration}
\end{figure*}

\boldheader{Action representation}
For embodied navigation tasks, the discrete action space is directly represented as texts, \eg ``\texttt{forward}'', ``\texttt{left}'', ``\texttt{right}'', ``\texttt{stop}''. For object manipulation tasks, the action is represented differently based on the robots. Overall,  
the positional change (\eg $(\Delta \texttt{PosX}, \Delta \texttt{PosY}, \Delta \texttt{PosZ})$), rotational change (\eg $(\Delta \texttt{RotX}, \Delta \texttt{RotY}, \Delta \texttt{RotZ})$), and gripper open or close are discretized using the same 1000 special tokens, and we use the text prompt to indicate the input and target format. For tasks that require multi-step planning (\eg VIMA~\cite{jiang2023vima}), the actions are represented as human-readable texts with the indication of steps, skills used (\eg pick, place, or push), and discretized positional and rotational parameters. 
Figure \ref{fig:embodied_supp_illustration} provides a detailed illustration of the robot tasks.

\newcolumntype{P}[1]{>{\centering\arraybackslash}p{#1}}
\newcommand\bothmodels[1]{\multicolumn{2}{P{4cm}}{\cellcolor{olive!5} #1}}

\begin{table}[t!]
\centering
\small
\begin{tabular}{@{} p{0.3cm} @{\hspace{0.05cm}} p{3.5cm} @{\hspace{0.1cm}} P{2cm} @{\hspace{0.1cm}} P{2cm} @{}}
\toprule
\multirow{9}{*}{\rotatebox[origin=c]{90}{{\footnotesize Audio Input}}} & Sample rate & \bothmodels{16000 Hz} \\
                                                                      & FFT hop length & \bothmodels{256 samples} \\
                                                                      & FFT window size & \bothmodels{1024} \\
                                                                      & Mel bins & \bothmodels{128} \\
                                                                      & Subsegment length & \bothmodels{256 hops, ($\approx$4.08 sec)} \\ 
                                                                      & Mel Spectrogram size & \bothmodels{128 mels $\times$ 256 hops}\\
                                                                      & fmin & \bothmodels{0} \\
                                                                      & fmax & \bothmodels{8000} \\
                                                                      & AST patch size & \bothmodels{16} \\
                                                                      & token size & \bothmodels{8 $\times$ 16} \\
                                                                      & Pretrain sub-sample &
                                                                      \bothmodels{64} \\
                                                                      \midrule
                                                                      & Final size & \bothmodels{64 or 128 tokens} \\  \midrule
\multirow{4}{*}{\rotatebox[origin=c]{90}{{\footnotesize Image Input}}}      & ViT patch size & \bothmodels{16} \\                                                              
                                                                      & Pretraining size & \bothmodels{384 $\times$ 384} \\ 
                                                                      & Token size & \bothmodels{24 $\times$ 24} \\ 
                                                                      & Pretrain sub-sample & \bothmodels{288} \\  \midrule
                                                                      & Final size & \bothmodels{288 or 576 tokens} \\  \midrule
\multirow{1}{*}{\rotatebox[origin=c]{90}{{\footnotesize Text}}}        & Seq length & \bothmodels{512} \\            
                                                                      \midrule
                                                                      & Final size & \bothmodels{512 tokens} \\ 
                                                                      \midrule
\multirow{6}{*}{\rotatebox[origin=c]{90}{{\footnotesize Image History}}}      & ViT patch size & \bothmodels{16} \\                                                              
                                                                      & Pretraining size & \bothmodels{256 $\times$ 256} \\ 
                                                                      & Token size & \bothmodels{16 $\times$ 16} \\ 
                                                                      & Pretrain sub-sample & \bothmodels{128} \\
                                                                      & Max num segments & \bothmodels{4} \\
                                                                      & Latent size & \bothmodels{32} \\
                                                                      \midrule
                                                                      & Final size & \bothmodels{32, 64, 96, 128 tokens} \\
                                                                     \midrule
\multirow{6}{*}{\rotatebox[origin=c]{90}{{\footnotesize Audio History}}}      & AST patch size & \bothmodels{16} \\                                                              
                                                                      & Pretraining size & \bothmodels{128 $\times$ 256} \\ 
                                                                      & Token size & \bothmodels{8 $\times$ 16} \\ 
                                                                      & Pretrain sub-sample & \bothmodels{64} \\
                                                                      & Max num segments & \bothmodels{4} \\
                                                                      & Latent size & \bothmodels{16} \\
                                                                      \midrule
                                                                      & Final size & \bothmodels{16, 32, 48, 64 tokens} \\                                                                     
 \bottomrule
\end{tabular}
\vspace{-1mm}
\caption{Input representations details}
\vspace{-3mm}
\label{tab:input_representation}
\end{table}

\boldheader{Images representation} 
Images are encoded with a pre-trained ViT~\cite{dosovitskiy2021vit}. We use the ViT-B checkpoint trained on LAION 2B dataset\footnote{\url{https://github.com/mlfoundations/open_clip}}. For image inputs, we use a maximum length of 576 tokens (\ie $24 \times 24$ patch encoding from a $384 \times 384$ image). We concatenate features from the second and second-last layers of the ViT to capture both low and high-level visual information. To generate the image, we encode these images as discrete tokens \cite{esser2021taming}. Different from \uio~\cite{lu2022unified}, which uses the VQ-GAN trained on ImageNet~\cite{deng2009imagenet} to convert $256 \times 256$ resolution image into $16 \times 16$ tokens, we use the VQ-GAN trained on the Open Images dataset \cite{open_images} with a compression ratio of 8 and a vocabulary size of 16384\footnote{\url{https://github.com/CompVis/taming-transformers}}. This converts $256 \times 256$ resolution image into $32 \times 32$ tokens. We also compare the VQ-GAN tokenizer with the ViT-VQGAN \cite{yu2021vector} and MoVQ \cite{zheng2022movq}. We empirically find VQ-GAN leads to best generation results. 

\boldheader{Dense structures representation}
To handle this modality, we convert per-pixel labels into RGB images. For depth, we construct a grayscale image by normalizing the depth map. For surface normal estimation, we convert the $x/y/z$ orientations into $r/g/b$ values. For segmentation, we train \uiot~to predict a single black-and-white mask for a particular object specified by a class and a bounding box. Instance segmentation (as done in GRIT~\cite{gupta2022grit}) can then be performed by first performing localization for the target class and then performing segmentation for each detected box.
\uio{} instead trains the model to produce an image with a randomly selected color for each instance. We found this makes post-processing difficult since output images sometimes do not exactly follow the color scheme, and the model could struggle with images with many different instances.

\boldheader{Audio representation}
This modality encodes a 4.08-second segment of audio. We take the waveform sampled at 16000 Hz and convert it to a log-mel-scaled spectrogram. We compute the spectrogram for an entire audio segment (4.08 seconds) simultaneously. Each window involves 1024 samples and 256 samples `hops' between windows. The resulting spectrogram has a size of 128 mel bins with 256 windows. We chose these hyperparameters largely around efficiency.
We then encode this with a pre-trained AST \cite{gong2021ast} with the patch size of $16 \times 16$, hence a total of 128 tokens. 

To generate audio, we use ViT-VQGAN \cite{yu2021vector} to convert the spectrograms into discrete tokens. Since the authors of \cite{yu2021vector} did not release the source code or any pre-trained models, we implement and train our own version of ViT-VQGAN with $8\times8$ patch size that encodes a $256 \times 128$ spectrogram into 512 tokens with a codebook size of 8196. The model is trained with the audio on AudioSet \cite{gemmeke2017audio}, ACAV100M \cite{lee2021acav100m}, and YT-Temporal-1B \cite{Zellers_2022_CVPR} datasets. 
After getting the log-mel-scaled spectrograms, we use HiFi-GAN\footnote{\url{https://github.com/jik876/hifi-gan}} \cite{kong2020hifi} vocoder to decode the spectrograms back to waveforms. We train the HiFi-GAN using the same parameters shown in Table \ref{tab:input_representation}. We trained the model on a mixture of AudioSet and LJSpeech \cite{ljspeech} to cover natural sound and human voice. 

\boldheader{History representation}
Images and audio inputs in this history are first encoded in the same way as image and audio inputs.
We then use a perceiver resampler \cite{Alayrac2022FlamingoAV} to further compress the image and audio features and produce a fixed number of visual outputs (32) and audio outputs (16) to reduce the total sequence length of the model. As shown in Table \ref{tab:input_representation}, we consider a maximum of 4 images and audio segments. In our experiments, we test with two different variants of perceiver implementations: 1) a small group of latent embeddings query each frame/segment individually \cite{Alayrac2022FlamingoAV, Awadalla2023OpenFlamingoAO}, 2) a large group of latent embeddings query all history at once. While the second implementation can finely represent the referenced image and audio, the first can preserve better temporal information. Thus, our final implementation uses the first one.

\subsection{2D Rotary Embedding}
\label{supp:rotary}
We use a rotary position encoding to model the relative location of input sequences \cite{su2021roformer}. We chose this primarily because we did not want to use absolute (additive) position embeddings, which would have to be added to the inputs of each encoder, and also wanted to be consistent with the LLaMA \cite{touvron2023llama} position encoding. 

The rotary encoding uses no parameters and instead uses a kernel trick to allow the model to recover relative distances between key and query elements in a transformer's attention head. For text, we apply rotary encoding at each layer of the network. For other modalities, we extend RoPE to two-dimensional cases by splitting each of the
query and key embeddings of transformer attention heads in half and apply separate rotary embeddings constructed by each of the two coordinates to the halves. 

We treat each token (image, audio, image history, and audio history) as having a 2-dimensional position corresponding to 1) $h, w$ coordinates in the image or audio spectrogram, 2) $(t, l)$ where $t$ and $l$ represent the indices of frame and perceiver latent vector in the image or audio history, respectively. Different from \cite{Zellers_2022_CVPR}, which uses a 4-dimensional position to represent all the inputs, we use a combination of learnable segment (modality) embeddings and rotary encoding. 

\newcommand\bothparams[1]{\multicolumn{3}{P{4.3cm}}{\cellcolor{olive!5} #1}}

\begin{table}[t!]
\centering
\small
\resizebox{\columnwidth}{!}{
\begin{tabular}{@{} p{0.3cm} @{\hspace{0.05cm}} p{3.cm} @{\hspace{0.1cm}} P{1.5cm} @{\hspace{0.1cm}} P{1.5cm} @{\hspace{0.1cm}} P{1.5cm} @{}}
\toprule
& & L & XL & XXL \\ 
\midrule
\multirow{12}{*}{\rotatebox[origin=c]{90}{{\footnotesize Transformer}}}  & Params & 1.1B & 3.2B & 6.8B \\ 
                                                              & Vocab size & \bothparams{33280} \\
                                                              & Image vocab size & \bothparams{16512} \\ 
                                                              & Audio vocab size & \bothparams{8320} \\
                                                              & Model dims & 1024 & 2048 & 3072 \\ 
                                                              & MLP dims & 2816 & 5120 & 8192 \\ 
                                                              & encoder layer & \bothparams{24} \\ 
                                                              & decoder layer & \bothparams{24} \\ 
                                                              & Heads & 16 & 16 & 24 \\
                                                              & MLP activations & \bothparams{silu, linear} \\
                                                              & Logits\_via\_embedding & \bothparams{True} \\
                                                              & Dropout & \bothparams{0} \\
                                                              \midrule
\multirow{7}{*}{\rotatebox[origin=c]{90}{{\footnotesize Image Resampler}}}  & Latents size & \bothparams{32} \\ 
                                                                & Model dims & 768 & 1024 & 1024 \\
                                                                & Heads & 12 & 16 & 16 \\
                                                                & Head Dims & \bothparams{64} \\
                                                                & Number layer & \bothparams{2} \\
                                                                & MLP Dims & 2048 & 4096 & 4096 \\
                                                                & MLP activations & \bothparams{gelu}\\
                                                              \midrule
\multirow{7}{*}{\rotatebox[origin=c]{90}{{\footnotesize Audio Resampler}}}  & Latents size & \bothparams{16} \\ 
                                                                & Model dims & 768 & 1024 & 1024 \\
                                                                & Heads & 12 & 16 & 16 \\
                                                                & Head Dims & \bothparams{64} \\
                                                                & Number layer & \bothparams{2} \\
                                                                & MLP Dims & 2048 & 4096 & 4096 \\
                                                                & MLP activations & \bothparams{gelu}\\   
                                                                \midrule
\multirow{7}{*}{\rotatebox[origin=c]{90}{{\footnotesize ViT}}}  & Patch size & \bothparams{16} \\ 
                                                                & Model dims & \bothparams{768} \\
                                                                & Heads & \bothparams{12} \\
                                                                & Head Dims & \bothparams{64} \\
                                                                & Number layer & \bothparams{11} \\
                                                                & MLP Dims & \bothparams{3072} \\
                                                                & MLP activations & \bothparams{gelu}\\   
                                                                \midrule
\multirow{7}{*}{\rotatebox[origin=c]{90}{{\footnotesize AST}}}  & Patch size & \bothparams{16} \\ 
                                                                & Model dims & \bothparams{768} \\
                                                                & Heads & \bothparams{12} \\
                                                                & Head Dims & \bothparams{64} \\
                                                                & Number layer & \bothparams{11} \\
                                                                & MLP Dims & \bothparams{2048} \\
                                                                & MLP activations & \bothparams{gelu}\\                                                             
\bottomrule
\end{tabular}}
\vspace{-1mm}
\caption{Model Hyperparameters}
\vspace{-3mm}
\label{tab:hyperparameters}
\end{table}

\subsection{Dynamic Packing}
\label{supp:dynamic_packing}
Here, we describe the dynamic packing algorithm in more detail.
As is standard practice, when batching together inputs, we pad input tensors to a maximum length and use attention masked to prevent the transformer from attending to padding elements. This, however, is highly inefficient in our multi-modal setting because many modalities are not present in most examples, which results in a huge amount of padding. For example, if one example in the batch has an image output, every other example must be padded with 1024 target image tokens, even if their output is in a different modality.

One solution is to arrange batches so that each batch contains examples with similar numbers of tokens in each modality. This is, however, complicated to do in practice since (1) our data does not fit in RAM, so we cannot easily sort and group data this way, especially if needing to match tokens across five input and three output modalities and (2) our coding framework, JAX~\cite{jax2018github}, does not support variable length tensors when constructing the execution graph which makes handling variable lengths between batches extremely difficult.

Instead, we use \textit{packing}, a process where the tokens of multiple examples are packed into a single sequence, and the attentions are masked to prevent the transformer from cross-attending between examples. Packing is often done as a pre-processing step when handling text, but this does not work in our setup since some parts of our network cannot operate on packed data (\eg, the \vae{} or image ViT). Instead, we start with an unpacked batch of examples, run these components first, and then dynamically pack the resulting tokens in a backdrop-compatible way before running the transformer. To run efficiently on TPUs we pack examples using matrix multiplication with carefully constructed one-hot matrices. 

To account for all modalities, the maximum sequence length our transformer needs to take as input is 1152, and the maximum target length is 2048. When packing, we can generally pack two examples into an input sequence of 864 and a target sequence of 1280, which gives a roughly 4x speed up due to reduced sequence length and the ability to process two examples simultaneously. When streaming data, packing cannot be done reliably. For example, if two consecutive examples have an image output, they cannot be packed since they will total over 1280 output tokens. To handle this, we use a heuristic algorithm to re-arrange data as it is being streamed. The algorithm keeps a small pool of examples in memory. Given a new example, it pairs it with the largest example in the pool it can be packed with and outputs both as a pair. If no such example exists, it adds the example to the pool. If the pool reaches a maximum size of 10, the largest example is emitted and processed without being packed with another example. We find this occurs less than 0.1\% of the time during training.

\subsection{Full Model Details}
\label{supp:full_model_details}

\begin{figure*}[t]
    \centering
    \includegraphics[width=0.8\textwidth]{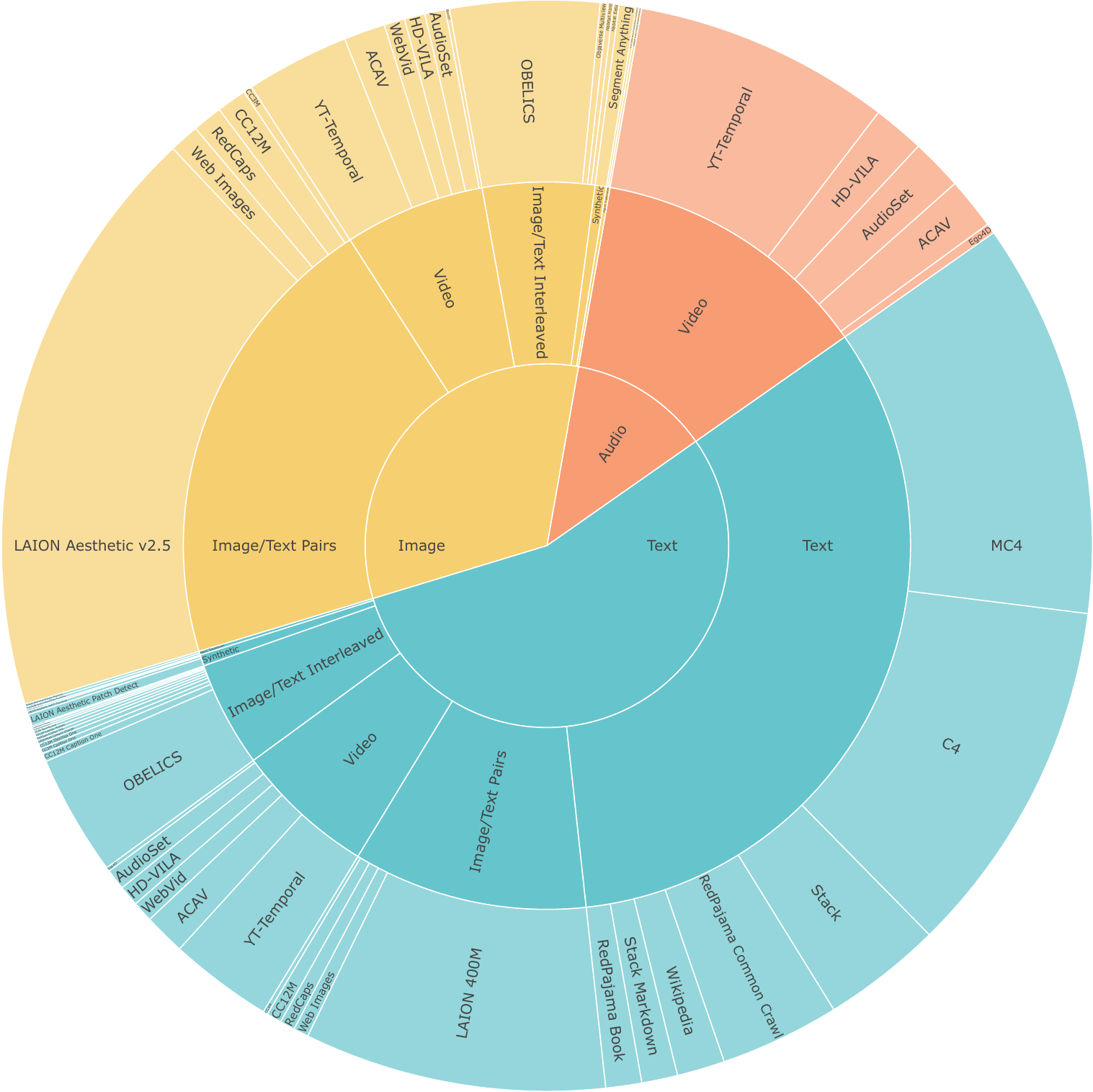}
    \caption{Pre-training data distribution, segments proportional to sampling rates. The inner section shows the target modality, the middle section shows the type of data, and the third shows particular datasets.}
    \vspace{-1mm}
    \label{fig:pretraining_sunburst}
\end{figure*}

In Table \ref{tab:hyperparameters}, we present the full hyperparameters of our model. 
During pre-training, we train the \uiolarge, \uioxl, and \uioxxl~with a batch size of 512 due to memory limit. We sub-sample 50\% of the image, audio, and history inputs patches. The total packing length is 864 for the encoder and 1280 for the decoder.
During instruction tuning, we train all of our models with a batch size 256 due to computing constraints. We sub-sample 87.5\% of the image, audio, and history input patches. The total packing length is 1024 for pretraining and 1280 for instruction tuning. 8-way in-layer parallelism and 64-way data parallelism were used to scale up to the 7B model training.

We train for 1.5 million steps with an effective batch size of 512. This results in training on approximately 1 trillion tokens. During pre-training, we keep at most 50\% of the image patches in the image history or image encoder, as is common practice with MAE pre-training~\cite{he2022masked}. We use up to four images/segments in image/audio history. 
\section{Pre-Training Details}
\label{supp:pretraining}

\begin{table*}[]
\footnotesize
\setlength\tabcolsep{2.5pt}
\renewcommand{\arraystretch}{1.25}
\renewcommand{\indent}{\hspace{0.2cm}}
    \centering
    \begin{tabular}{l|c c | c c c c c c c | c c c c c c }
\toprule
 & Size & Rate & Text & Sparse & Dense & Image & Audio & ImageH & AudioH & Text & Sparse & Dense & Image & Audio \\
\midrule
\band \textbf{Text} & 6.6b & 33.0 & \checkmark & - & - & - & - & - & - & \checkmark & - & - & - & - \\
\indent MC4~\cite{xue2020mt5} & 5.0b & 11.7 & \checkmark & - & - & - & - & - & - & \checkmark & - & - & - & - \\
\indent C4~\cite{habernal2016c4corpus} & 266m & 10.6 & \checkmark & - & - & - & - & - & - & \checkmark & - & - & - & - \\
\indent Stack~\cite{Kocetkov2022TheStack} & 147m & 3.55 & \checkmark & - & - & - & - & - & - & \checkmark & - & - & - & - \\
\indent RedPajama CC~\cite{together2023redpajama} & 1.2b & 3.55 & \checkmark & - & - & - & - & - & - & \checkmark & - & - & - & - \\
\indent Wikipedia & 6.8m & 1.42 & \checkmark & - & - & - & - & - & - & \checkmark & - & - & - & - \\
\indent RedPajama Book~\cite{together2023redpajama} & 13m & 1.06 & \checkmark & - & - & - & - & - & - & \checkmark & - & - & - & - \\
\indent Stack-Markdown~\cite{Kocetkov2022TheStack} & 34m & 1.06 & \checkmark & - & - & - & - & - & - & \checkmark & - & - & - & - \\
\band \textbf{Image/Text} & 970m & 31.3 & \checkmark & - & - & \checkmark & - & - & - & \checkmark & - & - & \checkmark & - \\
\indent LAION Aesthetics v2.5~\cite{laion-aesthetic} & 491m & 17.7 & \checkmark & - & - & \checkmark & - & - & - & - & - & - & \checkmark & - \\
\indent LAION-400M~\cite{schuhmann2021laion400m} & 346m & 8.95 & \checkmark & - & - & \checkmark & - & - & - & \checkmark & - & - & - & - \\
\indent CC12M~\cite{changpinyo2021cc12m} & 11m & 1.48 & \checkmark & - & - & \checkmark & - & - & - & \checkmark & - & - & \checkmark & - \\
\indent RedCaps~\cite{desai2021redcaps} & 12m & 1.39 & \checkmark & - & - & \checkmark & - & - & - & \checkmark & - & - & \checkmark & - \\
\indent Web Images & 107m & 1.33 & \checkmark & - & - & \checkmark & - & - & - & \checkmark & - & - & \checkmark & - \\
\indent CC3M~\cite{sharma2018cc3m} & 3.0m & 0.49 & \checkmark & - & - & \checkmark & - & - & - & \checkmark & - & - & \checkmark & - \\
\band \textbf{Video} & 181m & 25.0 & \checkmark & - & - & \checkmark & \checkmark & \checkmark & \checkmark & \checkmark & - & - & \checkmark & \checkmark \\
\indent YT-Temporal~\cite{Zellers_2022_CVPR} & 146m & 13.7 & \checkmark & - & - & \checkmark & \checkmark & \checkmark & \checkmark & \checkmark & - & - & \checkmark & \checkmark \\
\indent ACAV~\cite{lee2021acav100m} & 17m & 3.98 & \checkmark & - & - & \checkmark & - & \checkmark & - & \checkmark & - & - & \checkmark & \checkmark \\
\indent HD-VILA~\cite{xue2022hdvila} & 7.1m & 2.75 & \checkmark & - & - & \checkmark & \checkmark & \checkmark & \checkmark & \checkmark & - & - & \checkmark & \checkmark \\
\indent AudioSet~\cite{gemmeke2017audio} & 1.7m & 2.75 & \checkmark & - & - & \checkmark & \checkmark & \checkmark & \checkmark & \checkmark & - & - & \checkmark & \checkmark \\
\indent WebVid~\cite{bain2021webvid} & 9.2m & 1.23 & \checkmark & - & - & \checkmark & \checkmark & \checkmark & \checkmark & \checkmark & - & - & \checkmark & - \\
\indent Ego4D~\cite{Grauman_2022_CVPR} & 0.7m & 0.55 & \checkmark & - & - & \checkmark & \checkmark & \checkmark & \checkmark & \checkmark & - & - & \checkmark & \checkmark \\
\band \textbf{Interleaved Image/Text} & 157m & 8.70 & \checkmark & - & - & \checkmark & - & \checkmark & - & \checkmark & - & - & \checkmark & - \\
\indent OBELICS~\cite{laurenccon2023obelics} & 131m & 8.00 & \checkmark & - & - & \checkmark & - & \checkmark & - & \checkmark & - & - & \checkmark & - \\
\indent CC12M Interleaved & 11m & 0.35 & \checkmark & - & - & \checkmark & - & \checkmark & - & \checkmark & - & - & - & - \\
\indent CC3M Interleaved & 3.0m & 0.21 & \checkmark & - & - & \checkmark & - & \checkmark & - & \checkmark & - & - & - & - \\
\indent RedCaps Interleaved & 12m & 0.14 & \checkmark & - & - & \checkmark & - & \checkmark & - & \checkmark & - & - & - & - \\
\band \textbf{Multi-View} & 3.4m & 0.67 & \checkmark & - & - & \checkmark & - & \checkmark & - & - & \checkmark & - & \checkmark & - \\
\indent CroCo Habitat~\cite{weinzaepfel2022croco,savva2019habitat} & 2.6m & 0.33 & \checkmark & - & - & \checkmark & - & \checkmark & - & - & - & - & \checkmark & - \\
\indent Objaverse~\cite{deitke2023objaverse} & 0.8m & 0.33 & \checkmark & - & - & \checkmark & - & \checkmark & - & - & \checkmark & - & \checkmark & - \\
\band \textbf{Agent Trajectories} & 1.3m & 0.33 & \checkmark & - & - & \checkmark & - & \checkmark & - & \checkmark & - & - & \checkmark & - \\
\indent ProcTHOR~\cite{deitke2022} & 0.7m & 0.17 & \checkmark & - & - & \checkmark & - & \checkmark & - & \checkmark & - & - & \checkmark & - \\
\indent Habitat~\cite{savva2019habitat} & 0.6m & 0.17 & \checkmark & - & - & \checkmark & - & \checkmark & - & \checkmark & - & - & \checkmark & - \\
\band \textbf{Synthetic} & 504m & 1.00 & \checkmark & \checkmark & - & \checkmark & - & - & - & - & \checkmark & \checkmark & - & - \\
\indent Segment Anything~\cite{Kirillov_2023_ICCV} & 1.1m & 0.50 & \checkmark & \checkmark & - & \checkmark & - & - & - & - & - & \checkmark & - & - \\
\indent Laion Aesthetics Patches & 491m & 0.45 & \checkmark & - & - & \checkmark & - & - & - & - & \checkmark & - & - & - \\
\indent RedCaps Patches & 12m & 0.05 & \checkmark & - & - & \checkmark & - & - & - & - & \checkmark & - & - & - \\
\band \textbf{All} & 8.5b & 100 & \checkmark & \checkmark & - & \checkmark & \checkmark & \checkmark & \checkmark & \checkmark & \checkmark & \checkmark & \checkmark & \checkmark \\

\bottomrule
    \end{tabular}
    \vspace{-1mm}
    \caption{Datasets used for pre-training, rate shows the sampling percentage during pre-training and size shows the approximate number of examples if iterating through the data once.}
    \vspace{-1mm}

    \label{tab:pretrain}

\end{table*}

In this section, we provide additional details about the data \uiot{} is pre-trained on.
The datasets we use for pre-training are listed in Table~\ref{tab:pretrain}. 
Unless otherwise specified, we use the pre-training objective described in Section 3.3, where one of the present modalities is randomly selected as the target. We sample data to ensure all the output modalities are well represented and to balance how often our various corpora are used based on their size. The distribution is shown in Figure~\ref{fig:pretraining_sunburst}.

\subsection{Data Sources}
\boldheader{Text} 
Our data follows the mixture used by MPT-7B \cite{MosaicML2023Introducing}.

\boldheader{Image \& Text}
Image \& text paired data comes from various unsupervised corpora, shown in Table~\ref{tab:pretrain}. For LAION data, we only generate images from image/text pairs from LAION aesthetic, which contains higher quality images, while we generate text for image/text pairs from LAION 400M. We also only keep images from LAION if they are marked as being unlikely to be NSFW in the LAION metadata.
Web images is a dataset of images we download and focuses on icons and stylized images.

\boldheader{Video}
We gather a total of 180M short videos from various sources.
During training, we pick a random sequence of up to five frames from the video. The first four will be encoded with an image/audio history encoder, while the fifth frame will be encoded with the image/audio encoder. The text matching these frames is encoded with a text encoder along with marker tokens to show where each frame occurred as stated in~\ref{supp:detail_task_representation}, or, if the dataset only includes a single caption that is not aligned with individual frames, the entire caption is encoded instead. The text, audio, or image modality can be selected as the target modality. As usual, other modalities are randomly masked, and the target modality is randomly masked or injected with noise in the input.
Note we have sub-sampled data from many of these corpora to keep the dataset size more manageable, and sometimes due to broken video links.

\boldheader{Interleaved Image \& Text}
We primarily use OBELICS~\cite{laurenccon2023obelics}, which contains paragraphs and images interleaved together. For each document, we randomly select an image or a paragraph as the target and use up to the previous four (if the target is an image) or five (if the target is a paragraph) images as context. 
The last image is encoded with the image encoder, and the remaining images are encoded in the image history.
The text matching those images is concatenated and interjected with marker tokens to indicate where the images in the image history or image input occur.
 We either do de-noising, where a noisy version of the target is included in the input, or generation, where the target is not part of the input, although we always include both the text and image input modalities.

In addition, we construct interleaved data by interleaving multiple images and captions from several image/text pair corpora. The images are encoded as the image input and/or the image history, and matching text is constructed by specifying the caption for one, or all, of these images using special tokens to mark which image each caption refers to. For this task, we only target the text modality, and train the model to either (1) de-noise the caption of a single image, (2) generate a caption for a single image that is specified in an input prompt using a marker token or (3) generate a sequence of marker tokens and captions that describe each input image.
This task aims to ensure the model learns the semantics of the images in the history and understands the marker tokens.

\boldheader{Multi-View}
We train on the cross-view completion task from CroCo~\cite{weinzaepfel2022croco}, where the model must complete a heavily noised image using an image of the same scene, but from a slightly different angle, as context. The noised input is encoded as an image and the second image is encoded through the image history encoder. 
In addition, we generate data using Objaverse~\cite{deitke2023objaverse} objects by capturing multiple views of the object in 3D, and either specify the camera coordinates in the input text and train the model to generate a new image matching new camera coordinates, or train the model to predict how the camera has moved between different images. We further augment the view synthesis task by providing in-context examples. For example, by giving one or more examples of the views and transformations in the image history, the model predicts the new view from the new camera transformation specified by the prompt.
Both tasks aim to improve the model's 3D understanding during pre-training.

\boldheader{Agent Trajectory}
We use scripted shortest path trajectories in ProcTHOR~\cite{deitke2022} and human-collected demonstrations in Habitat~\cite{ramrakhya2022, savva2019habitat}. While the original datasets are for object navigation with relatively long episode lengths, we only subsample from the last few frames for image history and image input such that mostly the target object is within the observation. The task is randomly selected from 1) generating the next visual observation frame as the target image, 2) predicting the next positional observation coordinates as the text target, and 3) predicting the next action as the text target. 1) requires inferring from the image and image history input and the last action specified in the text input, 2) further requires the location information, and 3) is based on the target object name and visual observations for the next action prediction. 

\boldheader{Synthetic}
We add two synthetic tasks. First, we use the automatically annotated data from Segment Anything~\cite{Kirillov_2023_ICCV}. We give the model either a set of points or a bounding box as input and train it to generate a segmentation mask as output. Second, we add artificial patches of various shapes and colors to images from other unsupervised datasets and train the model to output their locations in order to train the model to generate sparse coordinates as output. We additionally train the model to output the total number of patches on the image to pre-train its counting abilities.

\begin{figure*}
    \centering
    \includegraphics[width=0.8\textwidth]{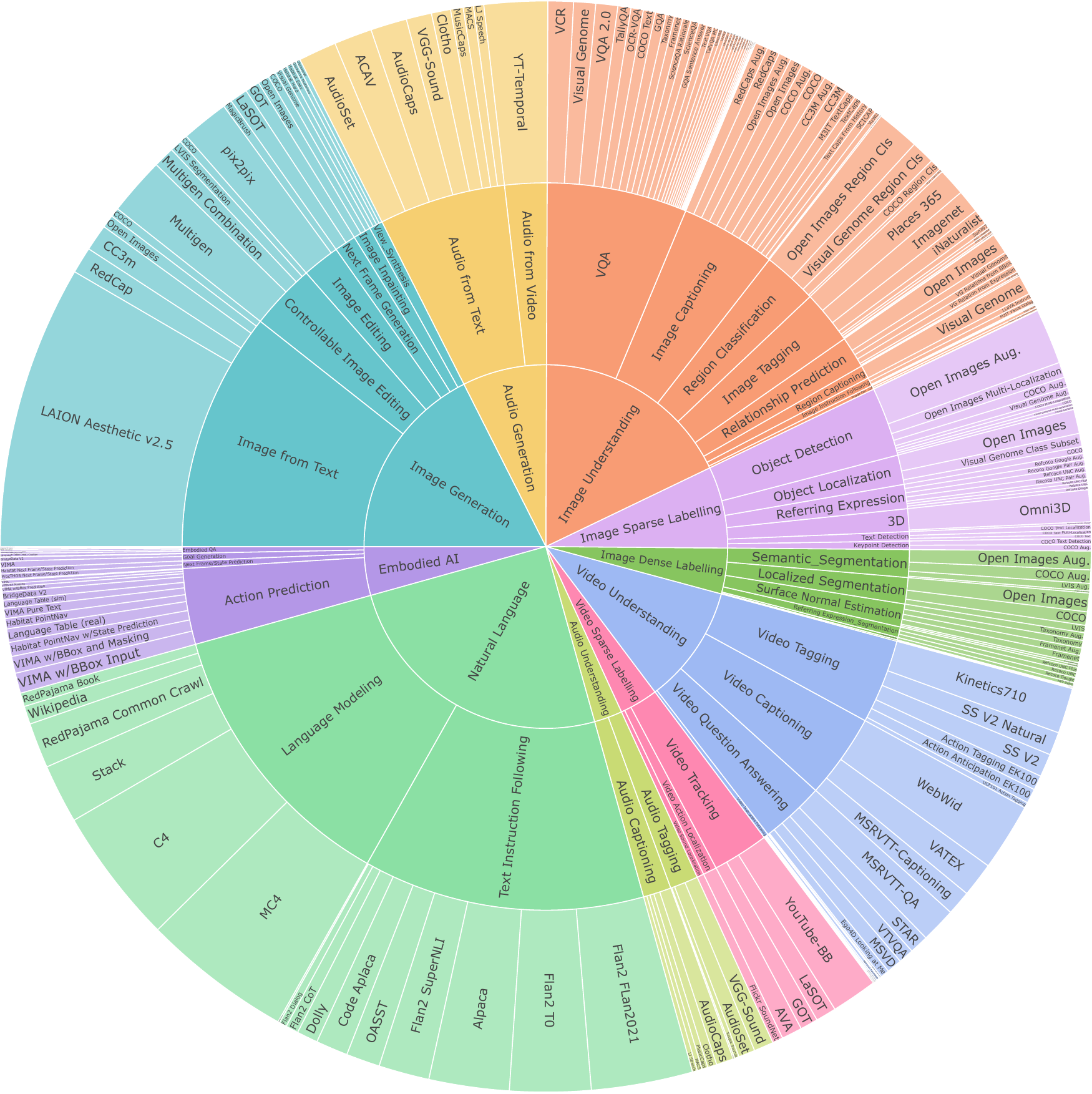}
    \caption{Sunburst chart of our instruction-tuning mixtures, sections are proportional to sampling rates.}
    \label{fig:instruction-tuning-rates}
\end{figure*}

\section{Instruction Tuning Details}
\label{supp:instruction-tuning}
\begin{table*}[]
\thisfloatpagestyle{empty}
\vspace{-0.4cm}
    \centering
    \footnotesize
\setlength\tabcolsep{2.5pt}
\renewcommand{\arraystretch}{1.13}
\renewcommand{\indent}{\hspace{0.2cm}}
\begin{tabular}{l | c c c | c c c c c c c | c c c c c}
\toprule
 & Size & Rate & Datasets & Text & Sparse & Dense & Image & Audio & ImageH & AudioH & Text & Sparse & Dense & Image & Audio \\
\midrule
\band \textbf{Image Generation} & 506m & 17.6 & 21 & \checkmark & \checkmark & \checkmark & \checkmark & - & \checkmark & \checkmark & \checkmark & - & - & \checkmark & - \\
\indent Image from Text & 497m & 10.6 & 5 & \checkmark & - & - & - & - & - & - & - & - & - & \checkmark & - \\
\indent Controllable Image Editing & 3.0m & 2.92 & 4 & \checkmark & - & \checkmark & \checkmark & - & \checkmark & - & - & - & - & \checkmark & - \\
\indent Image Editing & 1.1m & 1.66 & 3 & \checkmark & - & - & \checkmark & - & - & - & - & - & - & \checkmark & - \\
\indent Next Frame Generation & 24k & 0.96 & 2 & \checkmark & \checkmark & - & - & - & \checkmark & \checkmark & - & - & - & \checkmark & - \\
\indent Image Inpainting & 1.0m & 0.79 & 3 & \checkmark & \checkmark & - & \checkmark & - & - & - & - & - & - & \checkmark & - \\
\indent View Synthesis & 4.2m & 0.60 & 4 & \checkmark & - & - & \checkmark & - & \checkmark & - & \checkmark & - & - & \checkmark & - \\
\band \textbf{Audio Generation} & 164m & 7.50 & 9 & \checkmark & - & - & \checkmark & \checkmark & \checkmark & \checkmark & - & - & - & - & \checkmark \\
\indent Audio from Text & 19m & 5.62 & 8 & \checkmark & - & - & - & - & - & \checkmark & - & - & - & - & \checkmark \\
\indent Audio from Video & 145m & 1.88 & 1 & \checkmark & - & - & \checkmark & \checkmark & \checkmark & \checkmark & - & - & - & - & \checkmark \\
\band \textbf{Image Understanding} & 53m & 17.8 & 73 & \checkmark & \checkmark & - & \checkmark & - & \checkmark & - & \checkmark & - & - & - & - \\
\indent VQA & 5.8m & 6.23 & 31 & \checkmark & - & - & \checkmark & - & - & - & \checkmark & - & - & - & - \\
\indent Image Captioning & 32m & 4.25 & 14 & \checkmark & - & - & \checkmark & - & - & - & \checkmark & - & - & - & - \\
\indent Region Classification & 6.1m & 2.41 & 4 & \checkmark & \checkmark & - & \checkmark & - & - & - & \checkmark & - & - & - & - \\
\indent Image Tagging & 3.8m & 2.38 & 8 & \checkmark & - & - & \checkmark & - & - & - & \checkmark & - & - & - & - \\
\indent Relationship Prediction & 0.8m & 1.41 & 6 & \checkmark & \checkmark & - & \checkmark & - & - & - & \checkmark & - & - & - & - \\
\indent Region Captioning & 3.5m & 0.60 & 1 & \checkmark & \checkmark & - & \checkmark & - & - & - & \checkmark & - & - & - & - \\
\indent Image Instruction Following & 0.4m & 0.37 & 6 & \checkmark & - & - & \checkmark & - & - & - & \checkmark & - & - & - & - \\
\indent Image Pair  QA & 0.1m & 0.17 & 3 & \checkmark & - & - & \checkmark & - & \checkmark & - & \checkmark & - & - & - & - \\
\band \textbf{Image Sparse Labelling} & 13m & 7.25 & 26 & \checkmark & \checkmark & - & \checkmark & - & \checkmark & - & - & \checkmark & - & \checkmark & - \\
\indent Object Detection & 5.3m & 3.08 & 9 & \checkmark & - & - & \checkmark & - & - & - & - & \checkmark & - & - & - \\
\indent Object Localization & 6.0m & 1.31 & 3 & \checkmark & - & - & \checkmark & - & - & - & - & \checkmark & - & - & - \\
\indent Referring Expression & 0.2m & 1.08 & 7 & \checkmark & - & - & \checkmark & - & - & - & - & \checkmark & - & - & - \\
\indent 3D & 1.0m & 1.00 & 2 & \checkmark & - & - & \checkmark & - & \checkmark & - & - & \checkmark & - & \checkmark & - \\
\indent Text Detection & 37k & 0.41 & 3 & \checkmark & - & - & \checkmark & - & - & - & - & \checkmark & - & - & - \\
\indent Keypoint Detection & 0.3m & 0.38 & 2 & \checkmark & \checkmark & - & \checkmark & - & - & - & - & \checkmark & - & - & - \\
\band \textbf{Image Dense Labelling} & 6.9m & 4.06 & 19 & \checkmark & \checkmark & - & \checkmark & - & \checkmark & - & - & - & \checkmark & - & - \\
\indent Semantic Segmentation & 2.4m & 1.23 & 4 & \checkmark & - & - & \checkmark & - & - & - & - & - & \checkmark & - & - \\
\indent Localized Segmentation & 3.2m & 1.17 & 3 & \checkmark & \checkmark & - & \checkmark & - & - & - & - & - & \checkmark & - & - \\
\indent Surface Normal Estimation & 1.1m & 1.03 & 6 & \checkmark & - & - & \checkmark & - & - & - & - & - & \checkmark & - & - \\
\indent Referring Expression Segmentation & 0.1m & 0.47 & 3 & \checkmark & - & - & \checkmark & - & - & - & - & - & \checkmark & - & - \\
\indent Depth Estimation & 47k & 0.11 & 1 & \checkmark & - & - & \checkmark & - & - & - & - & - & \checkmark & - & - \\
\indent Optical Flow & 24k & 0.06 & 2 & \checkmark & - & - & \checkmark & - & \checkmark & - & - & - & \checkmark & - & - \\
\band \textbf{Video Understanding} & 13m & 10.6 & 24 & \checkmark & - & - & \checkmark & \checkmark & \checkmark & \checkmark & \checkmark & - & - & - & - \\
\indent Video Captioning & 9.1m & 3.75 & 3 & \checkmark & - & - & \checkmark & - & \checkmark & \checkmark & \checkmark & - & - & - & - \\
\indent Video Tagging & 1.1m & 3.75 & 6 & \checkmark & - & - & \checkmark & - & \checkmark & \checkmark & \checkmark & - & - & - & - \\
\indent Video Question Answering & 2.5m & 2.84 & 9 & \checkmark & - & - & \checkmark & \checkmark & \checkmark & \checkmark & \checkmark & - & - & - & - \\
\indent Video Instruction Following & 0.2m & 0.21 & 6 & \checkmark & - & - & \checkmark & - & \checkmark & \checkmark & \checkmark & - & - & - & - \\
\band \textbf{Video Sparse Labelling} & 0.4m & 3.42 & 5 & \checkmark & \checkmark & - & \checkmark & - & \checkmark & \checkmark & - & \checkmark & - & - & - \\
\indent Video Tracking & 0.2m & 2.50 & 3 & \checkmark & \checkmark & - & \checkmark & - & \checkmark & \checkmark & - & \checkmark & - & - & - \\
\indent Video Action Localization & 0.2m & 0.61 & 1 & \checkmark & - & - & \checkmark & - & \checkmark & \checkmark & - & \checkmark & - & - & - \\
\indent Video Sound Localization & 2.5k & 0.31 & 1 & \checkmark & - & - & \checkmark & - & \checkmark & \checkmark & - & \checkmark & - & - & - \\
\band \textbf{Audio Understanding} & 2.2m & 2.50 & 10 & \checkmark & - & - & \checkmark & \checkmark & \checkmark & - & \checkmark & - & - & - & - \\
\indent Audio Tagging & 2.1m & 1.25 & 5 & \checkmark & - & - & \checkmark & \checkmark & \checkmark & - & \checkmark & - & - & - & - \\
\indent Audio Captioning & 75k & 1.25 & 5 & \checkmark & - & - & - & \checkmark & - & - & \checkmark & - & - & - & - \\
\band \textbf{Natural Language} & 11m & 25.0 & 17 & \checkmark & - & - & - & - & - & - & \checkmark & - & - & - & - \\
\indent Text Instruction Following & 11m & 12.5 & 10 & \checkmark & - & - & - & - & - & - & \checkmark & - & - & - & - \\
\indent Language Modeling & - & 12.5 & 7 & \checkmark & - & - & - & - & - & - & \checkmark & - & - & - & - \\
\band \textbf{Embodied AI} & 7.2m & 4.33 & 23 & \checkmark & - & - & \checkmark & - & \checkmark & - & \checkmark & \checkmark & - & \checkmark & - \\
\indent Action Prediction & 4.3m & 3.37 & 12 & \checkmark & - & - & \checkmark & - & \checkmark & - & \checkmark & - & - & - & - \\
\indent Next Frame/State Prediction & 1.3m & 0.33 & 2 & \checkmark & - & - & \checkmark & - & \checkmark & - & \checkmark & - & - & \checkmark & - \\
\indent Goal Generation & 0.7m & 0.33 & 3 & \checkmark & - & - & \checkmark & - & \checkmark & - & - & - & - & \checkmark & - \\
\indent Embodied QA & 1.0m & 0.30 & 6 & \checkmark & - & - & \checkmark & - & \checkmark & - & \checkmark & \checkmark & - & - & - \\
\indent All Tasks & 775m & 100 & 227 & \checkmark & \checkmark & \checkmark & \checkmark & \checkmark & \checkmark & \checkmark & \checkmark & \checkmark & \checkmark & \checkmark & \checkmark \\
\bottomrule

    \end{tabular}
    \caption{Instruction tuning training mixture. Due to the number of datasets used, we group them by task and only show statistics for each group. The rate shows the sampling rate, size shows the number of examples of iterating through the data once, and datasets show the number of individual data sources used for the tasks.}
    \label{tab:instruction-tuning-data}
\end{table*}

In this section, we provide additional details about the instruction tuning data and individual tasks \uiot{} supports. An overview of the instruction tuning data is shown in Table~\ref{tab:instruction-tuning-data}. 
We show a visualization including individual datasets in Figure~\ref{fig:instruction-tuning-rates}. We sample broad categories of tasks evenly and then generally sample individual datasets in proportion to the square root of their size, although with some minor hand-engineered adjustments to downweight noisy datasets or upweight very rare tasks.

\subsection{Natural Language}
For natural language data we use the mixture from FlanV2~\cite{flanv2}, which in turn includes data from Muffin~\cite{wei2021finetuned}, T0-SF~\cite{sanh2021multitask}, NIV2~\cite{supernatural_instructions}, and CoT annotations, as well data from Alpaca~\cite{peng2023instruction}, Dolly~\cite{dolly}, Open Assistant~\cite{kopf2023openassistant}, and MDPP~\cite{austin2021program}. In addition, we continue pre-training on our unsupervised NLP mixture from our fine-tuning stage to ensure the model does not forget information learned from unsupervised data during the extensive instruction-tuning stage.

\subsection{Image Generation}
For text-to-image generation, we use the same image/text pairs we used during pre-training, as well as localized narratives from Open Images~\cite{open_images} and captions from COCO~\cite{coco} and Visual Genome (VG)~\cite{visual_genome}. 
Our prompts for these tasks specify that the image might be noisy or approximate for unsupervised corpora (\eg ``Generate an image that roughly matches this text: \{caption\}") and give hints as to the style for supervised corpora (\eg ``What do you see in this image? Plainly describe the individual element you observe." for localized narratives) to help disambiguate the stylistic differences between the datasets. We use simple prompts (\eg ``Caption this image.") for the COCO captions.

We additionally train the model to generate images through view synthesis~\cite{weinzaepfel2022croco,deitke2023objaverse} as was done during pre-training. We also integrate data for image editing~\cite{brooks2023instructpix2pix,zhang2023magicbrush} and image editing based on various dense control signals such as depth maps, edges, segmentation, etc. Following~\cite{qin2023unicontrol}, and the segmentation-based image generation from \uio{} using data from COCO and LVIS~\cite{lvis}. Finally, we train on inpainting by masking a region of an input image that contains an object and training the model to generate the complete image given the object name and location. We derive data for this task from the object annotation data in COCO, Open Images, and VG.

During inference, we use top-p sampling, also known as nucleus sampling~\cite{holtzman2020curious}, for generating images with the temperature $t=1.0$ and $p=0.95$. We also enable classifier-free guidance~\cite{ho2021classifier} by replacing the prompt with the un-informative prompt ``An image of a random picture.'' 10\% of the time during training. That prompt is then used as the classifier-free prompt with a guidance scale of $\alpha=10.0$ during inference.

\subsection{Audio Generation}
Datasets for audio generation from text include AudioCaps~\cite{audiocaps}, Clotho~\cite{drossos2020clotho}, MACS~\cite{martin2021diversity}, MusicCaps~\cite{agostinelli2023musiclm}, and LJSpeech~\cite{ljspeech}. During training, we divided the audio into 4-second-long segments and then generated one segment of the target audio, giving both the text and any previous segments as input. 
We also train on the next-frame prediction task, which aims to generate the audio for the next frame in a video from YT-Temporal-1B~\cite{Zellers_2022_CVPR}.

Our prompts for these tasks specify the characteristics of target audio; \eg, ``Generate the sound/music based on the description: \{caption\}'' for natural sound and music, respectively, and ``Speak: \{passage\}'' for speech.
We use the same sampling method as the image generation, the top-p sampling with the temperature $t=1.0$ and $p=0.95$. We do not use the classifier-free guidance because it can lead to poor performance. When generating audio longer than 4.08 seconds during inference, we generate an initial segment that is 4.08 seconds long and then extend it by generating additional segments using previous audio segments as the audio history input.

\subsection{Image Understanding}
These tasks require generating text in response to a query about an image or a pair of images.
We use the data from M$^3$IT~\cite{li2023m} and MIMIC-IT~\cite{Li2023OtterAM,li2023mimic}, as well as a variety of other additional sources. For VQA, we add GQA~\cite{hudson2019gqa}, TallyQA~\cite{acharya2019tallyqa}, OK-VQA~\cite{Marino2019OKVQAAV}, A-OKVQA~\cite{schwenk2022okvqa}, OCR-based VQA datasets~\cite{mishra2019ocrvqa,singh2019towards}, Visual Genome, ScienceQA~\cite{lu2022learn}, VCR~\cite{zellers2019vcr} and VizWiz~\cite{gurari2018vizwiz}. For image tagging we add Caltech Birds~\cite{caltechbirds}, iNaturalist~\cite{van2018inaturalist}, Sun397~\cite{xiao2010sun}, and Places365~\cite{zhou2017places}. For region classification, we add examples derived from object annotation from Open Images, VG, and COCO. We categorize datasets with open-ended responses such as LLaVa~\cite{llava}, Visual Storytelling~\cite{huang2016visual}, and Visual Dialog~\cite{das2017visual} as visual instruction following, and we categorize NLVR~\cite{suhr2018nlvr} and the ``spot the differences'' tasks from MIMIC-IT as image pair QA. For image pair QA tasks, we encode the second image in the image history modality.

We also add a grounded relationship prediction task using data from Visual Genome and VSR~\cite{liu2023vsr} as well as image captioning using the same supervised sources we use for image generation.

We again put stylistic hints in the prompts for these tasks. For example, in VQA and captioning datasets, we specify to return a short answer (\eg  ``Answer this question very succinctly: \{question\}"), which we find is critical to allow the model to produce longer, more natural responses when asked user questions. Likewise, we roughly specify the kind of class to output for image tagging, \eg, ``"What is the scientific name of this animal?" for the iNaturalist dataset.

\subsection{Image Sparse Labelling}
These tasks require outputting sparse coordinates based on an input image. We use Open Images, Visual Genome, and COCO for object detection and localization, which requires detecting all objects belonging to a specific class and three COCO referring expression datasets~\cite{referitgame,yu2016modeling,nagaraja2016modeling} for referring expressions.

In addition, we train on the OmniLabel~\cite{brazil2023omni3d} 3D detection dataset by generating the projected 3D center, virtual depth, log-normalized box size, and rotation of each 3D box, again by normalizing these values between 0 and 1 and then encoding them using the location tokens. We also added the camera pose prediction tasks using Objaverse objects that were used during pre-training.

We include 3 text detection datasets from COCO-Text~\cite{veit2016cocotext}, including finding the bounding box of an input text string for multiple text strings or finding and listing all text along with their bounding boxes in an image.

Lastly, we do keypoint detection using COCO pose data. For keypoint detection, we input a bounding box around a person in the image and train the model to return a list of keypoints for that person. During inference, we first localize all people in the image and then use each returned bounding box as a keypoint query to find that person's keypoints. During training, the model predicts ``MISSING" for keypoints that are not visible (\eg ``right elbow: MISSING"). During inference, we use a masking function over the model's logit to force it to guess a valid point for each keypoint since the keypoint metric does not award points for correctly identifying a keypoint as being not visible.

\subsection{Image Dense Labelling}
We do several image labeling tasks, including surface normal estimation on FramNet~\cite{huang2019framenet}, BlendedMVS~\cite{yao2020blendedmvs} and Taskonomy~\cite{zamir2018taskonomy}, depth on NYU Depth~\cite{nyu_depth}, and optical flow on Flying Chairs~\cite{dosovitskiy2015flyingchairs} and MPI Sintel~\cite{butler2012mpisintel}.

We additionally train on several segmentation tasks: semantic segmentation (segmenting a particular class), localization segmentation (segmenting an object in an input bounding box), and referring expression segmentation (segmenting an object matching a referring expression). Data comes from Open Images, COCO, LVIS, and referring expressions from the COCO refexp datasets~\cite{referitgame,yu2016modeling,nagaraja2016modeling}. To do instance segmentation, as needed for GRIT, we first do localization on the target class and then perform localized segmentation on each returned bounding box.

During inference, we do temperature sampling with a top-p of 0.95 as before, but without classifier-free guidance. For segmentation, we find it beneficial to increase the value of p to 0.97.

\subsection{Video Understanding}
These tasks require generating text in response to a query about a video. For video captioning, we add VATEX~\cite{wang2019vatex} and MSR-VTT~\cite{xu2016msr}. For action classification (video tagging), we add UCF101~\cite{soomro2012ucf101}, Kinetics-710~\cite{li2022uniformerv2}, Something-Something v2~\cite{goyal2017something} and EPIC-KITCHENS-100~\cite{Damen2022rescaling}. We also use examples from EPIC-KITCHENS-100 for action anticipation. For video question answering, we add MSRVTT-QA~\cite{xu2017video}, MSVD-QA~\cite{xu2017video}, STAR~\cite{wu2021star} and M4-ViteVQA~\cite{zhao2022towards}. Lastly, we use examples from M$^3$IT and MIMIC-IT for the video instruction following.

To cover the visual content of the entire video with a small number of frames (5), we use the segment-based sampling following~\cite{wang2019temporal}; we first divide the video into five segments of equal duration and then randomly sample one frame from each of the segments during training, and the middle frame at inference. We use the first four frames as the image history input and the final frame as the image input for action classification and video captioning. We empirically found that using the third frame as the image input while using the other frames as the image history input performs better for video question answering.

We use similar prompts to those for image understanding tasks, \eg, ``Write a short description of this video.'', ``The question \{question\} can be answered using the video. A short answer is'' and ``What are they doing in this video? Short answer:'' in video captioning, video question answering, and video tagging, respectively, for ensuring a short answer.

\subsection{Video Sparse Labelling}
We do single object tracking and spatial-temporal action localization on video data. We train on YouTube-BB~\cite{real2017youtube}, LaSOT~\cite{fan2021lasot} and GOT-10k~\cite{huang2019got} by inputting bounding boxes around a target object in each of previous frames and having the model return the next location as a bounding box (``Anticipate the object's next location from all previous images and the location of the object in those frames: \{locations\}.''). We also train the model on AVA~\cite{gu2018ava} by inputting a video snippet consisting of five frames and requiring the model to detect all actions of humans appearing in the middle (third) frame of the video snippet (``Given the temporal context from the video, detect all of the humans performing actions in the image.''). Note that we provide the video snippet, not a single video frame, because some of the actions require temporal context to answer (\eg, stand and sit) correctly. We use the final/middle frame of five consecutive frames in the video as the image input and the other frames as the image history input for single object tracking and action localization, respectively.

\subsection{Audio Understanding}
We train the model on audio tagging and audio captioning tasks. For audio tagging, we add AudioSet~\cite{gemmeke2017audio}, VGG-Sound~\cite{chen2020vggsound}, and MACS. For audio captioning, we use the same datasets as text-to-audio generation, that is, AudioCaps, Clotho, MACS, MusicCaps, and LJSpeech. For audio-visual action classification, we train on Kinetics-Sounds~\cite{arandjelovic2017look} and VGG-Sound.

We again use stylistic hints in the prompts for these tasks. For example, we specify the characteristics of target audio (\eg, ``Describe the music.'' and ``Transcribe the audio to text.'' for MusicCaps and LJSpeech, respectively), enforce a short answer (\eg, ``What is this in the audio? Short answer:'' and ``Give a short description of this audio.''), and specify the kind of class to output for audio tagging, \eg, ``This audio depicts a scene of a'' for MACS. We use the same prompts as video tagging for audio-visual action classification.

We use the same sampling strategy as the video understanding; we sample five audio segments with uniform intervals from the whole audio and use the middle/final audio segment as the audio input while using the other segments as the audio history input for audio classification and audio captioning, respectively.

\subsection{Embodied AI}

While many robot manipulation tasks can be formulated by multimodal prompts
that interleave language and images or video frames, we use VIMA-Bench~\cite{jiang2023vima} to evaluate the robot manipulation skills. 
We use the image input as the initial observation of the environment and the image history for the images or videos in the prompt. The text inputs, or the language instructions, also include special tokens to explicitly express the interleaved multimodal prompt. The action space consists of primitive actions of ``pick and place" for tasks with a suction cup as the end effector or ``push" for tasks with a spatula. Both primitive actions contain two poses and one rotation $\in \mathbb{R}^3$, specifying the start and target states of the end effector.

With the action representation described in~\ref{supp:detail_task_representation}, we seamlessly add large-scale manipulation datasets Language Table~\cite{lynch2023interactive}, BridgeData V2~\cite{walke2023bridgedata}, and FrankaKitchen~\cite{gupta2019relay} with the continuous control in both simulated and real-world environments. The model directly predicts the next action as the text target based on the current observation as image input, previous frames as image history, and language instruction and previous actions as text inputs. 

Due to the non-causality of the model and limited sequence length for the image history, we only added the PointNav task from Habitat~\cite{savva2019habitat} Gibson scenes for the navigation. The model is required to predict the next action, with random augmentation for predicting the next position and rotation state, based on the point goal (positions $\in\mathbb{R}^2$), visual observations, and previous actions and states, if any.

\subsection{Task Augmentation}
In addition to these sources, we derive several additional tasks that use the same supervised annotations as other tasks but require performing slightly different functions. We call this task augmentation. The new tasks include prompts that specify the desired output.
These tasks serve to add diversity to our instruction following data.
We review the task augmentation we construct below.

\boldheader{Segmentation}
We build several augmentations of the segmentation tasks, including (1) segmenting pixels belonging to one of a set of 2-4 categories, possibly including categories that do not exist in the image, (2) segmenting pixels belonging to a class and are within an input bounding box and (3) build a map of pixels that do not belong to a set 1-4 classes. Prompts are designed for these that state the requirement, \eg, ``Show pixels that are part of chair, paper and in 
\textless{}extra\_id\_289\textgreater{} \textless{}extra\_id\_871\textgreater{} \textless{}extra\_id\_781\textgreater{} \textless{}extra\_id\_1156\textgreater{}".

\begin{table*}[t]
\tablefont
\setlength\tabcolsep{3pt}
\renewcommand{\arraystretch}{1.15}
\center
\resizebox{\textwidth}{!}{
\begin{tabular}{l c}
\toprule
Prompt & Model Response \\
\midrule
A video of a \textcolor{red}{\textit{man}} (\textcolor{blue}{\textit{woman}}) saying \uiot~is a model that works with vision, language, audio, and action.  &  \href{https://ai2-prior-uio.s3.us-west-2.amazonaws.com/public/samples/pretrain/pretrain-1-man.wav}{\faVolumeUp} (\href{https://ai2-prior-uio.s3.us-west-2.amazonaws.com/public/samples/pretrain/pretrain-1-woman.wav}{\textcolor{blue}{\faVolumeUp}}) \\
A video of a man playing guitar. \href{https://ai2-prior-uio.s3.us-west-2.amazonaws.com/public/samples/pretrain/shred_guitar.wav}{\faVolumeUp} \parbox[c]{10em}{\includegraphics[height=0.7in]{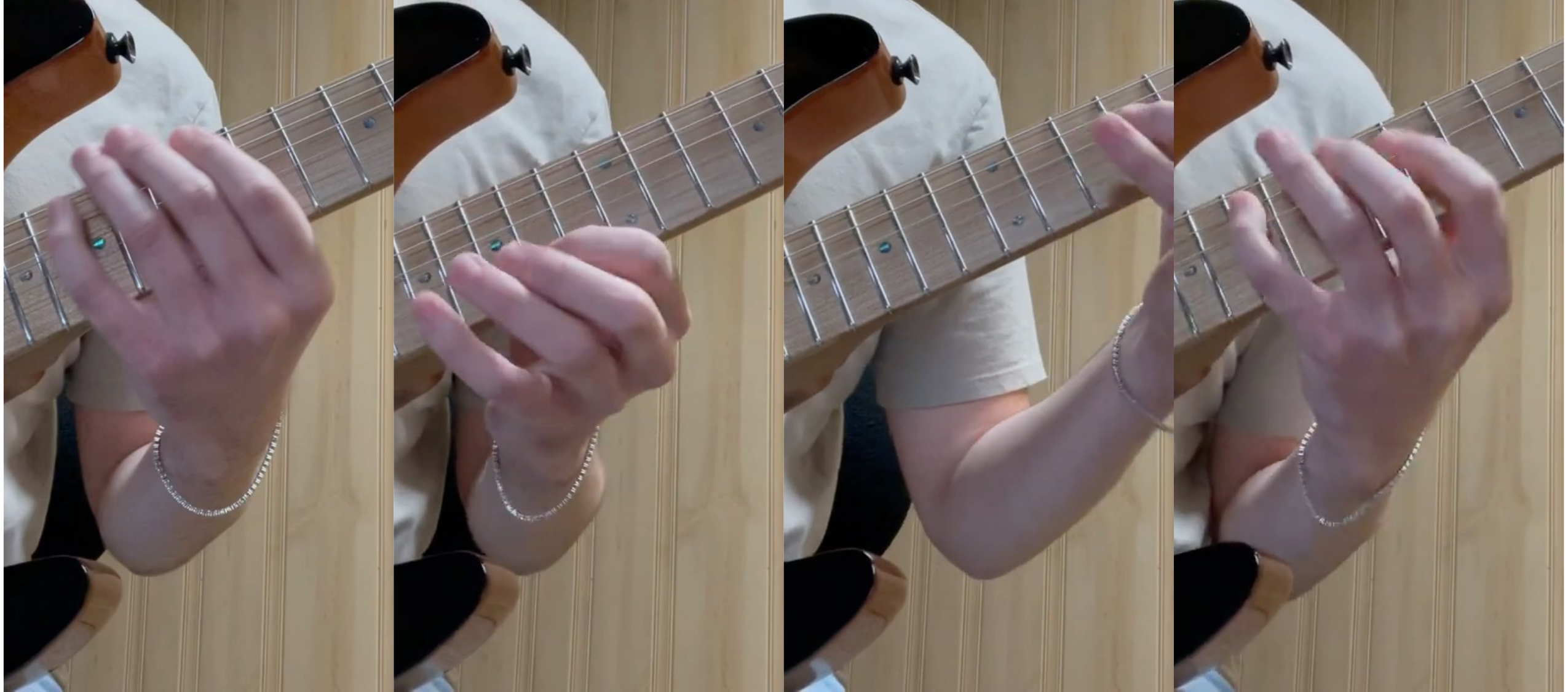}} & \href{https://ai2-prior-uio.s3.us-west-2.amazonaws.com/public/samples/pretrain/guitar_out.wav}{\faVolumeUp} \\
\bottomrule
\end{tabular}}
\vspace{-2mm}
\caption{Audio generation samples from the pre-trained model.}
\vspace{-1mm}
\label{tab:pretrain-audio-gen}
\end{table*}

\begin{figure*}[t]
    \centering
    \includegraphics[width=0.95\textwidth]{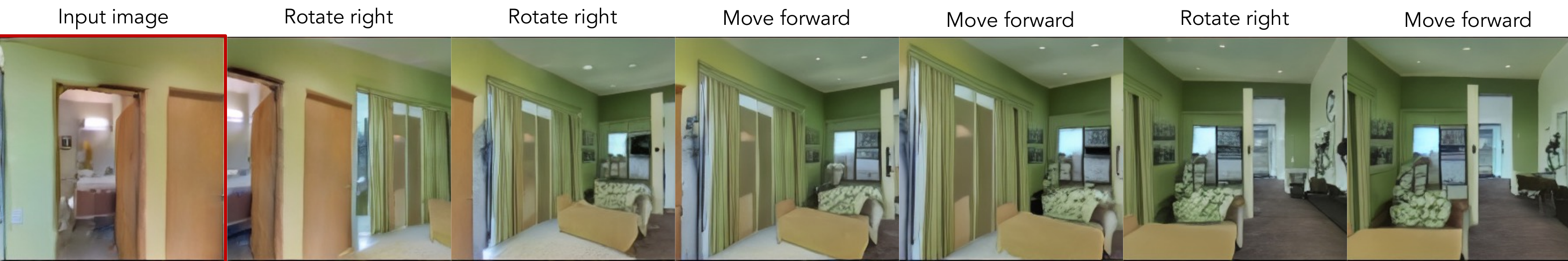}
    \captionof{figure}{\small Future frame prediction samples from the pre-trained model. Given the initial input image and action, the model can generate a plausible frame that reflects the result of the action. }
    \label{fig:pretrain-eai}
    \vspace{-2mm}
\end{figure*}




\boldheader{Detection and Referring Expression}
For detection, localization, and referring expressions, we also train the model to output various properties of the output bounding boxes instead of the boxes themselves. Properties include the width, height, area, left/right/top/bottom half, center coordinates, distance from the left/right/top/bottom edge of the image, or the coordinates of different corners of the bounding box. We also change the format of the output bounding box (\eg, $[x_1, y_1, w, h]$ instead of $[y_1, x_1, y_2, x_2]$ format), and change whether the model labels the boxes with the object category or not.

For detection, we train the model to detect any object belonging to a set of 1-4 classes. For referring expressions, we train the model to locate multiple referring expressions from a single query. In this case, we sometimes train the model to predict a property of both referenced boxes instead of outputting the directly, for example, which box is the smallest, which is the largest, the area of intersection, a box containing both boxes, etc.

\boldheader{Relationship Prediction}
We train the model to list all relationships between a particular object in the image and any other object. A bounding box and category specify the target object. Similarly, we train the model to predict all relationships between any instance of a particular class of objects and any other object in the image.

\boldheader{Captioning}
For captioning, we train the model to generate a caption that is longer or shorter than a given character or word length or contains a particular word or set of words. We also randomly require the caption to start with a particular prefix. Again, these requirements are specified in the prompt, for example, ``Generate a caption longer than five words for this image. Start your output with the text `My caption is:'".

\boldheader{Surface Normal Estimation}
For surface normal estimation, we train the model to generate RGB images that encode the pixel orientation differently. This includes changing which RGB channels correspond to the x, y, and z orientations and only including a subset of those orientations. We also include tasks that require specifying the x, y, and z orientation at a particular point specified in the prompt using location tokens. Finally, we include tasks requiring segmentation masks over pixels with particular orientations, \eg, ``Build a binary mask over surfaces with an upward orientation".

\begin{figure}[t]
    \centering
    \includegraphics[width=1.0\columnwidth]{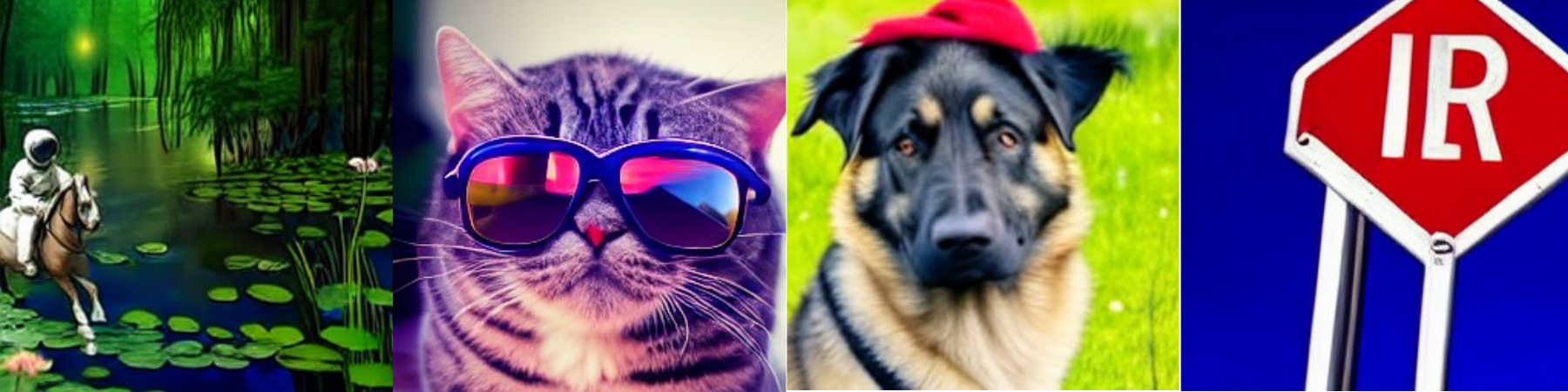}
    \captionof{figure}{\small Image generation samples from the pre-trained model. Prompts from left to right: 1) An image of an astronaut riding a horse in the forest. There is a river in front of them with water lilies, Fantasy, HD. 2) A image of a cat wearing sunglasses, HD. 3) A image of a black german shepherd wearing a red beret, HD. 4) An image of a stop sign in a Fantasy style with the text ``1991''.}
    \label{fig:pretrain-image-gen}
    \vspace{-2mm}
\end{figure}

\boldheader{Embodied AI}
We further augment the embodiment datasets with the video QA and goal image generation tasks. The QA augmentation aims for the robot's planning and affordance. For example, given a robot video trajectory, the model is supposed to predict the plan (caption), or whether a given action is reasonable from the language instruction. Applying image editing in embodied space, we further let the model generate the goal or subgoal images based on the initial visual observation in the image input and the language prompt in the text input. While recent works show that embodiment QA with VLM~\cite{driess2023palm, sermanet2023robovqa} and (sub-)goal generation with diffusion model~\cite{mishra2023generative} are effective in the decision-making downstream tasks, our model combines the both augmentation strategies.

\section{Experiment Details}
\label{supp:results}

\begin{table*}[tb]
\setlength\tabcolsep{3pt}
\renewcommand{\arraystretch}{1.25}
\center
  \resizebox{\textwidth}{!}{
  \begin{tabular}{c l @{\hspace{1.0\tabcolsep}} c  c c c c c  c c  c c c c c c  c c c c c}
\toprule
 &  & \multicolumn{2}{c}{Categorization} &  \multicolumn{2}{c}{Localization} & \multicolumn{2}{c}{VQA} & \multicolumn{2}{c}{Refexp} & \multicolumn{2}{c}{Segmentation} & \multicolumn{2}{c}{Keypoint} & \multicolumn{2}{c}{Normal} & \multicolumn{2}{c}{All} \\
  \cmidrule(r){3-4}
  \cmidrule(r){5-6}
  \cmidrule(r){7-8}
  \cmidrule(r){9-10}
  \cmidrule(r){11-12}
  \cmidrule(r){13-14}
  \cmidrule(r){15-16}
  \cmidrule(r){17-18}
 & & ablation & test & ablation & test & ablation & test & ablation & test & ablation & test & ablation & test & ablation & test & ablation & test \\
 \midrule
    \band \small\texttt{0} & NLL-AngMF [\citenum{bae2021estimating}] & - & - & - & - & - & - & - & - & - & - & - & - & \textbf{49.6} & \textbf{50.5} & 7.2 & 7.1 \\
 \small\texttt{1} & Mask R-CNN [\citenum{He17Mask}] & - & - & 44.7 & 45.1 & - & - & - & - & 26.2 & 26.2 & 70.8 & 70.6 & - & - & 20.2 & 20.3 \\
    \band \small\texttt{2} & GPV-1 [\citenum{Gupta2021GPV}] & 33.2 & 33.2 & 42.8 & 42.7 & 50.6 & 49.8 & 25.8 & 26.8 & - & - & - & - & - & - & 21.8 & 21.8 \\
   \small\texttt{3} & CLIP [\citenum{radford2021learning}]& 48.1 & - & - & - & - & - & - & - & - & - & - & - & - &- & 6.9 & - \\
  \band \small\texttt{4} & OFA$_{\texttt{LARGE}}$ [\citenum{wang2022OFA}] & 22.6 & - & - & - & 72.4 & - & 61.7 & - & - & - & - & - & - & - & 22.4 & - \\ 
   \small\texttt{5} & GPV-2 [\citenum{Kamath2022WeblySC}] & 54.7 & 55.1 & 53.6 & 53.6 & 63.5 & 63.2 & 51.5 & 52.1 & - & - & - & - & - & - & 31.9 & 32.0 \\
    \small\texttt{5} & DINO + SAM~\cite{Kirillov_2023_ICCV,oquab2023dinov2}& - & - & 66.0 & 66.0 & - & - & - & - & \textbf{60.2} & \textbf{60.1} & - & - & - &- & 18.0 & 18.0 \\
 \midrule
 \band \small\texttt{6} &\uio$_{\texttt{SMALL}}$ & 42.6 & - & 50.4 & - & 52.9 & - & 51.1 & - & 40.7 & - & 46.5 & - & 33.5 & - & 45.4 & - \\
   \small\texttt{7} &\uio$_{\texttt{BASE}}$ & 53.1 & - & 59.7 & - & 63.0 & - & 68.3 & - & 49.3 & - & 60.2 & - & 37.5 & - & 55.9 & - \\
\band \small\texttt{8} &\uio$_{\texttt{LARGE}}$ & 57.0 & - & 64.2 & - & 67.4 & - & 74.1 & - & 54.0 & - & 67.6 & - & 40.2 & - & 60.7 & -  \\
  \small\texttt{9} &\uio$_{\texttt{XL}}$ & 61.7 & 60.8 & 67.0 & 67.1 & \textbf{74.5} & \textbf{74.5} & \textbf{78.6} & \textbf{78.9} & 56.3 & 56.5 & 68.1 & 67.7 & 45.0 & 44.3 & 64.5 & 64.3 \\
\midrule
\band \small\texttt{9} & \uiolarge & 70.1 & - & 66.1 & - & 67.6 & - & 66.6 & - & 53.8 & - & 56.8 & - & 44.5 & - & 60.8 & - \\
\small\texttt{10} & \uioxl & 74.2 & - & 69.1 & - & 69.0 & - & 71.9 & - & 57.3 & - & 68.2 & - & 46.7 & - & 65.2 & - \\
\band \small\texttt{11} & \uioxxl & \textbf{74.9} & \textbf{75.2} & \textbf{70.3} & \textbf{70.2} & 71.3 & 71.1 & 75.5 & 75.5 & 58.2 & 58.8 & \textbf{72.8} & \textbf{73.2} & 45.2 & 44.7 & \textbf{66.9} & \textbf{67.0} \\
\bottomrule
\end{tabular}}
\caption{\small{GRIT results and additional baselines from the GRIT leaderboard.}}
\label{tab:grit_full}
\end{table*}

\subsection{Pre-training Visualization}
\label{supp:pretrain_visualization}
In the main paper, we evaluate the effectiveness of our pre-training by evaluating \uiot~quantitively on a variety of benchmarks. Here, we qualitatively show the visualizations from the pre-trained \uioxxl model. Table ~\ref{tab:pretrain-audio-gen} shows audio generation from text (top) and text + video (bottom). We can see the pre-trained model learns text-to-speech synthesis through video pre-training, and the model can also synthesize music that matches the video input. Figure \ref{fig:pretrain-eai} shows the future frame prediction samples given the initial input image and action sequence. Figure~\ref{fig:pretrain-image-gen} shows the image generation samples given prompts. The model has a good understanding of different objects. However, it struggles to generate the correct text from the given caption. 
 
\subsection{NLP Results}
\label{supp:nlp}

\begin{table}[t]
\footnotesize
    \centering
    \setlength\tabcolsep{3pt}
    \resizebox{\columnwidth}{!}{
    \begin{tabular}{l | c c c c c }
        \toprule
         &  HellaSwag & MMLU & Arc Easy & Arc Cha. & BoolQ\\
         \midrule
        \uiolarge{} & 39.4 & 28.4 & 41.8 & 26.2 & 66.6\\
        \uioxl{} & 49.9 & 29.7 & 49.5 & 31.3 & 72.8 \\ 
        \uioxxl{} & 52.7 & 30.4 & 55.3 & 33.5 & 77.3\\
        \midrule
        Open LLaMA 3B & 52.0 & 23.9 & 69.3 & 33.8 & 67.0 \\
        LLaMA 7B & 57.1 & 42.6 & 76.4 & 43.5 &  77.7 \\
        LLaMA 7B Chat & 57.7 & 47.6 & 74.4 & 44.0 & 80.7\\
        \bottomrule
    \end{tabular}}
    \vspace{-1mm}
    \caption{Results on NLP tasks.}
    \label{tab:nlp}  
\end{table}

We present results on a set of NLP tasks to evaluate the model's language understanding abilities. We 
evaluate using the EleutherAI LM-Eval harness~\cite{eval-harness}, tasks are evaluated zero-shot using the default prompts without any adjustments aside from adding the \texttt{[Text] [S]} prefix used for all text generation tasks. We evaluate on HellaSwag~\cite{zellers2019hellaswag} and a selection of other question answering benchmarks: MMLU~\cite{hendrycks2021measuring}, ARC~\cite{clark2018arc}, and BoolQ~\cite{clark2019boolq}. Results are shown in Table~\ref{tab:nlp}. Baselines were evaluated in the same setting, \ie, zero-shot, with the default prompts, and using LM-Eval.
\uiot{} is generally ahead of Open LLaMA 3B but behind LLaMA. 

\subsection{GRIT Details}
\label{supp:grit}
We present GRIT results in more detail in Table~\ref{tab:grit_full}. Notably, \uiot{} is the first unified model to pass the Masked R-CNN baseline for localization and goes a long way toward closing the gap between SAM and unified models on segmentation.

For GRIT VQA, looking at the scores from GRIT on different VQA subsets, we find that \uiot{} does better on the same-source subset (84.6 vs 58.5) but worse on the new-source subset (57.7 vs 67.2). Same-source questions come from VQA 2.0, and new-source questions come from VG, so the difference can be attributed to the kinds of questions being asked. Qualitatively, it is hard to understand why the scores differ on these subsets since the GRIT ablation questions lack ground truth annotations. However, we notice the models often produce different answers when faced with ambiguous questions (\eg ``What color is black on the horse'', ``hair'' for \uio{} \vs ``mane'' for \uiot{}), so one possibility is that \uiot{} does not match the VG answer style as well as \uio{}, which would likely be due to differences in the kind of VQA training data the models were trained on.

For GRIT localization, we find the model can struggle with images with many instances of the target class, particularly when using beam search. We hypothesize that this is because the probability mass can get split between many similar location tokens, resulting in EOS becoming the most probable token even if its probability is low. As a solution, during inference, we only output EOS if the EOS token itself has a probability of over 0.5, which we find significantly improves the performance on crowded images. In rare cases, we observe this leads to the model generating bounding boxes for the same instance multiple times. As a solution, we apply Non-maximum suppression with a higher threshold of 0.8 to remove these duplicates. We apply this inference trick for localization and when doing the initial localization step for the keypoint and segmentation tasks.
\subsection{Multimodal Benchmark Details}
\label{supp:multimodal_benchmark}

\begin{table}[t]
\centering
\setlength\tabcolsep{4pt}
\resizebox{\columnwidth}{!}{%
\begin{tabular}{l|l|ccccccc}
\toprule
Splits & Metrics & \uiotacronym$_\texttt{XXL}$ & \uiotacronym$_\texttt{XL}$ & \uiotacronym$_\texttt{L}$ & ~\cite{ye2023mplug_owl2} & ~\cite{you2023ferret} & ~\cite{chen2023shikra} \\
\cmidrule(lr){1-2}\cmidrule(lr){3-9}
\multirow{5}{*}{Random}
& Accuracy ($\uparrow$)     &   90.90 & 88.27  & 84.03  & 88.28 &  90.24 & 86.90   \\
& Precision ($\uparrow$)    &   94.30 &  97.44 & 77.73 & 94.34 & 97.72 & 94.40  \\
& Recall ($\uparrow$)        &  87.07 &  78.60 & 95.40 & 82.20 & 83.00 & 79.27   \\
& F1-Score ($\uparrow$)      & \textbf{90.54} & 87.01 & 85.66 & 87.85 & 89.76 & 86.19   \\
& \% Yes    &  46.17 & 40.33 & 61.37 & 44.91 & 43.78 & 43.26  \\
\cmidrule(lr){1-2}\cmidrule(lr){3-9}
\multirow{5}{*}{Popular}
& Accuracy ($\uparrow$)      & 88.17 & 87.47  & 77.27  & 86.20 & 84.90 & 83.97      \\
& Precision ($\uparrow$)    &   89.13 & 95.69 & 70.03 & 89.46 & 88.24 & 87.55    \\
& Recall ($\uparrow$)       &  86.93 & 78.47 & 95.33 & 82.06 & 80.53 & 79.20    \\
& F1-Score ($\uparrow$)     &  \textbf{88.02} & 86.23 & 80.75 & 85.60 & 84.21 & 83.16  \\
& \% Yes   &   48.77 & 41.00  & 68.07 & 45.86 & 45.63 & 45.23  \\
\cmidrule(lr){1-2}\cmidrule(lr){3-9}
\multirow{5}{*}{Adversarial}
& Accuracy ($\uparrow$)     &  84.17 & 85.77 & 72.00  & 84.12 & 82.36 & 83.10  \\
& Precision ($\uparrow$)    &   82.17 & 92.01 & 65.00 & 85.54 & 83.60 & 85.60   \\
& Recall ($\uparrow$)       &   87.27   & 78.33 & 95.33 & 82.13 & 80.53 & 79.60  \\
& F1-Score ($\uparrow$)     &  \textbf{84.64} & 84.62 & 77.30 & 83.80 & 82.00 & 82.49  \\
& \% Yes                   &  53.10 & 42.57 & 73.30 & 48.00 & 48.18 & 46.50  \\
\bottomrule
\end{tabular}%
}
\caption{\small Object hallucination benchmark POPE results, in comparison with
mPLUG-Owl2~\cite{ye2023mplug_owl2}, Ferret~\cite{you2023ferret}, and Shikra~\cite{chen2023shikra}.
}
\label{tab:pope_results}
\end{table}
\begin{table}[t]
\footnotesize
    \centering
    \setlength\tabcolsep{7pt}
    \resizebox{0.7\columnwidth}{!}{
    \begin{tabular}{l | c c c }
        \toprule
         &  Img & Video & All\\
                 \midrule
        InstructBLIP~\cite{dai2023instruct_blip} & 58.8 & 38.1 & 53.4 \\
        VideoChat-7B~\cite{Li2023VideoChatCV} & 39.0 & 33.9 & 37.6 \\
        Otter-7B~\cite{Li2023OtterAM} & 42.9 & 30.6 & 39.7 \\
        Qwen-VL-7B~\cite{bai2023qwen} & 62.3 & 39.1 & 56.3 \\
        Qwen-VL-chat-7B~\cite{bai2023qwen} & 65.4 & 37.8 & 58.2 \\
        mPLUG-Owl2-7B~\cite{ye2023mplug_owl2} & 64.1 & 39.8 & 57.8 \\
        LLaVA-1.5-7B~\cite{llava15} & - & - & 58.6 \\
        LLaVA-1.5-13B~\cite{llava15} & 68.2 & 42.7 & 61.6 \\
        \midrule
        \uiot{}$_\texttt{L}$ & 56.0 & 37.5 & 51.1 \\
        \uiot{}$_\texttt{XL}$ & 64.1 & 45.6 & 60.2 \\
        \uiot{}$_\texttt{XXL}$ & \textbf{65.7} & \textbf{46.8} & \textbf{61.8} \\
        \bottomrule
    \end{tabular}}
    \caption{Results on SEED-Bench~\cite{li2023seed}. Our XXL model outperforms all 7B vision language models and is even slightly better than the LLaVA-1.5 13B model.}
    \label{tab:seed-res}
\end{table}

We now provide the breakdown results for the evaluation-only multimodal benchmarks, POPE~\cite{Li2023EvaluatingOH} and SEED-Bench~\cite{li2023seed}. POPE is the object hallucination benchmark, requiring `yes' or `no' answers. As shown in Table~\ref{tab:pope_results}, our largest model achieves the highest F1 score in all 3 dimensions. Interestingly, smaller models favored `no' responses, possibly due to a bias from negative examples encountered during the instruction tuning phase. SEED-Bench offers 19k multiple-choice questions with human annotations for evaluating multimodal models across 12 dimensions, including spatial (Image) and temporal (Video) understanding. As shown in Table~\ref{tab:seed-res}, our XXL model outperforms all other 7B vision/video language models, and is even slightly better than the LLaVA-1.5 13B model. Notably, our XL (3B) model has already outperformed all other counterparts in the temporal understanding split. 
While recent video language models~\cite{video_chatgpt, Li2023VideoChatCV, Li2023OtterAM} have shown proficiency in conventional video tasks like video tagging and captioning, their performance in SEED-Bench's temporal understanding is even worse than that of vision language models, which might be attributed to their limited instruction-following capabilities.

\subsection{Image Generation Details}
\label{supp:image_gen}

\begin{figure}[t]
    \centering
    \includegraphics[width=1.0\columnwidth]{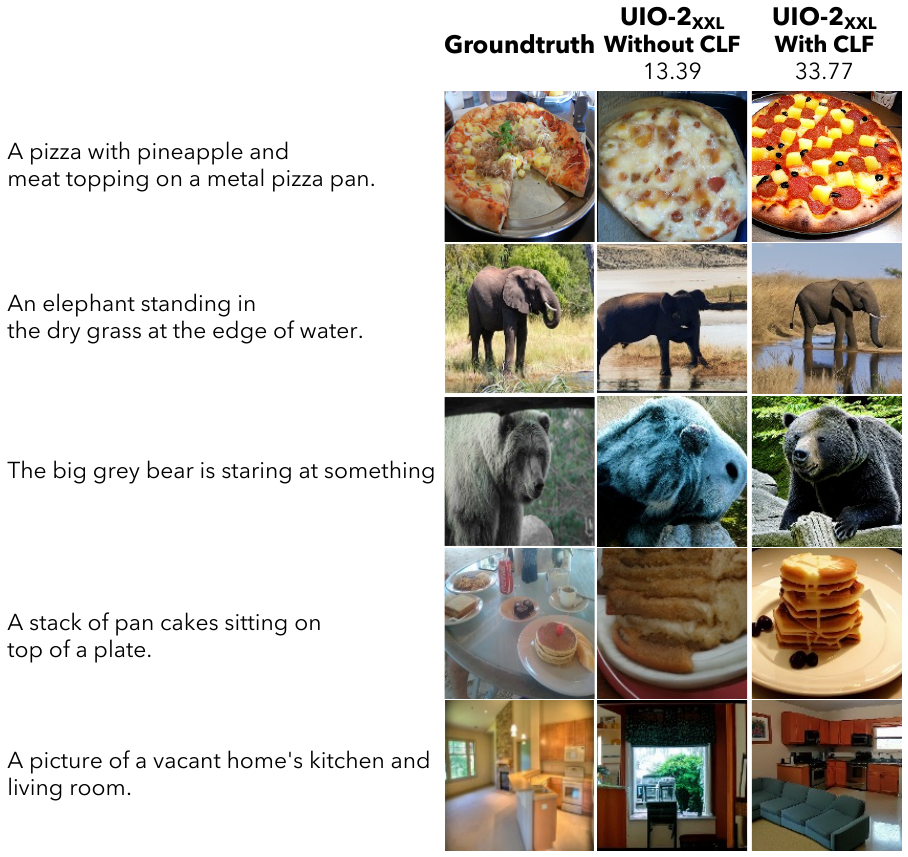}
    \captionof{figure}{\small Samples generated by \uiotacronym$_\texttt{XXL}$ for the MS COCO captions~\cite{coco}. While the classifier-free guidance~\cite{ho2021classifier} significantly boosts image quality and fidelity, it achieves poor FID.}
    \label{fig:coco_examples}
\end{figure}

\begin{figure*}[t]
    \centering
    \includegraphics[width=0.95\textwidth]{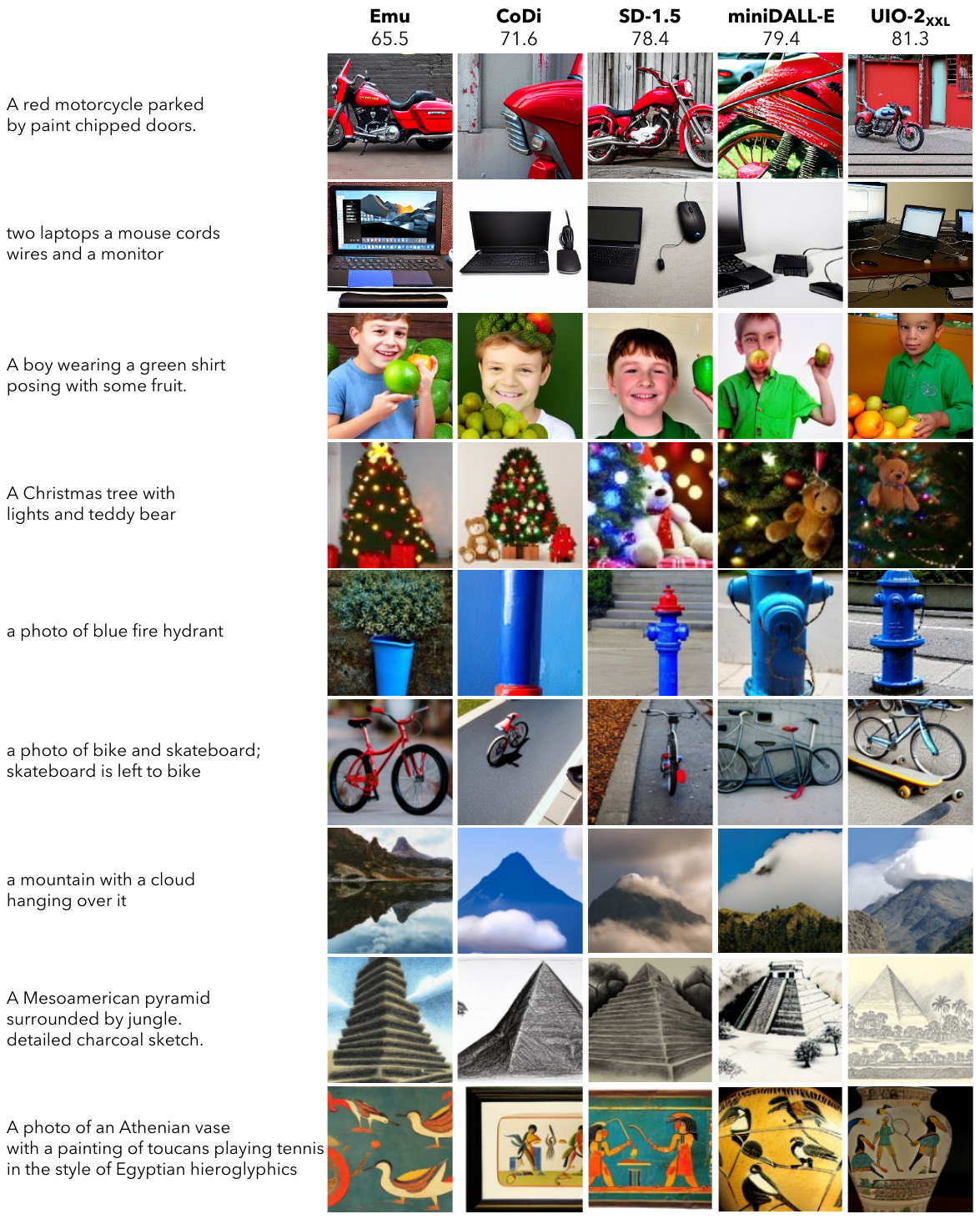}
    \captionof{figure}{\small Samples generated for the TIFA bnechmark captions~\cite{hu2023tifa}. Some of the images generated by baselines (\eg, rows 1-2, 6, 9) have high quality but do not fully follow the input text while \uiot{} generates faithful images.}
    \label{fig:tifa_examples}
\end{figure*}

\begin{table*}[t]
\tablefont
\setlength\tabcolsep{3pt}
\renewcommand{\arraystretch}{1.15}
\center
\resizebox{\textwidth}{!}{
\begin{tabular}{l c}
\toprule
Prompt & Model Response \\
\midrule
What is the sound of a duck quacking? & \href{https://ai2-prior-uio.s3.us-west-2.amazonaws.com/public/samples/a+duck+quacking.wav}{\faVolumeUp} \\
A bell is ringing loudly and quickly. & \href{https://ai2-prior-uio.s3.us-west-2.amazonaws.com/public/samples/a+bell+is+ringing+loudly+and+quickly.wav}{\faVolumeUp} \\
\midrule
Speak out the text ``Police protection was better and more effective;'' for me.  &  \href{https://ai2-prior-uio.s3.us-west-2.amazonaws.com/public/samples/tts_output_1.wav}{\faVolumeUp} \\
Speak: Printing, then, for our purpose, may be considered as the art of making books by means of movable types. & \href{https://ai2-prior-uio.s3.us-west-2.amazonaws.com/public/samples/tts_output_2.wav}{\faVolumeUp} \\
Read ``Many animals of even complex structure which live parasitically within others are wholly devoid of an alimentary cavity.'' & \href{https://ai2-prior-uio.s3.us-west-2.amazonaws.com/public/samples/tts_output_3.wav}{\faVolumeUp} \\
\midrule
Generate the music based on the description ``Slow tempo, bass-and-drums-led reggae song.'' & \href{https://ai2-prior-uio.s3.us-west-2.amazonaws.com/public/samples/music_gen_1.wav}{\faVolumeUp} \\
Based on the given description ``Industrial techno sounds, repetitive, hypnotic rhythms'', produce a corresponding piece of music. & \href{https://ai2-prior-uio.s3.us-west-2.amazonaws.com/public/samples/music_gen_2.wav}{\faVolumeUp} \\
\bottomrule
\end{tabular}}
\caption{Audio generation examples. \uiot{} can generate not only environmental sound (rows 1-2), but also speech (rows 3-5) and music (rows 6-7). Note that some of the outputs longer than 4.08 seconds have discontinuity in sound, or changes in tone, speed or melody (rows 4-5, 7). Since our model can output 4.08-second audio at a time, we complete the audio clip by using any previously generated clips as additional input. Click \textcolor{magenta}{\faVolumeUp}~for audio samples.}
\label{tab:audio-gen-ex}
\end{table*}

Figure~\ref{fig:tifa_examples} shows generated images for the TIFA benchmark captions~\cite{hu2023tifa} using several baselines as well as \uioxxl{}. We use the official implementation code (Emu~\cite{sun2023emu} and CoDi~\cite{tang2023codi}) or the images shared in the official GitHub repository of TIFA\footnote{\url{https://github.com/Yushi-Hu/tifa/tree/main/human_annotations}} (Stable Diffusion v1.5~\cite{SD15} and miniDALL-E~\cite{mindalle}) for baselines. All the baselines except miniDALL-E use the Stable Diffusion decoder trained on large-scale, high-quality image datasets, generating images of high fidelity. However, they often generate images that do not fully follow the input captions while \uiot{} generates faithful images.

For text-to-image generation on MS COCO~\cite{coco}, we follow the standard convention~\cite{zhu2019dm}; we evaluate on a subset of 30K captions sampled from the validation set.\footnote{We use the evaluation code at \url{https://github.com/MinfengZhu/DM-GAN}} Following~\cite{ding2021cogview}, we generate 8 images for each caption and select the best one using CLIP text-image similarity~\cite{radford2021learning}. Despite classifier-free guidance~\cite{ho2021classifier} resulting in generated images of qualitatively higher quality, the computed FID score~\cite{NIPS2017_8a1d6947} is significantly worse compared to what would have been achieved without employing it (33.77 vs 13.39); see Figure~\ref{fig:coco_examples}.

\subsection{Audio Generation Details}
\label{supp:audio_gen}

\begin{table}[t]
\tablefont
\footnotesize
\setlength\tabcolsep{3pt}
\renewcommand{\arraystretch}{1.15}
\center
\resizebox{\columnwidth}{!}{
\begin{tabular}{l l c}
\toprule
Input Image & Prompt & Model Response \\
\midrule
\parbox[c]{1em}{\includegraphics[width=0.6in]{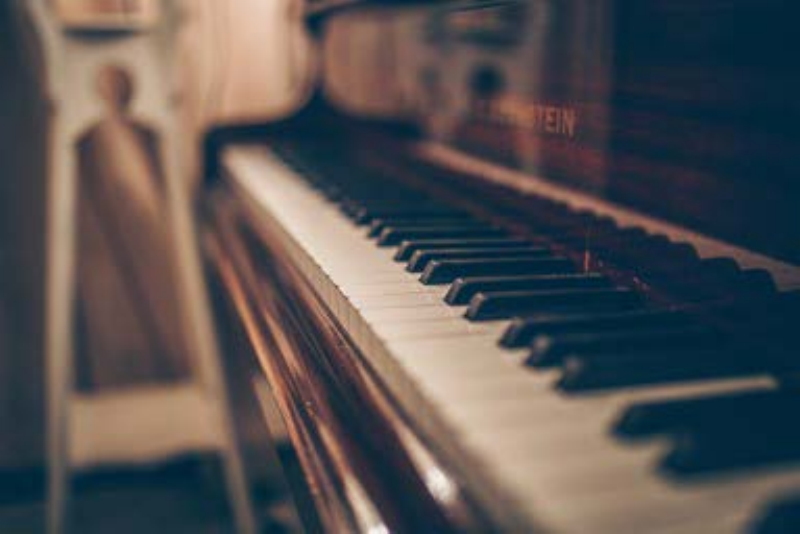}} &
What is the sound of this instrument? & \href{https://ai2-prior-uio.s3.us-west-2.amazonaws.com/public/samples/piano_sound.wav}{\faVolumeUp} \\
\parbox[c]{1em}{\includegraphics[width=0.6in]{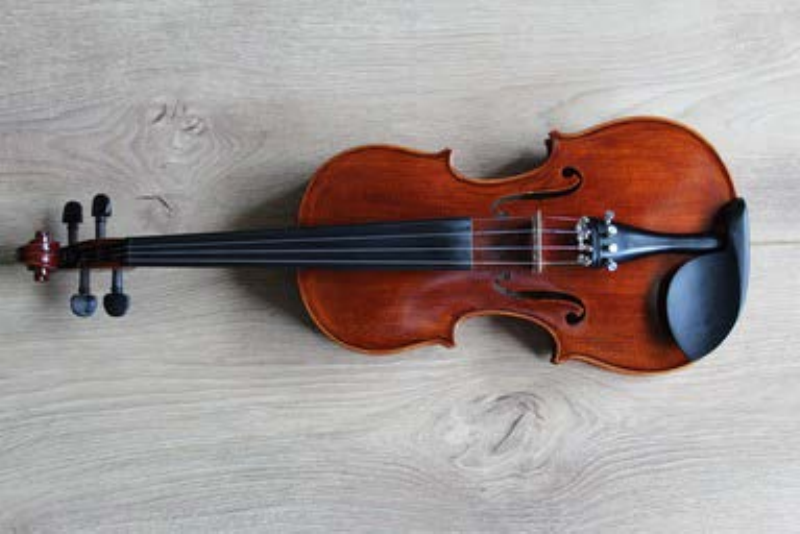}} &
What is the sound of this instrument? & \href{https://ai2-prior-uio.s3.us-west-2.amazonaws.com/public/samples/violin_sound.wav}{\faVolumeUp} \\
\midrule
\parbox[c]{1em}{\includegraphics[width=0.6in]{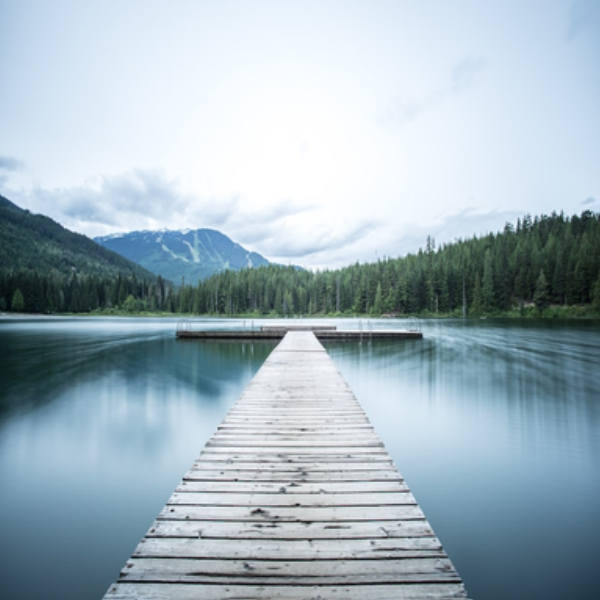}} &
Generate music about this scene. & \href{https://ai2-prior-uio.s3.us-west-2.amazonaws.com/public/samples/dock.wav}{\faVolumeUp} \\
\parbox[c]{1em}{\includegraphics[width=0.6in]{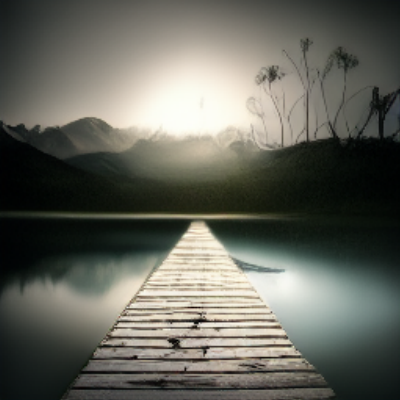}} &
Generate music about this scene. & \href{https://ai2-prior-uio.s3.us-west-2.amazonaws.com/public/samples/dock_haunted.wav}{\faVolumeUp} \\
\midrule
\parbox[c]{1em}{\includegraphics[width=0.6in]{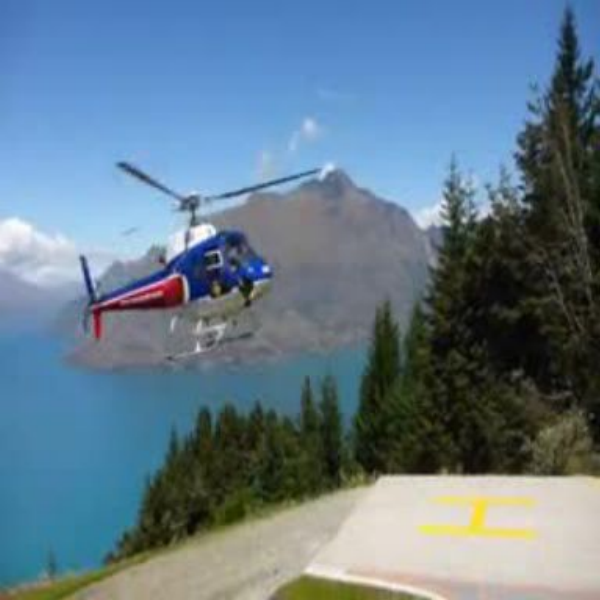}} &
\href{https://ai2-prior-uio.s3.us-west-2.amazonaws.com/public/samples/visual_sound_localization_input_1.wav}{\faVolumeUp} Locate the bounding boxes of the sound sources in the given image. &
\multicolumn{1}{l}{\parbox[c]{1em}{\includegraphics[width=0.6in]{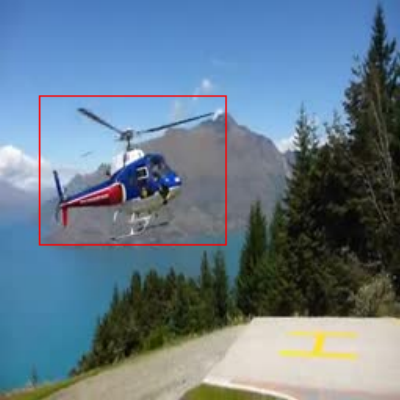}}} \\
\parbox[c]{1em}{\includegraphics[width=0.6in]{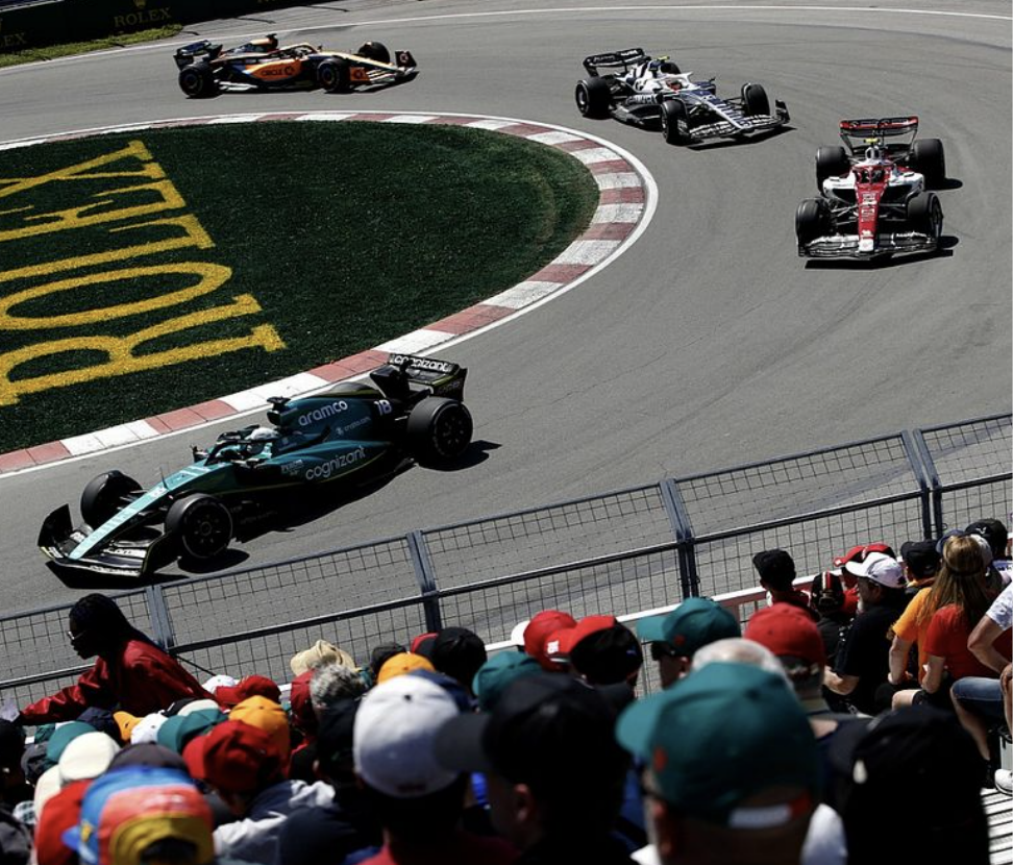}} &
\href{https://ai2-prior-uio.s3.us-west-2.amazonaws.com/public/samples/visual_sound_localization_input_2.mp3}{\faVolumeUp} Identify the locations of the sound sources in the given image. &
\multicolumn{1}{l}{\parbox[c]{1em}{\includegraphics[width=0.6in]{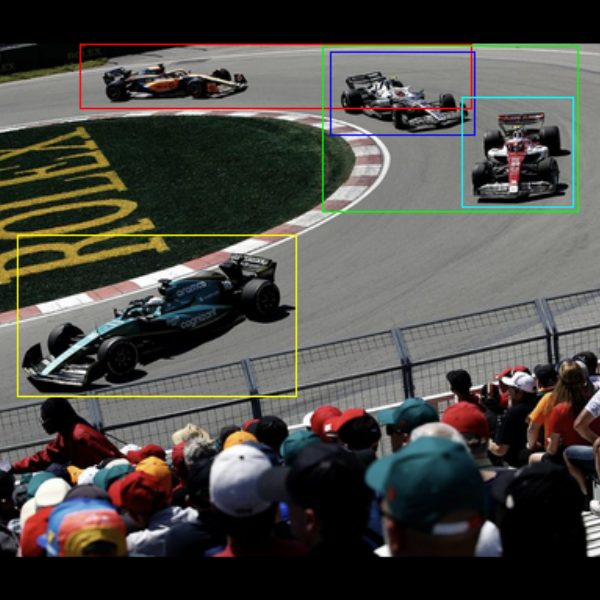}}} \\
\parbox[c]{1em}{\includegraphics[width=0.6in]{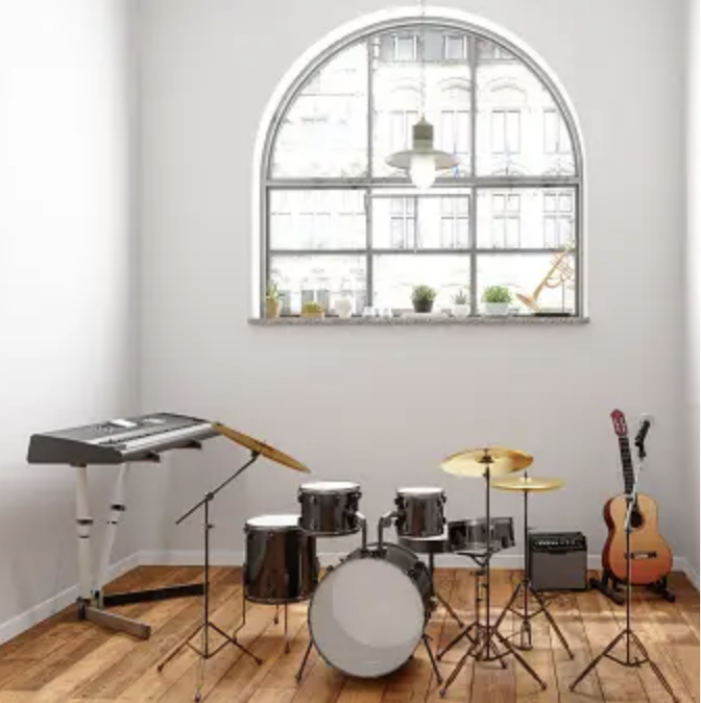}} &
\href{https://ai2-prior-uio.s3.us-west-2.amazonaws.com/public/samples/visual_sound_localization_input_3.mp3}{\faVolumeUp} Identify the locations of the instruments producing the given sound. &
\multicolumn{1}{l}{\parbox[c]{1em}{\includegraphics[width=0.6in]{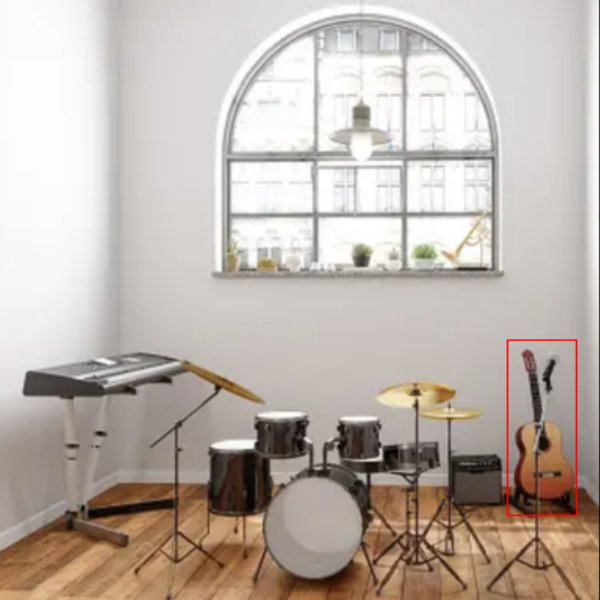}}} \\
\parbox[c]{1em}{\includegraphics[width=0.6in]{figure/visual_sound_localization_input_3.pdf}} &
\href{https://ai2-prior-uio.s3.us-west-2.amazonaws.com/public/samples/visual_sound_localization_input_4.mp3}{\faVolumeUp} Identify the locations of the instruments producing the given sound. &
\multicolumn{1}{l}{\parbox[c]{1em}{\includegraphics[width=0.6in]{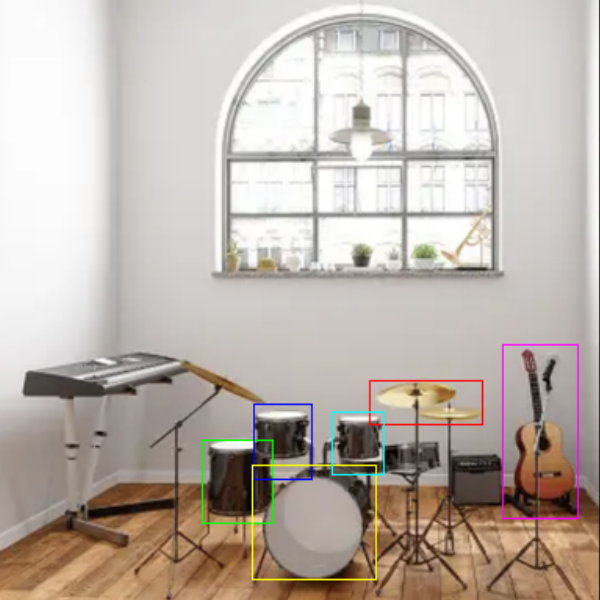}}} \\
\bottomrule
\end{tabular}
}
\caption{Audio-visual qualitative examples showcasing the ability of \uiot{} to reason across modalities. \uiot{} can generate the sound of the instrument in the input image (rows 1-2), and generate the music that matches the mood of the input image (rows 3-4). The last four examples (rows 5-8) show the results of visual sound localization. Note that \uiot{} can accurately identify the instruments that make and do not make sounds (rows 7-8). Click \textcolor{magenta}{\faVolumeUp}~for audio samples.}
\label{tab:audio-visual-ex}
\end{table}

For text-to-audio generation, we evaluate on the AudioCaps~\cite{audiocaps} test set. Note that we cannot do an apples-to-apples comparison with other methods because AudioCaps consists of 10-second audio clips while our model can generate 4.08-second audio at a time. Instead, we evaluate the dataset in the following setup: we first sample four 2-second audio segments, convert them to log-mel-spectrograms with zero-padding, and generate the following audio with the prompt ``Generate the following sound based on what you heard and the description: \{caption\}''. We convert the model output, that is, a log-mel-scaled spectrogram, into a waveform using the pretrained HiFi-GAN, and compare the ground-truth audio and generated audio using computational metrics including Fr{\'e}chet Audio Distance~\cite{kilgour2019fr}, Inception Score~\cite{salimans2016improved} and Kullback–Leibler divergence. We use the same evaluation code as AudioLDM\footnote{\url{https://github.com/haoheliu/audioldm_eval}}~\cite{liu2023audioldm}. We show the audio generation examples in Table \ref{tab:audio-gen-ex} and audio-visual qualitative examples in Table \ref{tab:audio-visual-ex}. 

\subsection{Video and Audio Understanding Details}

\begin{figure*}[t]
    \centering
    \includegraphics[width=0.95\textwidth]{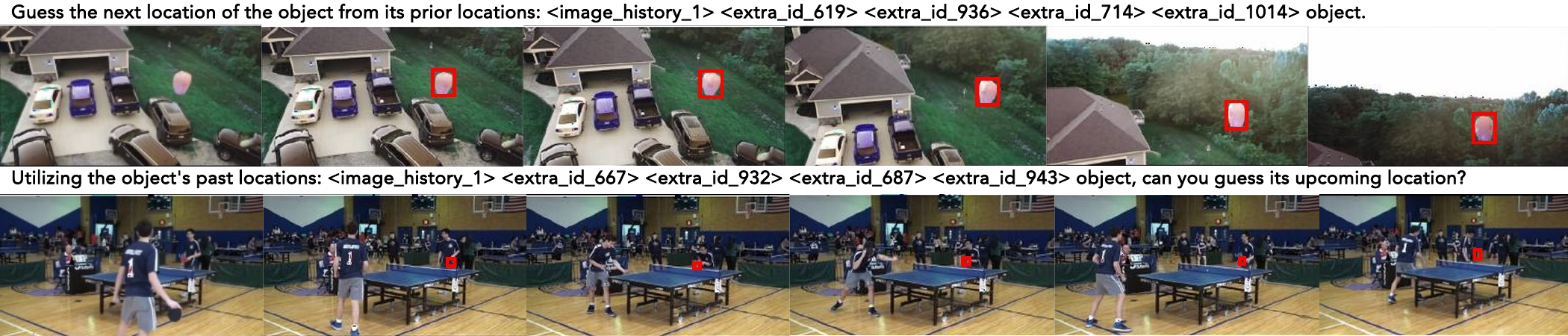}
    \captionof{figure}{\small Single object tracking examples from LaSOT \cite{fan2021lasot}. Given the first input image and the corresponding bounding box of the target object, our model can track the object across video frames.}
    \vspace{-3mm}
    \label{fig:example:tracking}
\end{figure*}

\begin{figure}[t]
    \centering
    \includegraphics[width=1.0\columnwidth]{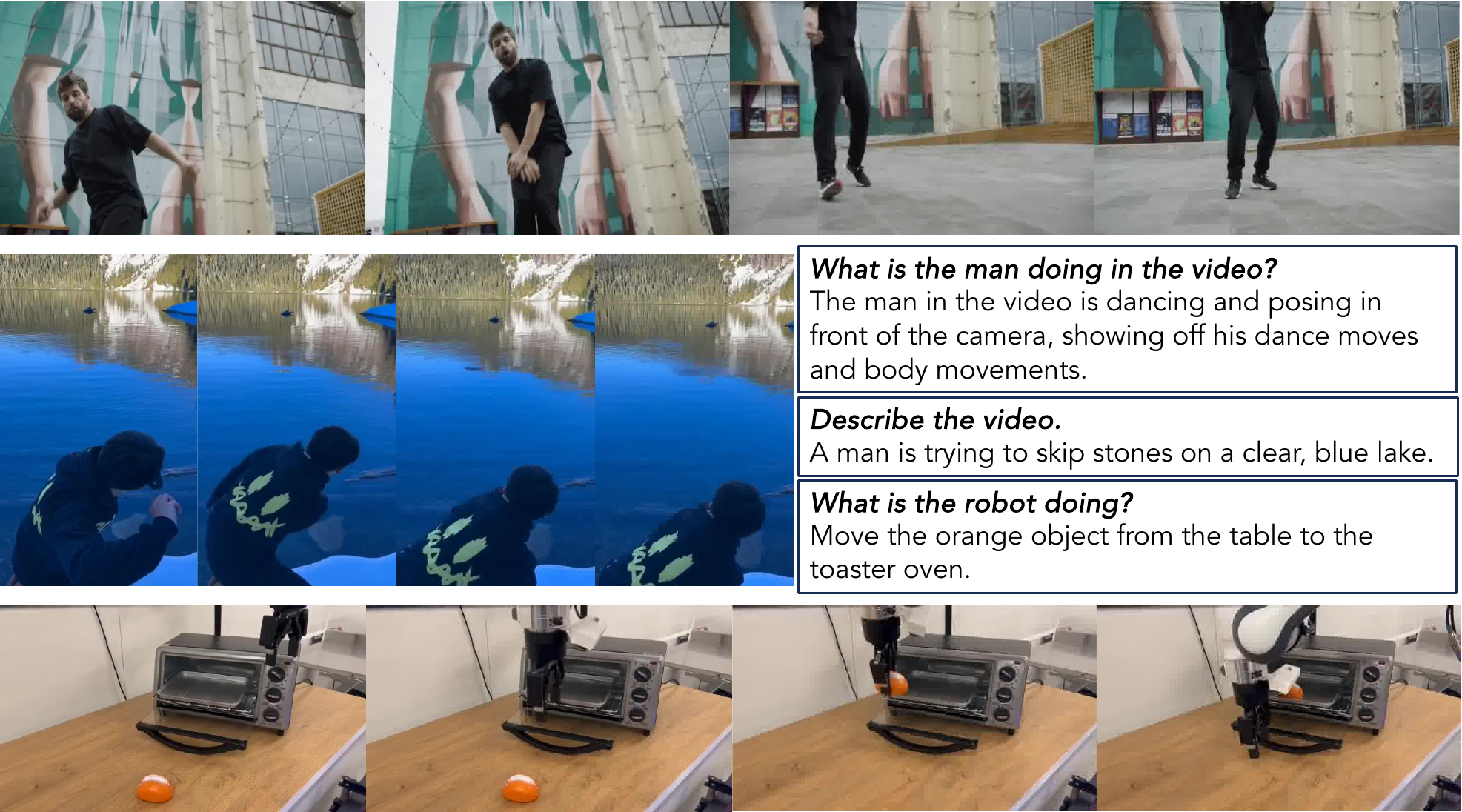}
    \captionof{figure}{\small Video understanding qualitative examples. }
    \label{fig:example:video_understanding}
\end{figure}

\noindent
We consider classification and question-answering tasks as open-ended answer generation and use the Exact Match (EM) to measure the performance. We also tried to formulate the classification task as multiple-choice answering and generate answers by computing the logit for each dataset label and selecting the one with the highest logit, but the performance boost was quite marginal. Note that we do not train our model directly on the Kinetics-400~\cite{kay2017kinetics}; we instead train on Kinetics-710, a mixture of three different datasets belonging to the \textit{Kinetics} family, that is, Kinetics-400, 600, and 700. Our model achieves top-1 accuracy 79.1 (\vs. instruction tuning only: 73.8) when further finetuning on Kinetics-400 for 5 epochs, following~\cite{li2022uniformerv2}. For Kinetics-Sounds, leveraging both audio and visual inputs largely improves performance (audio-visual: 89.3 \vs video-only: 87.4 \vs audio-only: 38.2). For captioning tasks, we use CIDEr~\cite{vedantam2015cider} as the evaluation metric. Figure~\ref{fig:example:video_understanding} shows the qualitative examples for video understanding tasks.

\subsection{Embodiment Details}

\begin{figure}[h]
    \centering
    \includegraphics[width=1.0\columnwidth]{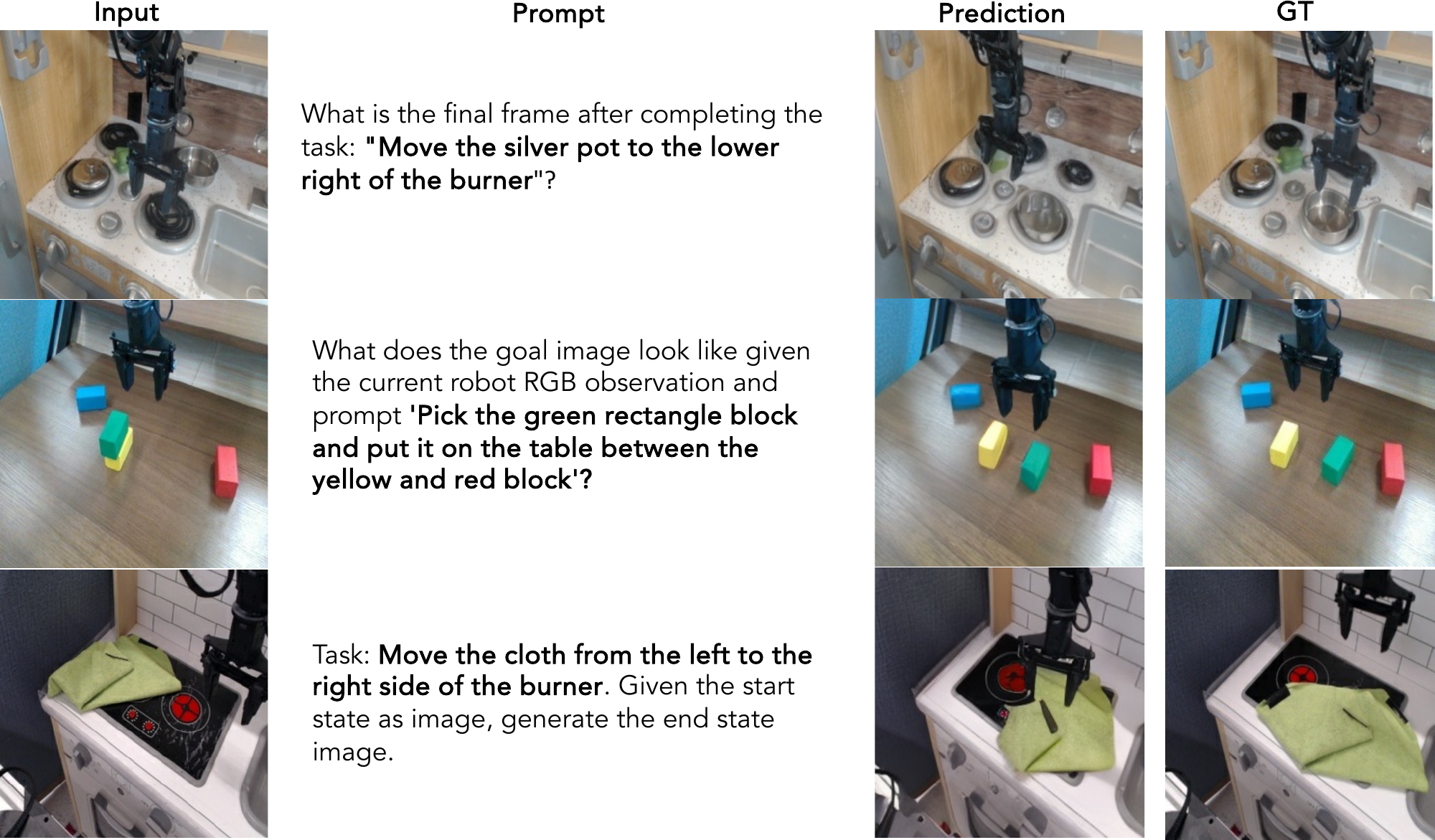}
    \captionof{figure}{\small Future state prediction examples on robotic manipulation tasks. Given the input image and instructions, our model can successfully generate the target state after the prompt instruction. }
    \vspace{-3mm}
    \label{fig:example:robotic_state_prediction}
\end{figure}

\begin{table}[h]
    \small
    \centering
    \setlength\tabcolsep{5pt}
    \resizebox{0.85\columnwidth}{!}{
    \begin{tabular}{l c c c c | c}
        \toprule
         & L1 & L2 & L3 & L4 & Avg. \\ 
         \midrule
         VIMA~\cite{jiang2023vima} & 81.5 & 81.5 & 78.7 & 48.6 & 72.6 \\
         VIMA-Gato~\cite{reed2022generalist} & 57.0 & 53.9 & 45.6 & 13.5 & 42.5 \\
         VIMA-Flamingo~\cite{Alayrac2022FlamingoAV} & 47.4 & 46.0 & 40.7 & 12.1 & 36.6 \\    
         VIMA-GPT~\cite{brown2020language} & 46.9 & 46.9 & 42.2 & 12.1 & 37.0 \\
         \midrule
         \uiot{}$_\texttt{L}$ & 66.9 & 63.8 & 57.5 & 12.6 & 50.2 \\
         \uiot{}$_\texttt{XL}$  & 70.3 & 69.8 & 64.5 & 13.1 & 54.2 \\
         \uiot{}$_\texttt{XXL}$  & 71.3 & 70.4 & 68.0 & 15.5 & 56.3 \\
         \bottomrule
    \end{tabular}}
    \vspace{-1mm}
    \caption{\small Evaluations on VIMA-Bench~\cite{jiang2023vima}}
    \vspace{-3mm}
    \label{tab:vima-full-eval}
\end{table}

In VIMA-Bench~\cite{jiang2023vima}, there are 4 levels of evaluation protocols: L1 object placement, L2 novel combination, L3 novel object, and L4 novel task. Results and comparisons are shown in Table~\ref{tab:vima-full-eval}. The inputs for the autoregressive transformer model VIMA~\cite{jiang2023vima} are object tokens consisting of cropped images and bounding boxes; image patch tokens encoded by ViT for VIMA-Gato~\cite{reed2022generalist}; image patch tokens encoded by ViT, further downsampled by a perceiver module for VIMA-Flamingo~\cite{Alayrac2022FlamingoAV}; and single image token encoded by ViT for VIMA-GPT~\cite{brown2020language}. The output of those baselines is all next-step action prediction. Since our model has to predict all actions at once only with the initial observation, the task setting is then more challenging than the casual policy learning baselines. Nevertheless, our models still outperform counterparts that input image or image patches for all 4 levels and are only behind the object-centric method~\cite{jiang2023vima}. In Figure~\ref{fig:example:robotic_state_prediction}, we show the future state prediction examples on robotic manipulation tasks. Given the input state image and natural language prompt, our model can successfully synthesize the target image state.

\subsection{Other Tasks}

\begin{figure}[t]
    \centering
    \includegraphics[width=1.0\columnwidth]{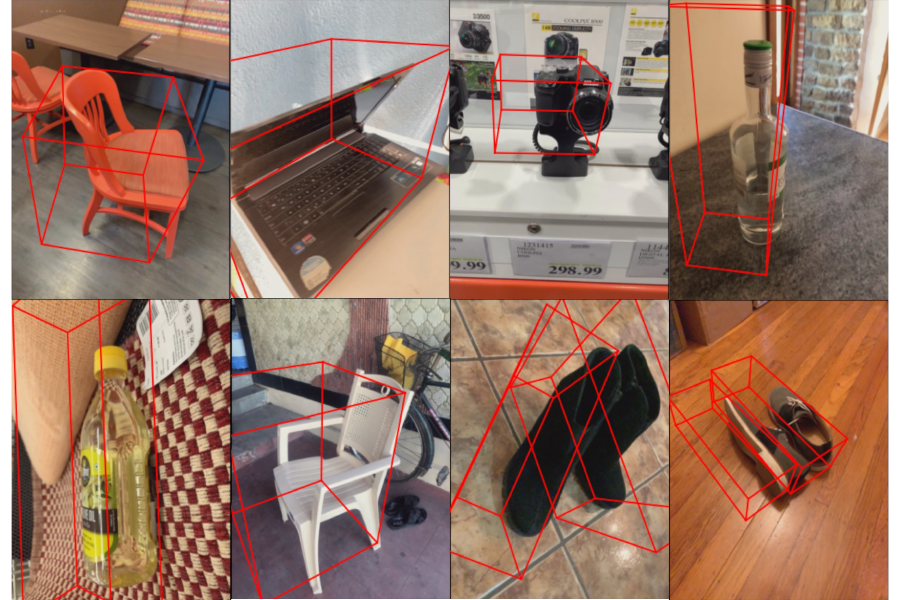}
    \captionof{figure}{\small 3D object detection qualitative examples from Objectron \cite{ahmadyan2021objectron}.}
    \vspace{-3mm}
    \label{fig:example:3d_vis}
\end{figure}

\begin{figure}[t]
    \centering
    \includegraphics[width=1.0\columnwidth]{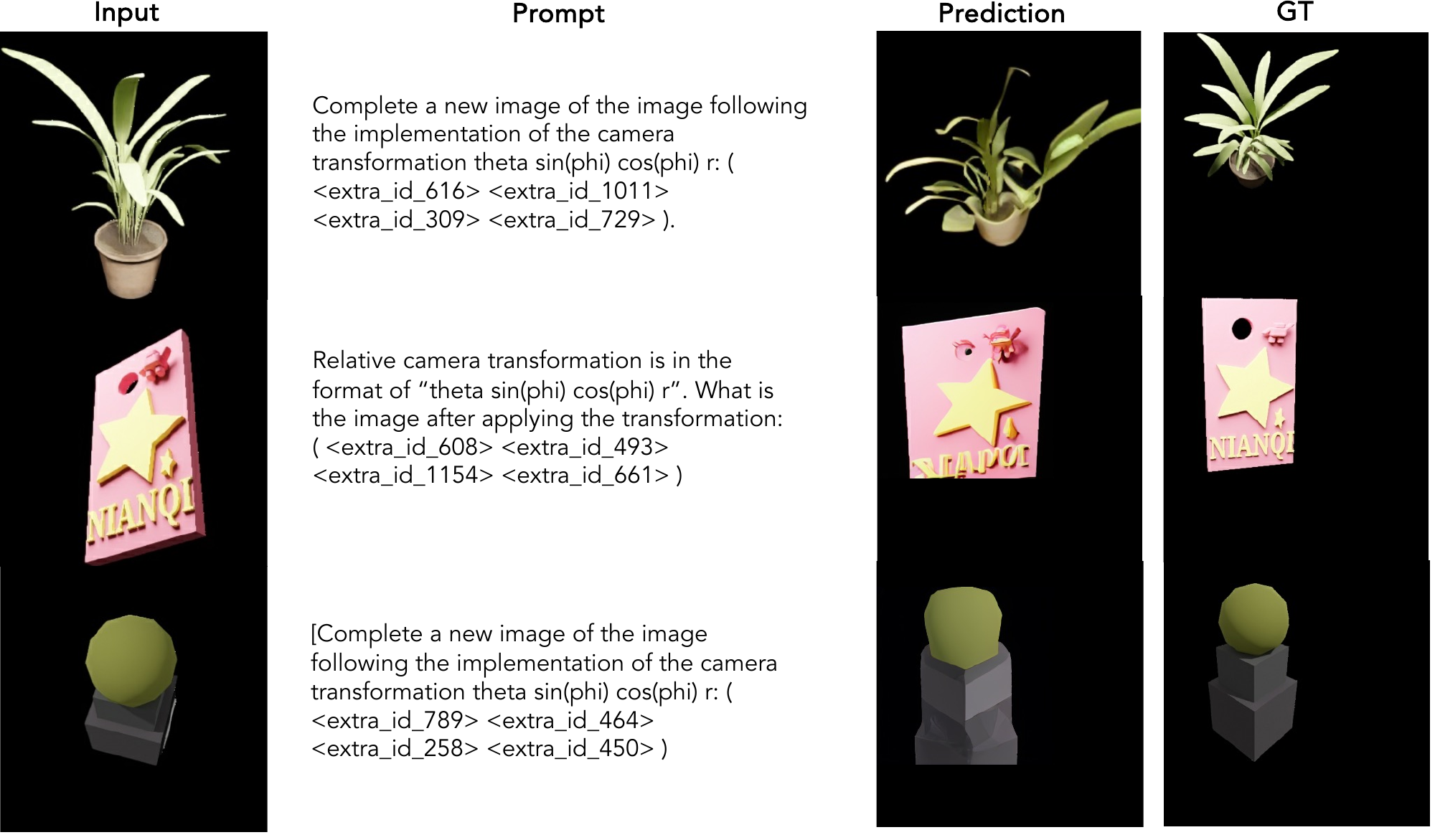}
    \captionof{figure}{\small Image-based 3D view synthesis examples from Objaverse \cite{deitke2023objaverse}.}
    \label{fig:example:objaverse}
    \vspace{-3mm}
\end{figure}

Figure~\ref{fig:example:tracking} shows single object tracking examples from the LaSOT \cite{fan2021lasot} dataset. Note that \uiot{} does not use specific class labels for tracking and tracks small moving objects such as a table tennis paddle well.
Figure~\ref{fig:example:3d_vis} presents qualitative examples of 3D object detection from the Objectron dataset \cite{ahmadyan2021objectron}. As outlined in our main paper, \uiot{} exhibits suboptimal performance in benchmarks for multi-object 3D detection. Additionally, Figure~\ref{fig:example:objaverse} illustrates examples of image-based 3D view synthesis using the Objaverse dataset \cite{deitke2023objaverse}. While the model produces coherent results, it faces challenges in accurately representing relative camera transformations.

\end{document}